\documentclass[nohyperref]{article}

\usepackage{microtype}
\usepackage{graphicx}
\usepackage{subfigure}
\usepackage{booktabs} 
\usepackage{multirow}
\usepackage{longtable}
\usepackage[table]{xcolor}

\usepackage{hyperref}
\usepackage{tabularx}

\usepackage[accepted]{icml2023}

\usepackage{amsmath}
\usepackage{amssymb}
\usepackage{mathtools}
\usepackage{amsthm}

\usepackage[capitalize,noabbrev]{cleveref}
\usepackage{soul}

\theoremstyle{plain}
\newtheorem{theorem}{Theorem}[section]

\newtheorem{example}[theorem]{Example}
\theoremstyle{definition}

\theoremstyle{remark}

\usepackage{csquotes}
\usepackage{dirtytalk}

\icmltitlerunning{Interpretable Neural-Symbolic Concept Reasoning}

\begin{document}

\twocolumn[
\icmltitle{Interpretable Neural-Symbolic Concept Reasoning}



\icmlsetsymbol{equal}{*}

\begin{icmlauthorlist}
\icmlauthor{Pietro Barbiero}{equal,cam}
\icmlauthor{Gabriele Ciravegna}{equal,inria}
\icmlauthor{Francesco Giannini}{equal,siena}
\icmlauthor{Mateo Espinosa Zarlenga}{cam}
\icmlauthor{Lucie Charlotte Magister}{cam}
\icmlauthor{Alberto Tonda}{inra}
\icmlauthor{Pietro Li\'o}{cam}
\icmlauthor{Frederic Precioso}{inria}
\icmlauthor{Mateja Jamnik}{cam}
\icmlauthor{Giuseppe Marra}{equal,kul}
\end{icmlauthorlist}

\icmlaffiliation{cam}{University of Cambridge, Cambridge, UK}
\icmlaffiliation{inria}{Université Côte d’Azur, Inria, CNRS, I3S, Maasai, Nice, France}
\icmlaffiliation{siena}{University of Siena, Siena, Italy}
\icmlaffiliation{inra}{INRA, Université Paris-Saclay, Thiverval-Grignon, France}
\icmlaffiliation{kul}{KU Leuven, Leuven, Belgium}

\icmlcorrespondingauthor{Pietro Barbiero}{pb737@cam.ac.uk}

\icmlkeywords{Machine Learning, ICML, XAI, Explainability, Logic, Neurosymbolic, Interpretability, Concepts}

\vskip 0.3in
]



\printAffiliationsAndNotice{\icmlEqualContribution} 

\begin{abstract}
Deep learning methods are highly accurate, yet their opaque decision process prevents them from earning full human trust. Concept-based models aim to address this issue by learning tasks based on a set of human-understandable concepts.
However, state-of-the-art concept-based models rely on high-dimensional concept embedding representations which lack a clear semantic meaning, thus questioning the interpretability of their decision process.
To overcome this limitation, we propose the \textit{Deep Concept Reasoner} (DCR), the first interpretable concept-based model that builds upon concept embeddings.
In DCR, neural networks do not make task predictions directly, but they build syntactic rule structures using concept embeddings. DCR then executes these rules on meaningful concept truth degrees to provide a final interpretable and semantically-consistent prediction in a differentiable manner.
Our experiments show that DCR: (i) improves up to $+25\%$ w.r.t.\ state-of-the-art interpretable concept-based models on challenging benchmarks (ii) discovers meaningful logic rules matching known ground truths even in the absence of concept supervision during training, and (iii), facilitates the generation of counterfactual examples providing the learnt rules as guidance.
\end{abstract}

\section{Introduction}
The opaque decision process of deep learning (DL) models has failed to inspire human trust despite their state-of-the-art performance across multiple tasks~\citep{rudin2019stop, bussone2015role}, raising ethical~\citep{duran2021afraid, lo2020ethical} and legal~\citep{wachter2017counterfactual, gdpr2017} concerns. 
For this reason, interpretability is now a core research topic in the field of responsible AI~\citep{rudin2019stop}.



\begin{figure*}[!t]
    \centering
    \includegraphics[width=.5\textwidth]{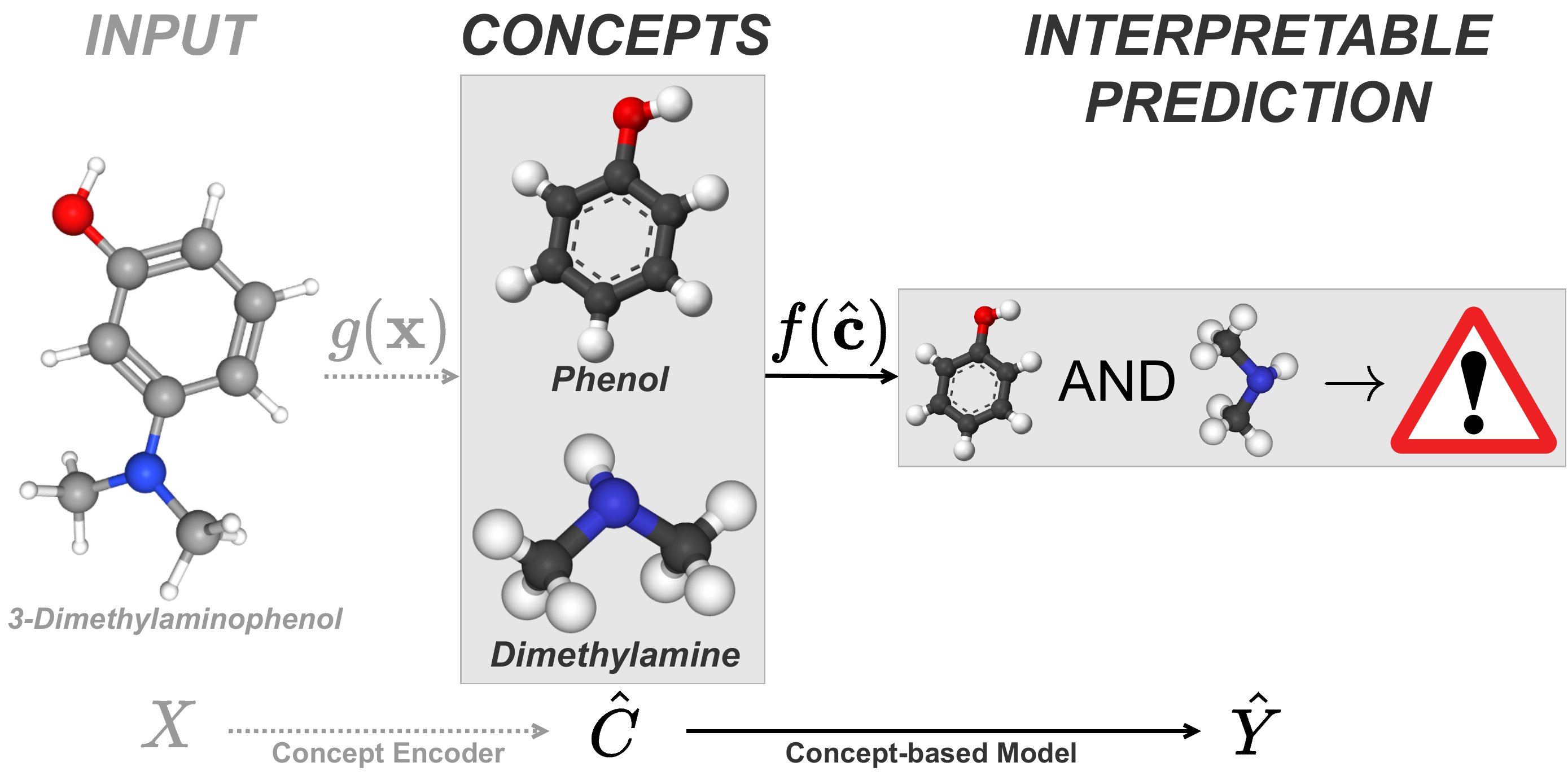} \qquad
    \includegraphics[width=.2\textwidth]{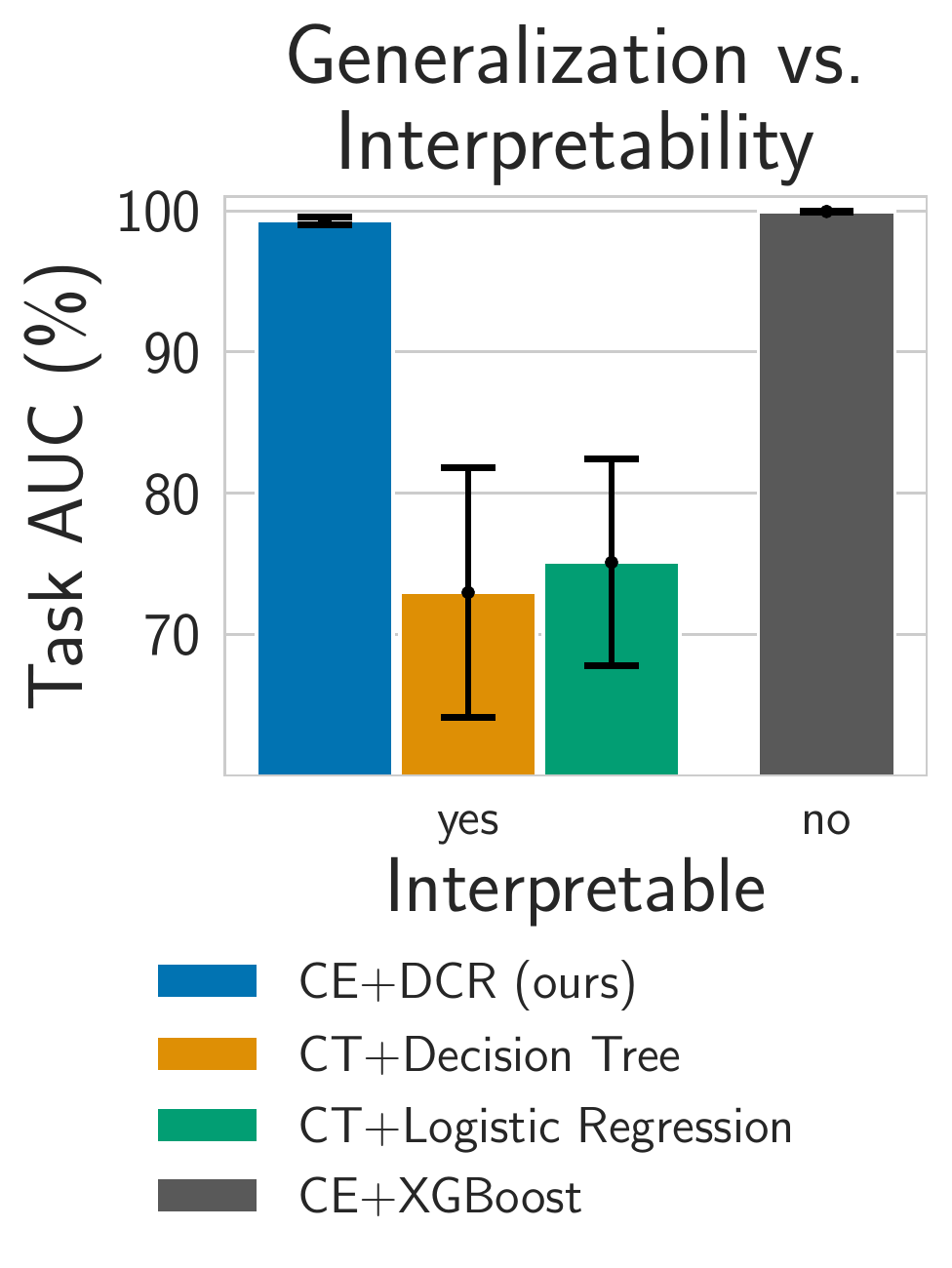}
    \caption{(a) An interpretable concept-based model $f$ maps concepts $\hat{C}$ to tasks $\hat{Y}$ generating an interpretable rule. When input features are not semantically meaningful, a concept encoder $g$ can map raw features to a concept space. (b) The proposed approach (DCR) outperforms interpretable concept-based models in the \emph{Dot} dataset. \emph{CE} stands for concept embeddings and \emph{CT} for concept truth values.}
    \label{fig:pre_results}
\end{figure*}

Concept-based models~\cite{kim2018interpretability, chen2020concept} aim to increase human trust in deep learning models by using human-understandable concepts to train interpretable models---such as logistic regression or decision trees~\cite{rudin2019stop,koh2020concept,kazhdan2020now} (Figure~\ref{fig:pre_results}).
This approach significantly increases human trust in the AI predictor~\cite{rudin2019stop,shen2022trust} as it allows users to clearly understand a model's decision process. 
However, state-of-the-art concept-based models, which rely on concept embeddings~\cite{yeh2020completeness,kazhdan2020now,mahinpei2021promises,zarlenga2022concept} to attain high performance, are not completely interpretable. Indeed, concept embeddings lack clear semantics on individual dimensions, e.g., $\mathbf{\hat{c}}_\text{yellow} = [2.3, 0.3, -3.5, \dots]^T$ does not have semantics assigned to each of its dimensions. This sacrifice of interpretability in favour of model capacity leads to a possible reduction in human trust when using these models, as argued by~\citet{rudin2019stop,mahinpei2021promises}.


In this paper, we propose the \textit{Deep Concept Reasoner}\footnote{Code available in public repository: \url{https://github.com/pietrobarbiero/pytorch_explain}.} (DCR, Section~\ref{sec:DCR}), the first interpretable concept-based model building on concept embeddings. DCR applies differentiable and learnable modules on concept embeddings to build a set of fuzzy rules which can then be executed on semantically meaningful concept truth degrees to provide a final interpretable prediction.
Our experiments (Section~\ref{sec:exp}) show that DCR: (i) attains better task accuracy than state-of-the-art interpretable concept-based models (Figure~\ref{fig:pre_results}), (ii) discovers meaningful logic rules, matching known ground truths even in absence of training concept supervision, and (iii) facilitates the generation of counterfactual examples thanks to the highly-interpretable learnt rules.
\section{Preliminaries}
\label{sec:back}
\paragraph{Concept-based models}
Concept-based models $f: C \rightarrow Y$ learn a map from a concept space $C$ to a task space $Y$~\cite{yeh2020completeness}. If concepts are semantically meaningful, then humans can interpret this mapping by tracing back predictions to the most relevant concepts~\cite{ghorbani2019interpretation}. When the features of the input space are hard for humans to reason about (such as pixel intensities), concept-based models work on the output of a concept-encoder mapping $g: X \rightarrow C$ from the input space $X$ to the concept space $C$~\cite{ghorbani2019towards,koh2020concept}. 
In general, training a concept-based model may require a dataset 
where each sample consists of input features $\mathbf{x}\in \mathcal{X} \subseteq \mathbb{R}^n$ (e.g., an image's pixels), $k$ ground truth concepts $\mathbf{c}\in  \mathcal{C} \subseteq \{0, 1\}^k$ (i.e., a binary vector with concept annotations, when available) and $o$ task labels $\mathbf{y} \in  \mathcal{Y} \subseteq \{0, 1\}^o$ (e.g., an image's classes). 
During training, a concept-based model is encouraged to align its predictions to task labels i.e., $\mathbf{y} \approx \mathbf{\hat{y}}=f(g(\mathbf{x}))$. Similarly, a concept encoder can be supervised when concept labels are available i.e., $\mathbf{c} \approx \mathbf{\hat{c}} = g(\mathbf{x})$. 
When concept labels are not available, they can still be extracted from pre-trained models associating concept labels to clusters found in their embeddings as proposed by~\citet{ghorbani2019towards,magister2021gcexplainer}. We indicate concept and task predictions as $\hat{c}_i=(g(\mathbf{x}))_i$ and $\hat{y}_j=(f(\mathbf{\hat{c}}))_j$ respectively.

\paragraph{Concept truth values vs. concept embeddings}
Usually, concept-based models represent concepts using their truth degree, that is, $\hat{c}_1,\ldots,\hat{c}_k\in [0,1]$.
However, this representation might significantly degrade task accuracy as observed by~\citet{mahinpei2021promises} and~\citet{zarlenga2022concept}. To overcome this issue, concept-based models may represent concepts using concept embeddings $\mathbf{\hat{c}}_i \in \mathbb{R}^m$ alongside their truth degrees $\hat{c}_i \in [0,1]$.\footnote{
With an abuse of notation, we use the same symbol for a concept embedding and its corresponding truth degree, with the former in bold to distinguish it.} While this increases task accuracy of concept-based models~\cite{zarlenga2022concept}, it also weakens their interpretability as concept embeddings lack clear semantics.

\paragraph{Fuzzy logic rules}
Continuous fuzzy logics~\cite{hajek2013metamathematics} extend Boolean logic by relaxing discrete truth-values in $\{0,1\}$ to truth degrees in $[0,1]$, and Boolean connectives to (differentiable) real-valued operators. In particular, a \mbox{t-norm
$\wedge:[0,1]\times[0,1]\rightarrow[0,1]$} generalises the Boolean conjunction
while a \mbox{t-conorm $\vee:[0,1]\times[0,1]\rightarrow[0,1]$} generalises the disjunction. These two operators are connected by the strong negation $\neg$, defined as $\neg x=1-x$.
For example, the product (fuzzy) logic can be defined by the operators $x\wedge y := x\cdot y$ and $x\vee y := x+y-xy$. 
As in Boolean logic, the syntax of a t-norm fuzzy rule includes: \mbox{(i) Atomic} formulas consisting of \textit{propositional variables} $z$, and logical \textit{constants} $\bot$ (\texttt{false}, ``0'') and $\top$ (\texttt{true}, ``1''),
\mbox{(ii) \textit{Literals}} representing atomic formulas or their negation, and \mbox{(iii) Logical} \textit{connectives} $\neg,\wedge,\vee,\Rightarrow,\Leftrightarrow$ joining formulas in arbitrarily complex compound formulas.


\section{Deep Concept Reasoning}
\label{sec:DCR}
\begin{figure*}[!t]
    \centering
    \includegraphics[clip, trim=0.4cm 0cm 7cm 0cm, width=0.2\textwidth]{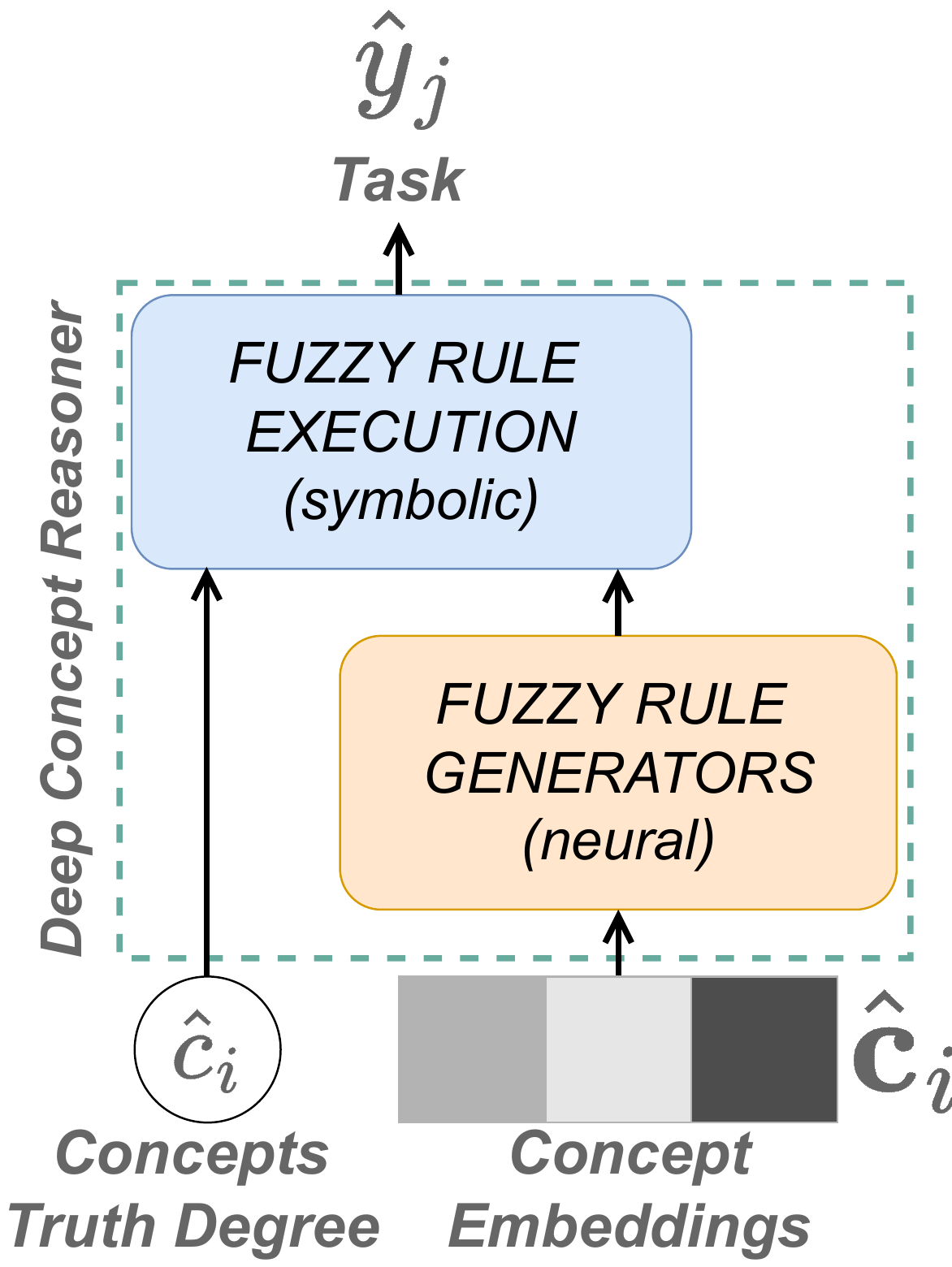}
    \qquad
    \includegraphics[clip, trim=0cm 0cm 2.9cm 0cm, width=0.6\textwidth]{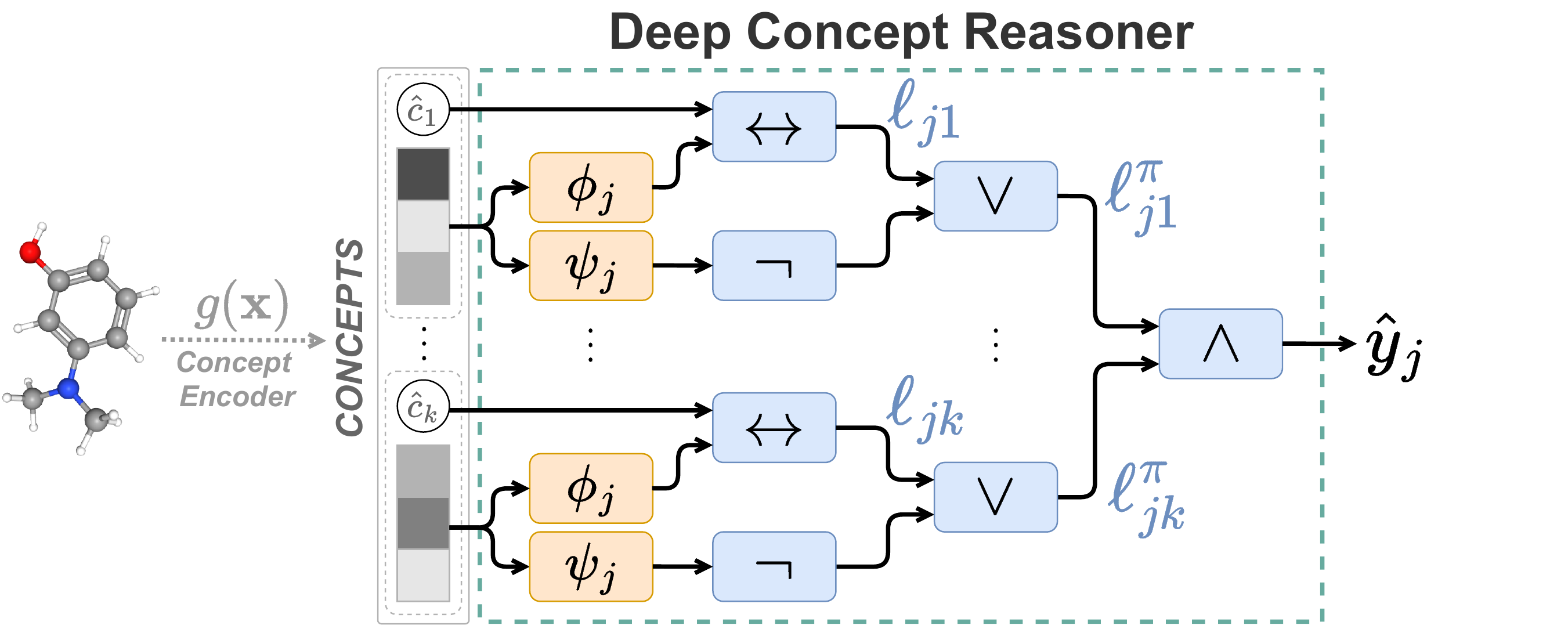}
    \caption{(left) Deep Concept Reasoner (DCR) generates fuzzy logic rules using neural models on concept embeddings. Then DCR executes the rule using the concept truth degrees to evaluate the rule symbolically. (right) Schema of DCR modules: first neural models $\phi$ and $\psi$ generate the rule, and then the rule is executed symbolically.}
    \label{fig:dcr-method}
\end{figure*}

Here we describe the ``Deep Concept Reasoner'' (DCR, Figure~\ref{fig:dcr-method}), the first interpretable concept-based model based on concept embeddings.  Similarly to existing models based on concept embeddings, DCR exploits high-dimensional representations of the concepts. However, in DCR, such representations are only used to compute a logic rule. The final prediction is then obtained by evaluating such rules on the concepts' truth values and not on their embeddings, thus maintaining clear semantics and providing a totally interpretable decision. 
Being differentiable, DCR is trainable as an independent module on concept databases, but it can also be trained end-to-end with differentiable concept encoders.
In the following section, we describe \mbox{(1) the} syntax of the rules we aim to learn (Section~\ref{sec:rulepred}), \mbox{(2) how} to (neurally) generate and execute learnt rules to predict task labels (Section~\ref{sec:ruleexec}), (\mbox{3) how} DCR learns simple rules in specific t-norm semantics (Section~\ref{sec:ruleexec}), and \mbox{(4) how} we can generate global and counterfactual explanations with DCR (Section~\ref{sec:ruleadv}). We provide Figure~\ref{fig:dcr-method} as a reference to graphically follow the discussion.

\subsection{Rule syntax}
\label{sec:rulepred}
To understand the rationale behind DCR's design, we begin with an illustrative toy example:
\begin{example}
\label{ex:banana}
Consider the problem of defining the fruit ``banana'' given the vocabulary of concepts ``soft'', ``round'', and ``yellow''. A simple definition can be $y_{\textit{banana}} \Leftrightarrow \neg c_{\textit{round}} \land c_{\textit{yellow}}$. From this rule we can deduce that (i) being ``soft'' is irrelevant for being a ``banana'' (indeed bananas can be both soft or hard), and (ii) being both ``\underline{not} round" and ``yellow'' is relevant to being a ``banana''.
\end{example}
As in this example, DCR rules can express whether a concept is \textbf{\emph{relevant}} or not (e.g., ``soft''), and whether a concept plays a positive (e.g., ``yellow'') or negative (e.g., ``\underline{not} round'') \textbf{\emph{role}}. 
To formalize this description of rule syntax, we let $l_{ji}$ denote the literal of concept $c_i$ (i.e., $\hat{c}_i$ or $\neg \hat{c}_i$) representing the \emph{role} of the concept $i$ for the $j$-th class. Similarly, we let $r_{ji} \in \{0,1\}$ representing whether $\hat{c}_i$ is \emph{relevant} for predicting the class $y_j$.
For each sample $\mathbf{x}$ and predicted class $\hat{y}_j$, DCR learns a rule with the following syntax\footnote{Here and in all equations we omit the explicit dependence on $\mathbf{x}$ for simplicity, i.e., we write $\hat{y}_j$ for $\hat{y}_j(\mathbf{x})$.}:
\begin{equation} \label{eq:target-rule}
    \hat{y}_j \Leftrightarrow \bigwedge_{i:\; r_{j i} = 1}{l_{ji}}
\end{equation}
Such a rule defines a logical statement for why a given sample is predicted to have label $\hat{y}_j$ using a conjunction of relevant concept literals (i.e., $\hat{c}_i$ or $\neg \hat{c}_i$).

\subsection{Rule generation and execution}
\label{sec:ruleexec}
Having defined the syntax of DCR rules, we describe how to \textit{generate} and \textit{execute} these rules in a differentiable way. To generate a rule we use two neural modules $\phi_j$ and $\psi_j$ which determine the role and relevance of each concept, respectively. Then, we execute each rule using the concepts' truth degrees of a given sample.
We split this process into three steps: (i) learning each concept's roles, (ii) learning each concept's relevance, and (iii) predicting the task using the relevant concepts.



\paragraph{Concept role}
\underline{Generation:} To determine the \emph{role} (positive/negative) of a concept, we use a feed-forward neural network $\phi_j: \mathbb{R}^m \rightarrow [0,1]$, with $m$ being the dimension of each concept embedding.
The neural model $\phi_j$ takes as input a concept embedding $\hat{\mathbf{c}}_i \in \mathbb{R}^m$ and returns a soft indicator representing the role of the concept in the formula, that is, whether in literal $l_{ji}$ the concept should appear negated (e.g., $\phi_{\textit{banana}}(\hat{\mathbf{c}}_{\textit{round}}) = 0$)  or not (e.g., $\phi_{\textit{banana}}(\hat{\mathbf{c}}_{\textit{yellow}}) = 1$).
\underline{Execution:} When we execute the rule, we need to compute the actual truth degree of a literal $l_{ji}$ given its role $\phi(\hat{\mathbf{c}}_i)$. We define this truth degree  $\ell_{ji} \in [0,1]$. In particular, we want to (i) forward the same truth degree of the concept, i.e.  $\ell_{ji} = \hat{c}_i$,  when $\phi(\hat{\mathbf{c}}_i)=1$,  and (ii) negate it, i.e. $\ell_{ji} = \neg \hat{c}_i$,  when $\phi(\hat{\mathbf{c}}_i)=0$.
This behaviour can be generalized by a fuzzy equality $\Leftrightarrow $ when both $\phi_j$ and $\hat{c}$ are fuzzy values, i.e.: 
\begin{equation}
    \label{eq:iff}
     \ell_{ji} = (\phi_j(\hat{\mathbf{c}}_i) \Leftrightarrow \hat{c}_{i})
\end{equation}
\begin{example}
    For a given object consider $\hat{c}_{\textit{round}}=0$ and $\phi_{\textit{banana}}(\hat{\mathbf{c}}_{\textit{round}})=0$. Then we get $\ell_{\textit{banana},\textit{round}}=(\phi_{\textit{banana}}(\hat{\mathbf{c}}_{\textit{round}}) \Leftrightarrow \hat{c}_{\textit{round}})=\neg \hat{c}_{\textit{round}}=1$. If instead we had $\phi_{\textit{banana}}(\hat{\mathbf{c}}_{\textit{round}}) = 1$, then $\ell_{\textit{banana},\textit{round}}=(\phi_{\textit{banana}}(\hat{\mathbf{c}}_{\textit{round}}) \Leftrightarrow \hat{c}_{\textit{round}})=0$.
\end{example}
\paragraph{Concept relevance.}
\underline{Generation:} To determine the \emph{relevance} of a concept $\mathbf{\hat{c}}_i$, we use another feed-forward neural network $\psi_j: \mathbb{R}^m \rightarrow [0,1]$. The model $\psi_j$ takes as input a concept embedding $\hat{\mathbf{c}}_i \in \mathbb{R}^m$ and returns a soft indicator representing the likelihood of a concept being relevant for the formula (e.g., $\psi_{\textit{banana}}(\hat{\mathbf{c}}_{\textit{soft}}) = 1$) or not (e.g., $\psi_{\textit{banana}}(\hat{\mathbf{c}}_{\textit{yellow}}) = 0$).
\underline{Execution:} When we execute the rule, we need to compute the truth degree of a literal given its relevance $r_{ji}$. We define the truth degree of a relevant literal as $\ell^r_{ji} \in [0,1]$, where $r$ stands for ``relevant''. In particular, we want to  \mbox{(i) filter} irrelevant concepts when $\psi_j(\hat{\mathbf{c}}_i) = 0$ by setting $\ell^r_{ji}=1$, and  \mbox{(ii) retain} relevant literals when $\psi_j(\hat{\mathbf{c}}_i) = 1$ by setting $\ell^r_{ji}= \ell_{ji}$. This behaviour can be generalized to fuzzy values of $\psi_j$ as follows: 
\begin{equation}~\label{eq:filter}
    \ell^r_{ji} = (\psi_j(\hat{\mathbf{c}}_i)\Rightarrow \ell_{ji})= (\neg \psi_j(\hat{\mathbf{c}}_i) \vee \ell_{ji})
\end{equation}

Note that setting $\ell^r_{ji}=1$ makes the literal $l_{ji}$ irrelevant since ``$1$'' is neutral w.r.t.\ the conjunction in Equation~\ref{eq:ruletarget}. 


\begin{example}
    For a given object of type ``banana'', let the concept ``soft'' be irrelevant, that is $\psi_{\textit{banana}}(\hat{\mathbf{c}}_{\textit{soft}}) = 0$. Then we get $\ell^r_{\textit{banana},\textit{soft}}=(\psi_{\textit{banana}}(\hat{\mathbf{c}}_{\textit{soft}}) \Rightarrow \ell_{\textit{banana},\textit{soft}})=1$, independently from the content of $\hat{c}_{\textit{soft}}$ or $\ell_{\textit{banana},\textit{soft}}$. Conversely, let the concept ``yellow'' by relevant, that is $\psi_{\textit{banana}}(\hat{\mathbf{c}}_{\textit{yellow}}) = 1$, and let its concept literal be $\ell_{\textit{banana},\textit{yellow}}=\hat{\mathbf{c}}_{\textit{yellow}}=1$. As a result, we get
    $\ell^r_{\textit{banana},\textit{yellow}}=(\psi_{\textit{banana}}(\hat{\mathbf{c}}_{\textit{yellow}}) \Rightarrow \ell_{\textit{banana},\textit{yellow}})=1$.
\end{example}


\paragraph{Task prediction}
Finally, we conjoin the relevant literals $\ell^r_{ji}$ to obtain the task prediction $\hat{y}_j$:
\begin{equation}
\label{eq:ruletarget}
    \hat{y}_j = \bigwedge_{i=1}^k \ell^r_{ji} 
\end{equation}
\begin{example}
    For a given object of type ``banana'', consider the following truth degrees for the concepts:  $\hat{c}_{soft}= 1, \hat{c}_{round} = 0,\hat{c}_{yellow} = 1$. Consider also the following values for the role and relevance of the class ``banana'': $\phi_{\textit{banana}}(\hat{\mathbf{c}}_{i})=[0,0,1]$ and $\psi_{\textit{banana}}(\hat{\mathbf{c}}_{i})=[0,1,1]$ for $i \in \{\textit{soft}, \textit{round}, \textit{yellow}\}$. Then, we obtain the final prediction for class $banana$ as: 
    \[
    \begin{array}{r}
    \hat{y}_{\textit{banana}} = \bigwedge_{i=1}^3\left(\neg \psi_{\textit{banana}}(\hat{\mathbf{c}}_i) \vee (\phi_{\textit{banana}}(\hat{\mathbf{c}}_i) \Leftrightarrow \hat{c}_{i})\right) =    \\
    =(1\vee(0\Leftrightarrow 1))\wedge (0\vee(0\Leftrightarrow 0))\wedge (0\vee(1\Leftrightarrow 1))=
    \\
    =(1\vee 0)\wedge (0\vee 1)\wedge (0\vee 1)=1 \wedge 1 \wedge 1 = 1
    \end{array}
    \]
\end{example}
We remark that the models $\phi_j$ and $\psi_j$: (a) generate fuzzy logic rules using concept embeddings which might hold more information than just concept truth degrees, and (b) do not depend on the number of input concepts which makes them applicable---without retraining---in testing environments where the set of concepts available differs from the set of concepts used during training.
We also remark that the whole process is differentiable as the neural models $\phi_j$ and $\psi_j$ are differentiable as well as the fuzzy logic operations as we will see in the next section.

\subsection{Rule parsimony and fuzzy semantics}
\paragraph{Rule parsimony} 
Simple explanations and logic rules are easier to interpret for humans~\cite{miller1956magical,rudin2019stop}. We can encode this behaviour within the DCR architecture by enforcing a certain degree of competition among concepts to make only relevant concepts survive. To this end, we design a special activation function for the neural network $\psi_j$ rescaling the output of a log-softmax activation:
\begin{align}
    \gamma_{ji} &= \log \Bigg( \frac{\exp(\text{MLP}_j(\hat{\mathbf{c}}_i))}{\sum_{i^\prime=1}^k \exp(\text{MLP}_j(\hat{\mathbf{c}}_{i^\prime}))} \Bigg) 
    \label{eq:comp}\\
    r_{ji} &= \psi_j(\hat{\mathbf{c}}_i) = \sigma \Bigg(\gamma_{ji} - \frac{1}{k} \sum_{i^\prime=1}^k \gamma_{ji^\prime} \Bigg)\label{eq:rel}
\end{align}
This way, if the scores $\gamma_{ji}$ are uniformly distributed, then we expect the network $\psi_j$ to select half of the concepts. 
We can also parametrise this function by introducing a parameter $\tau \in [-\infty, \infty]$ that allows a user to bias the default behaviour of the activation function: 
$r_{ji} = \sigma (\gamma_{ji} - \frac{\tau}{k} \sum_{i^\prime=1}^k \gamma_{ji^\prime} )$.
A user can increase $\tau$ to get more relevance scores closer to $1$ (more complex rules) or decrease it to get more relevance scores closer to $0$ (simpler rules).

\paragraph{Fuzzy semantics}
\label{sec:semirings}
To create a semantically valid model, we enforce the same semantic structure in all logic and neural operations. 
Moreover, to train our model end-to-end, we need these semantics to be differentiable in all its operations, including logic functions. \citet{marra2020lyrics} describe a set of possible t-norm fuzzy logics which can serve the purpose. In our experiments, we use the G\"odel t-norm. With this semantics, we can rewrite Equation~\ref{eq:iff} as:
\[
\begin{array}{ll}
    \ell_{ji} & =\phi_j(\hat{\mathbf{c}}_i)\Leftrightarrow \hat{c}_i =  (\phi_j(\hat{\mathbf{c}}_i)\Rightarrow \hat{c}_i)\wedge (\hat{c}_i\Rightarrow \phi_j(\hat{\mathbf{c}}_i)) = \nonumber\\   
    &=(\neg\phi_j(\hat{\mathbf{c}}_i)\vee \hat{c}_i)\wedge (\neg \hat{c}_i\vee \phi_j(\hat{\mathbf{c}}_i)) = \nonumber\\   
    &=\min\{\max\{1-\phi_j(\hat{\mathbf{c}}_i), \hat{c}_{i}\}, \max\{1-\hat{c}_{i},\phi(\hat{\mathbf{c}}_i)\}\}
\end{array}
\]
and Equation~\ref{eq:ruletarget} as: $
    \hat{y}_j = \min_{i=1}^k\{\max\{1 - \psi_j(\hat{\mathbf{c}}_i), \ell_{ji}\}\}
$

\subsection{Global and counterfactual explanations}
\label{sec:ruleadv}

\paragraph{Interpreting global behaviour}
In general, DCR rules may have different weights and concepts for different samples. However, we can still globally interpret the predictions of our model without the need for an external post-hoc explainer. To this end, we collect a batch of (or all) fuzzy rules generated DCR on the training data $\mathcal{X}_{\text{train}}$. Following~\citet{barbiero2022entropy}, we then Booleanize the collected rules and aggregate them with a global disjunction to get a single logic formula valid for all samples of class $j$:
\begin{equation} \label{eq:global-explanation}
    \hat{y}^C_j = \bigvee_{\mathbf{x} \in \mathcal{X}_{\text{train}}} \hat{y}_j(\mathbf{x})
\end{equation}
This way we obtain a global overview of the decision process of our model for each class.


\paragraph{Counterfactual explanations}
Logic rules clearly reveal which concepts play a key role in a prediction. This transparency, typical of interpretable models, facilitates the extraction of simple counterfactual explanations without the need for an external algorithm as in~\citet{abid2021meaningfully}. 
In DCR we extract simple counter-examples $x^\star$ using the logic rule as guidance. Following~\citet{wachter2017counterfactual}, we generate counter-examples as close as possible to the original sample $|x - x^\star|< \epsilon$. In particular, \citet{wachter2017counterfactual} proposes to perturb the input features of a model starting from the most relevant features. As the decision process depends mostly on the most relevant features, perturbing a small set of features is usually enough to find counter-examples. To this end, we first rank the concepts present in the rule according to their relevance scores. Then, starting from the most relevant concept, we invert their truth value until the prediction of the model changes. The new rule represents a counterfactual explanation for the original prediction.




\section{Experiments}
\label{sec:exp}

\subsection{Research questions}
In this section, we analyze the following research questions:
\begin{itemize}
    \item \textbf{Generalization ---} How does DCR generalize on unseen samples compared to interpretable and neural-symbolic models? How does DCR generalize when concepts are unsupervised?
    \item \textbf{Interpretability ---} Can DCR discover meaningful rules? Can DCR re-discover ground-truth rules? How stable are DCR rules under small perturbations of the input compared to interpretable models and local post-hoc explainers?
    How long does it take to extract a counterfactual explanation from DCR compared to a non-interpretable model?
\end{itemize}

\begin{figure*}[t]
    \centering
    \includegraphics[width=0.98\linewidth]{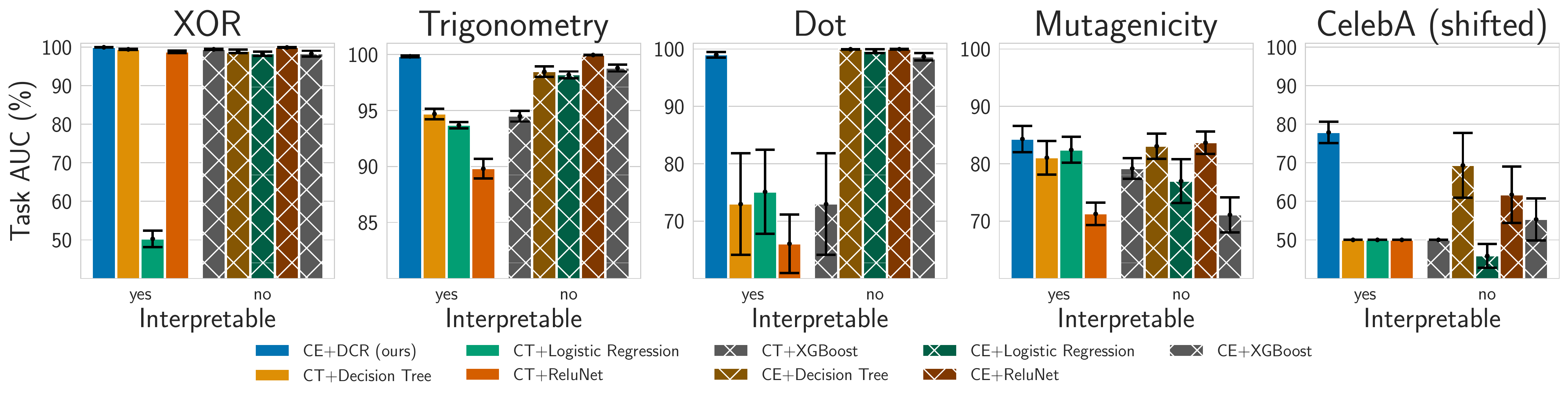}
    \caption{Mean ROC AUC for task predictions for all baselines across all tasks (the higher the better). DCR often outperforms interpretable concept-based models. \emph{CE} stands for concept embeddings, while \emph{CT} for concept truth degrees. Models trained on concept embeddings are not interpretable as concept embeddings lack a clear semantic for individual embedding dimensions.
    }
    \label{fig:accuracy}
\end{figure*}

\subsection{Experimental setup}
\paragraph{Data \& task setup}
We investigate our research questions using six datasets spanning three of the most common data types used in deep learning: tabular, image, and graph-structured data.
We use the three benchmark datasets (\textit{XOR}, \textit{Trigonometry}, and \textit{Dot}) proposed by~\citet{zarlenga2022concept} as they capture increasingly complex concept-to-label relationships, therefore challenging concept-based models. To test the DCR's ability to re-discover ground-truth rules we use the \textit{MNIST-Addition} dataset~\cite{manhaeve2018deepproblog}, a standard benchmark for neural-symbolic systems where one aims to predict the sum of two digits from the MNIST's dataset. 
Furthermore, we evaluate our methods on two real-world benchmark datasets: the Large-scale CelebFaces Attributes (\emph{CelebA},~\citep{liu2015deep}) and the \emph{Mutagenicity}~\cite{morris2020tudataset} dataset. In particular, we define a new \emph{CelebA} task to simulate a real-world condition of concept ``shifts'' where train and test concepts are correlated (e.g., ``beard'' and ``mustaches'') but do not match exactly. To this end, we split the set of \emph{CelebA} attributes defined by~\citet{zarlenga2022concept} in two partially disjoint sets and use one set of attributes for training models and one for testing.
Finally, we use \emph{Mutagenicity} as a real-world scenario the concept encoder is unsupervised.
As \emph{Mutagenicity} does not have concept annotations, we first train a graph neural network (GNN) on this dataset, and then we use the Graph Concept Explainer (GCExplainer, ~\cite{magister2021gcexplainer}) to extract a set of concepts from the embeddings of the trained GNN.
For dataset with concept labels instead, we generate concept embeddings and truth degrees by training a Concept Embedding Model~\cite{zarlenga2022concept}.

\paragraph{Baselines}
We compare DCR against interpretable models, such as logistic regression~\cite{verhulst1845resherches}, decision trees~\citep{breiman2017classification}, as well as state-of-the-art black-box classifiers, such as extreme gradient boosting (\mbox{XGBoost})~\citep{chen2016xgboost}, and locally-interpretable neural models, such as the Relu Net \citep{ciravegna2023logic}. 
We train all baseline models in two different conditions mapping concepts to tasks either using concept truth degrees or using concept embeddings (baselines marked with \emph{CT} and \emph{CE} in figures, respectively). We consider interpretable only baselines trained on concept truth degrees only, as concept embeddings lack of clear semantics assigned to each dimension. However, baselines trained on concept embeddings still provide a strong reference for task accuracy w.r.t. interpretable models.
On the \emph{MNIST-Addition} dataset we compare DCR with state-of-the-art neural-symbolic baselines including: DeepProbLog \cite{manhaeve2018deepproblog}, DeepStochLog \cite{winters2022deepstochlog}, Logic Tensor Networks \cite{badreddine2022logic}, and Embed2Sym \cite{aspis2022embed2sym}. This is possible as the \emph{MNIST-Addition} dataset provides access to the full set of ground-truth rules, allowing us to train these neural-symbolic systems.
Finally, we compare DCR interpretability with interpretable models, such as logistic regression and decision trees, and with local post-hoc explainers, such as the Local Interpretable Model-agnostic Explanations (LIME,~\cite{ribeiro2016should}) applied on XGBoost. 

\paragraph{Evaluation}
We assess each model's performance and interpretability based on four criteria. First, we measure task generalization using the Area Under the Receiver Operating Characteristic Curve (ROC AUC) from prediction scores~\cite{hand2001simple} (the higher the better).
Second, we evaluate DCR interpretability by comparing the learnt logic formulae with ground-truth rules in \emph{XOR}, \emph{Trigonometry}, and \emph{MNIST-Addition} datasets, and indirectly on \emph{Mutagenicity} by checking whether the learnt rules involve concepts corresponding to functional groups known for their harmful effects, as done by~\citet{ying2019gnnexplainer}. Third, to further assess interpretability, we measure 
the sensitivity of the predictions under small perturbations following~\citet{yeh2019fidelity} (the lower the better). Finally, we measure how receptive our model is to extract meaningful counterfactual examples from its rules by computing the number of concept perturbations required to obtain a counterfactual example following~\citet{wachter2017counterfactual} (the lower the better).
For each metric, 
we report their mean and 95\% Confidence Intervals (CI) on our test sets using $5$ different initialization seeds. We report further details and results in the appendix.

\subsection{Task generalization}
\paragraph{DCR outperforms interpretable models (Figure~\ref{fig:accuracy})} 
Our experiments show that DCR generalizes significantly better than interpretable benchmarks in our most challenging datasets. This improvement peaks when concept embeddings hold more information than concept truth degrees, as in  the \emph{CelebA} and \emph{Dot} tasks where this deficit of information is imposed byconstruction~\cite{zarlenga2022concept}. This grants DCR a significant advantage (up to $\sim 25\%$ improvement in ROC-AUC) over the other interpretable baselines. This phenomenon confirms the findings by~\citet{mahinpei2021promises} and~\citet{zarlenga2023towards}. In particular, the concept shift in \emph{CelebA} causes interpretable models to behave almost randomly as the set of test concepts is different from the set of train concepts (despite being correlated). DCR however still generalizes well as the mechanism generating rules only depends on concept embeddings and the embeddings hold more information on the correlation between train and test concepts w.r.t. concept truth degrees.
To further test this hypothesis, we compare DCR against XGBoost, decision trees (DTs), and logistic regression trained on concept embeddings. In most cases, concept embeddings allow DTs and logistic regression to improve task generalization, but the predictions of such models are no longer interpretable. In fact, even a logic rule whose terms correspond to dimensions of a concept embedding is not semantically meaningful as discussed in Section~\ref{sec:back}. In contrast, DCR uses concept embeddings to assemble rules whose terms are concept truth degrees, which makes it possible to keep the rules semantically meaningful.

\paragraph{DCR matches the accuracy of neural-symbolic systems trained using human rules (Table~\ref{tab:mnist-addition-accuracy})}
Our experiments show that DCR generates rules that, when applied, obtain accuracy levels close to neural-symbolic systems trained using human rules, currently representing the gold standard to benchmark rule learners.
We show this result on the \emph{MNIST-Addition} dataset~\cite{manhaeve2018deepproblog}, a standard benchmark in neural-symbolic AI, where the labels on the concepts are not available. We learn concepts without supervision by adding another task classifier, which only uses very crisp $\hat{c}_i$ to make the task predictions (see Appendix \ref{app:mnist}). 
DCR achieves similar performance to state-of-the-art neural-symbolic baselines (within $1\%$ accuracy from the best baseline). However, DCR is the only system discovering logic rules directly from data, while all the other baselines are trained using ground-truth rules. Therefore, this experiment indicates how DCR can learn meaningful rules also without concept supervision while still maintaining state-of-the-art performance.
\begin{table}[]
\centering
\caption{Task accuracy on the \emph{MNIST-addition} dataset. The neural-symbolic baselines use the knowledge of the symbolic task 
to distantly supervise the image recognition task. DCR achieves similar performances even though it learns the rules from scratch.}
\label{tab:mnist-addition-accuracy}
\resizebox{0.6\columnwidth}{!}{%
\begin{tabular}{ll}
\hline
\textbf{\textbf{\textsc{Model}}} & \textbf{\textbf{\textsc{Accuracy} (\%)}} \\ \hline
\multicolumn{2}{c}{With ground truth rules} \\
DeepProbLog & $97.2 \pm 0.5$ \\
DeepStochLog & $97.9 \pm 0.1$ \\
Embed2Sym & $97.7 \pm 0.1$ \\
LTN & $98.0 \pm 0.1$ \\ \hline
\multicolumn{2}{c}{Without ground truth rules} \\
DCR(ours) & $97.4 \pm 0.2$ \\ \hline
\end{tabular}%
}
\end{table}
\subsection{Interpretability}
\paragraph{DCR discovers semantically meaningful logic rules (Table~\ref{tab:global-rules})}
Our experiments show that DCR induces logic rules that are both accurate in predicting the task and formally correct when compared to ground-truth logic rules. We evaluate the formal correctness of DCR rules on the \emph{XOR}, \emph{Trigonometry}, and \emph{MNIST-Addition} datasets where we have access to ground-truth logic rules. We report a selection of Booleanized DCR rules with the corresponding ground truth rules in Table~\ref{tab:global-rules}. 
Our results indicate that DCR's rules align with human-designed ground truth rules, making them highly interpretable. For instance, DCR predicts that the sum of two MNIST digits is $17$ if either the first image is a \includegraphics[scale=0.4]{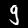} (i.e., $c'_9$) and the second is an \includegraphics[scale=0.4]{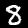} (i.e., $c''_8$) or vice-versa which we can interpret globally using Equation~\ref{eq:global-explanation} as: $y_{17} \Leftrightarrow (c'_9 \land  c''_8) \vee (c'_8 \land  c''_9)$. We list all logic rules discovered by DCR on the \emph{MNIST-Addition} dataset in Appendix~\ref{app:mnist}.
It is interesting to investigate the potential of DCR also in settings where we do not have access to the ground-truth logic rules, such as the \emph{Mutagenicity} dataset. Here, unlike the \textit{MNIST addition} dataset, not only
there is no supervision on the concepts, but we don't even know which are the concepts.
We use GCExplainer~\cite{magister2021gcexplainer} to generate a set of concept embeddings from the embeddings of a trained GNN. We then use these embeddings to train DCR.
In this setting, we can only evaluate the correctness of a DCR rules indirectly by checking whether the concepts appearing in the rules correspond to functional groups known for their harmful effects within the \emph{Mutagenicity} dataset following~\citet{ying2019gnnexplainer}. Interestingly, many of DCR's rules predicting mutagenic effects include functional groups such as phenols~\cite{hattenschwiler2000role} and dimethylamines~\cite{acgih2016american}, which can be highly toxic when combined in molecules such as \mbox{3-Dimethylaminophenols}~\cite{sabry2011synthesis}. This suggests that DCR has the potential to unveil semantically meaningful relations among concepts and to make them explicit to humans by means of the learnt rules. 
We provide experimental details with the full list of concepts and rules discovered in \emph{Mutagenicity} in Appendix~\ref{app:mutag}.

\begin{table}[!t]
\centering
\caption{Error rate of Booleanised DCR rules w.r.t.\ ground truth rules. Error rate represents how often the label predicted by a Booleanised rule differs from the fuzzy rule generated by our model. The error rate is reported with the mean and standard error of the mean. A full list of logic rules for MNIST is in Appendix~\ref{app:mnist}.}
\label{tab:global-rules}
\resizebox{\columnwidth}{!}{%
\begin{tabular}{lll}
\hline
\multicolumn{1}{l}{\textbf{\textsc{Ground-truth Rule}}} & \multicolumn{1}{l}{\textbf{\textsc{Predicted Rule}}} & \multicolumn{1}{l}{\textbf{\textsc{Error (\%)}}} \\ 
\hline
\multicolumn{3}{c}{\textbf{XOR}} \\
$y_0 \leftarrow \neg c_0 \wedge \neg c_1$ & $y_0 \leftarrow \neg c_0 \wedge \neg c_1$ & $0.00 \pm 0.00$ \\
$y_0 \leftarrow c_0 \wedge c_1$ & $y_0 \leftarrow c_0 \wedge c_1$ & $0.00 \pm 0.00$ \\
$y_1 \leftarrow \neg c_0 \wedge c_1$ & $y_1 \leftarrow \neg c_0 \wedge c_1$ & $0.02 \pm 0.02$ \\
$y_1 \leftarrow c_0 \wedge \neg c_1$ & $y_1 \leftarrow c_0 \wedge \neg c_1$ & $0.01 \pm 0.01$ \\
\multicolumn{3}{c}{\textbf{Trigonometry}} \\
$y_0 \leftarrow \neg c_0 \wedge \neg c_1 \wedge \neg c_2$ & $y_0 \leftarrow \neg c_0 \wedge \neg c_1 \wedge \neg c_2$ & $0.00 \pm 0.00$ \\
$y_1 \leftarrow c_0 \wedge c_1 \wedge c_2$ & $y_1 \leftarrow c_0 \wedge c_1 \wedge c_2$ & $0.00 \pm 0.00$ \\ 
\multicolumn{3}{c}{\textbf{MNIST-Addition}} \\
$y_{18} \leftarrow c'_9 \land  c''_9$ & $y_{18} \leftarrow c'_9 \land  c''_9$ & $0.00 \pm 0.00$ \\
$y_{17} \leftarrow c'_9 \land  c''_8$ & $y_{17} \leftarrow c'_9 \land  c''_8$ & $0.00 \pm 0.00$ \\
$y_{17} \leftarrow c'_8 \land  c''_9$ & $y_{17} \leftarrow c'_8 \land  c''_9$ & $0.00 \pm 0.00$ \\
\hline
\end{tabular}%
}
\end{table}

\begin{figure}[t]
    \centering
    \includegraphics[trim=0 0 0 0, clip, width=0.9\columnwidth]{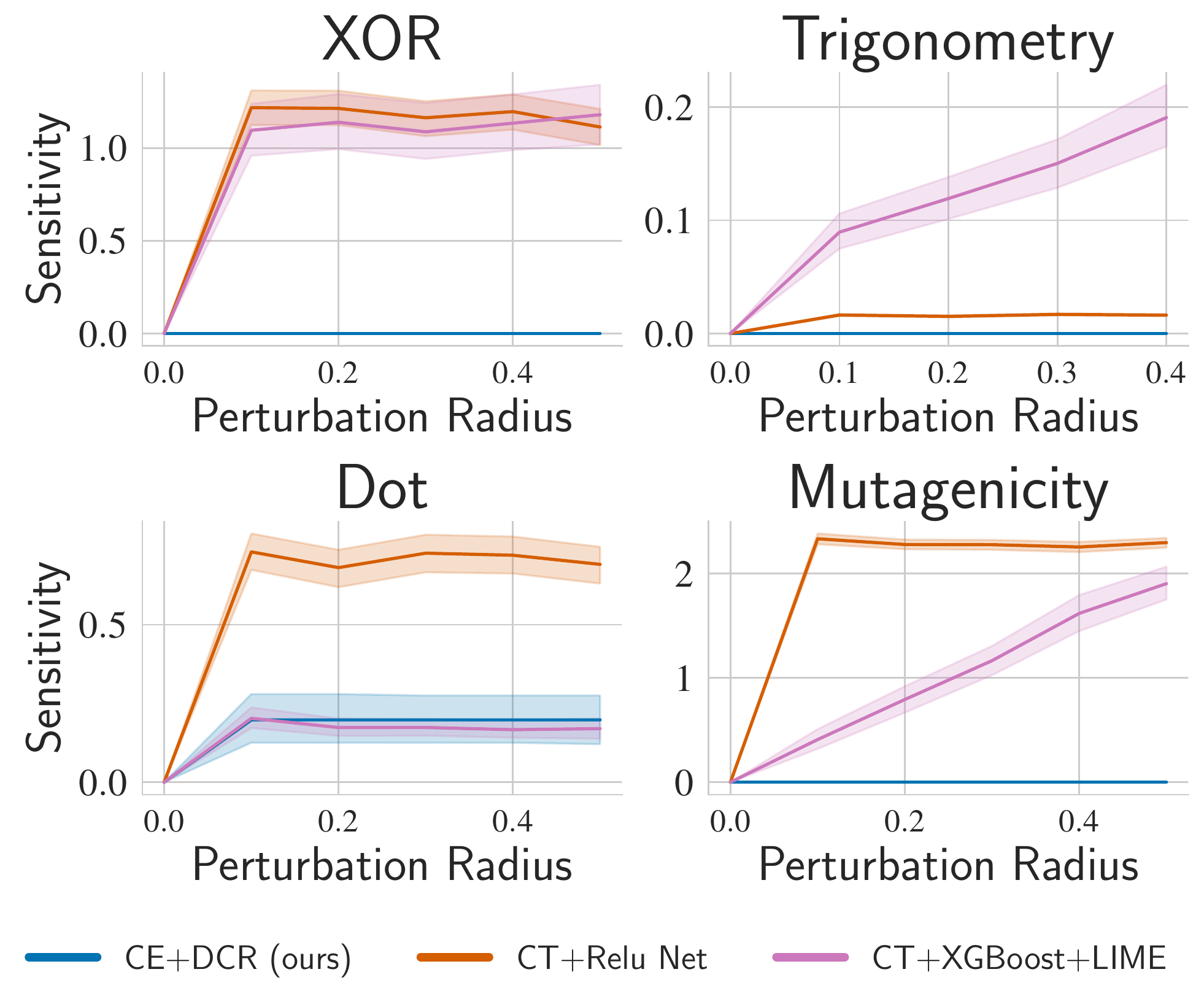}
    \caption{Sensitivity of model explanation when changing the radius of the input perturbation. The lower, the better. DCR explanations engender trust as they are stable under small perturbations of the input. The same does not hold generally for LIME explanations of XGBoost or Relu Net decision rules.}
    \label{fig:sensitivity}
\end{figure}

\paragraph{
DCR rules are stable under small perturbations (Figure~\ref{fig:sensitivity})}
An important characteristic of local explanations is to be stable under small perturbations \citep{yeh2019fidelity}. Indeed, users do not trust explanations if they change significantly on very similar inputs for which the model makes the same prediction. This metric, also known as explanation sensitivity, is generally computed as the maximum change in the explanation of a model $\Phi(f)$ on a slightly perturbed input ($x^{\star}$), that is, $|\Phi(f(\mathbf{x}^\star)) - \Phi(f(\mathbf{x}))|,  |\mathbf{x}-\mathbf{x}^\star|_\infty< \epsilon$. We compare the DCR explanations w.r.t. our interpretable baselines as well as w.r.t. LIME~\cite{ribeiro2016should} explaining the output of XGBoost. Since we are using different types of models, we use a normalised version of the sensitivity $ |\Phi(f(\mathbf{x}^\star)) - \Phi(f(\mathbf{x}))| / |\Phi(f(\mathbf{x}))|$. 
We compute the distance between two explanations considering the feature importance of the original explanation w.r.t. to the feature importance of the explanation for the perturbed example. For decision tree's rules, we consider the distance between the original path and the path of the perturbed example. 
As highlighted in Figure \ref{fig:sensitivity}, in all datasets the explanations provided by DCR are very stable, particularly w.r.t. LIME and ReluNet. Notice that the figure does not report the explanation sensitivity of logistic regression and decision tree because it is trivially zero as they learn fixed rules for the entire dataset. The area under the sensitivity curves of all methods together with further details concerning this experiment has been reported in Appendix~\ref{app:sensitivity}.

\paragraph{DCR explains prediction mistakes}
In DCR, task predictions are obtained by executing the logic rules. In this sense, the rules transparently represent the model behavior, and they can explain misclassifications at the level of tasks. For example, reading DCR rules we can observe that a task was mispredicted because some concepts have been predicted wrongly, or the relevance scores are selecting a suboptimal set of concepts. To test this, we analyze the mispredicted test samples in datasets where we have access to ground truth rules as reference (XOR and Trigonometry). Interestingly, DCR is able to identify a mislabeled sample in the XOR dataset (first row of Table~\ref{tab:mistakes}), highlighting an error in the data generation process. In fact, the rule learnt by DCR is correct $y=0 \leftarrow \neg c_0 \wedge \neg c_1$ but the ground truth label was incorrect $y=1$. In the Trigonometric dataset instead, concepts were mispredicted, thus leading to incorrect rules. 

\begin{table}[H]
\centering
\caption{DCR explains prediction errors.}
\label{tab:mistakes}
\resizebox{\columnwidth}{!}{%
\begin{tabular}{llll}
\hline
\textbf{Dataset} & \textbf{Concepts} & \textbf{DCR rule} & \textbf{Ground truth label} \\ \hline
XOR & {[}0.0, 0.0{]} & $y=0 \leftarrow \neg c_0 \wedge \neg c_1$ & $y=1$ \\
Trigonometry & {[}0.0, 1.0, 1.0{]} & $y=1 \leftarrow \neg c_0 \wedge c_1 \wedge c_2$ & $y=0$ \\
Trigonometry & {[}0.0, 1.0, 0.0{]} & $y=0 \leftarrow \neg c_0 \wedge c_1 \wedge \neg c_2$ & $y=1$ \\
Trigonometry & {[}0.0, 1.0, 1.0{]} & $y=1 \leftarrow \neg c_0 \wedge c_1 \wedge c_2$ & $y=0$ \\
Trigonometry & {[}0.0, 1.0, 1.0{]} & $y=1 \leftarrow \neg c_0 \wedge c_1 \wedge c_2$ & $y=0$ \\ \hline
\end{tabular}
}
\end{table}

\begin{figure}[t]
    \centering
    \includegraphics[width=.965\columnwidth]{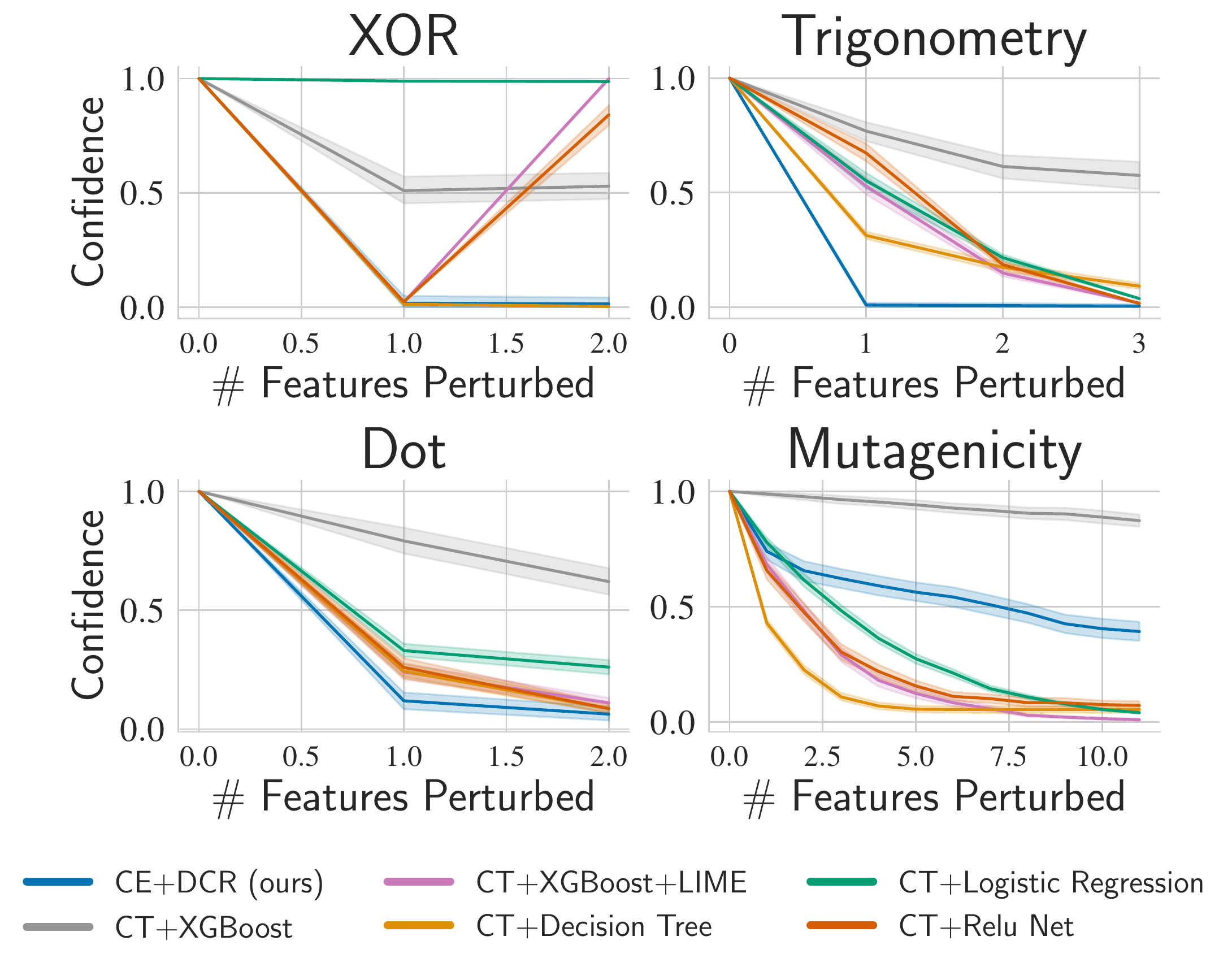}
    \caption{Model confidence as a function of the number of perturbed features on counterfactual examples. The lower, the better. Similarly to interpretable methods, DCR prediction confidence quickly drops after inverting the truth degree of a small set of relevant concepts, facilitating the discovery of counterfactual examples. }
    \label{fig:counterfactuals}
\end{figure}


\paragraph{DCR enables discovering counterfactual examples (Figure~\ref{fig:counterfactuals})}
Besides being stable, DCR rules can be used to find simple counterfactual examples, as introduced in Section~\ref{sec:ruleadv}. In Figure~\ref{fig:counterfactuals} we show a model's confidence in its predictions as we increase the number of concept perturbations. In making perturbations, we sort concepts from the most relevant to the least using DCR rules, as suggested by~\citet{wachter2017counterfactual}. 
Our results show that DCR confidence in its predictions drops quickly when we perturb the most relevant concepts according to a given rule. 
This enables us to discover counterfactual examples where the concept literals are very similar to the original one rule. 
This behaviour is emblematic of interpretable models such as decision trees and logistic regression, for which similar conclusions can be drawn. 
We also observe how in \emph{Mutagenicity} DCR confidence is a bit higher than interpretable baselines. We can explain this behavior as for this challenging dataset DCR rules give equal relevance to a larger set of concepts. Still, DCR confidence is much lower than a black box such as XGBoost.
Local explainers such as LIME can only partially explain the decision process of black box models such as XGBoost: LIME areas under the model confidence curve are generally higher than the other methods. We report the actual values for all methods in Table \ref{tab:counterfactuals} together with further details and counterfactual examples.

\section{Key findings \& significance}
\label{sec:disc}

\paragraph{Limitations}
One of the main limitations of DCR is that its global behavior may not be directly interpretable, which means that global rules may not perfectly align with the exact reasoning of the model. This could be an issue in cases where a user requires a precise understanding of the global model behavior. Also, the complexity of DCR rules may increase significantly when the difference between two tasks can only be determined by using a very high number of concepts. However, in most real-world cases, and in current benchmark datasets for concept-based models, this issue rarely arises. Finally, DCR requires concept embeddings as inputs, which assumes the existence of concept-based datasets or high-quality concept-discovery methods.

\paragraph{Relations with concept-based methods}
Interpretable concept-based models~\cite{koh2020concept} address the lack of human trust in AI systems as they allow their users to understand their decision process~\cite{rudin2019stop,ghorbani2019interpretation,barbiero2023categorical}.
These approaches come with several advantages over other explainability methods as they circumvent the brittleness of post-hoc methods~\citep{adebayo2018sanity, kindermans2019reliability} and provide a semantic advantage in settings where input features are naturally hard to reason about (e.g., raw image pixels) by providing explanations in terms of human-interpretable concepts~\cite{ghorbani2019interpretation,georgiev2022algorithmic,azzolin2022global,magister2022encoding,xuanyuan2022global}.
However,~\citet{zarlenga2022concept} and~\citet{mahinpei2021promises} emphasise how state-of-the-art concept-based models
either struggle to efficiently solve real-world tasks using concept truth-values only or they weaken their interpretability using concept embeddings to increase their learning capacity. This is true even when concept-based models use a simple logistic regression or decision tree to map concept embeddings to tasks because concept embedding dimensions do not have a clear semantic meaning, and models composing such dimensions generate prediction rules that are not human-interpretable. Our work solves this issue by introducing the first interpretable concept-based model that learns logic rules from concept embeddings. 

\paragraph{Relations with neural-symbolic methods} A common paradigm in neural-symbolic is to exploit deep learning models to map subsymbolic information (e.g. images) to an intermediate logical representation, which is then manipulated using weighted logic formalisms, such as probabilistic logic (DeepProbLog~\citep{manhaeve2018deepproblog}, NeurASP~\citep{yang2020neurasp}), fuzzy logic (Lyrics\citep{marra2020lyrics}, LTN~\citep{badreddine2022logic,wagner2021neural}) or both (DLM~\citep{marra2020integrating}, RNM\citep{marra2020relational}). This sets DCR between concept-based and neural-symbolic models. However, while these neural symbolic models focus on how to maximally exploit available logic knowledge (e.g. a logic program) to improve neural predictions, DCR focuses on learning such logical knowledge. Other neural symbolic approaches, such as Neuro-Symbolic Concept Learner~\citep{mao2019neuro}, the Neural Logic Machines~\citep{dong2019neural}, and the Neural State Machine~\citep{hudson2019learning}, are actually closer in spirit to concept based models as they exploit intermediate symbolic representations. However, the decision-making process on top of the concepts/symbols still relies on (or is uniquely) an uninterpretable neural component. In contrast, DCR encodes its decision process in a logical rule that is executed explicitly giving the user full knowledge and control over the concept-to-task decision-making process. This would be impossible in these neural symbolic approaches, as the decision process is implicit in the weights of the networks. 

\paragraph{Key advantages of DCR} 
The main advantage of DCR w.r.t. existing interpretable and black-box methods arises when dealing with challenging tasks where both interpretability and accuracy should be maximized. For simpler tasks, existing interpretable methods, such logistic regression, could be enough. On the other side, when interpretability is not a hard user requirement, then a simple black-box model would be easier to set up (e.g., it does not require concept labels or concept encoders). However, in all cases where interpretability plays a crucial role for the end user and existing interpretable models fail, then DCR could be preferable. Finally, compared to existing neural-symbolic approaches, DCR has an edge in all settings where the rules are unknown, while other methods (like DeepProbLog) might be more stable when the full set of rules is known in advance. For other limitations/drawbacks, please see our reply to common questions.



\paragraph{Conclusion}
This work presents the \textit{Deep Concept Reasoner} (DCR), the new state-of-the-art of interpretable concept-based models. To achieve this, DCR builds for each sample a weighted logic rule combining neural and symbolic algorithms on concept embeddings in a unified end-to-end differentiable system. In our experiments, we compare DCR with state-of-the-art interpretable concept-based models and black-box models using datasets spanning three of the most common data types used in deep learning: tabular, image, and graph data. 
Our experiments show that Deep Concept Reasoners: (i) attain better task accuracy w.r.t.\ state-of-the-art interpretable concept-based models, (ii) discover meaningful logic rules, and (iii) facilitate the generation of counterfactual examples. 
While the global behaviour of the model is still not directly interpretable, our results show how aggregating Boolean DCR rules provides an approximation for the global behaviour of the model which matches known ground truth relationships. As a result, our experiments indicate that DCR represents a significant advance over the current state-of-the-art of interpretable concept-based models, and thus makes progress on a key research topic within the field of explainability.

\section*{Acknowledgements}
The authors would like to thank Nikola Simidjievski for his insightful comments on earlier versions of this manuscript. PB acknowledges support from the European Union's Horizon 2020 research and innovation programme under grant agreement No 848077. GC and FP acknowledges support from the EU Horizon 2020 project AI4Media, under contract no. 951911 and by the French government, through Investments in the Future projects managed by the National Research Agency (ANR), 3IA Cote d’Azur with the reference number ANR-19-P3IA-0002. MEZ acknowledges support from the Gates Cambridge Trust via a Gates Cambridge Scholarship. FG was supported by TAILOR and by HumanE-AI-Net projects funded by EU Horizon 2020 research and innovation programme under GA No 952215 and No 952026, respectively.




\bibliography{references}
\bibliographystyle{icml2023}

\newpage
\appendix
\onecolumn

\section{Datasets \& Experimental Setup}
\label{app:datasets}

\paragraph{XOR dataset} 
The first dataset used in our experiments is inspired by the exclusive-OR (XOR) problem proposed by~\cite{minsky2017perceptrons} to show the limitations of Perceptrons. We draw input samples from a uniform distribution in the unit square $\mathbf{x} \in [0,1]^2$ and define two binary concepts \{$c_1, c_2\}$ by using the Boolean (discrete) version of the input features $c_i = \mathbb{1}_{x_i > 0.5}$. Finally, we construct a downstream task label using the XOR of the two concepts $y = c_1 \oplus c_2$.

\paragraph{Trigonometric dataset}
The second dataset we use in our experiments is inspired by that proposed by ~\citet{mahinpei2021promises} (see Appendix D of their paper). Specifically, we construct synthetic concept-annotated samples from three independent latent normal random variables $h_i \sim \mathcal{N}(0, 2)$. Each of the 7 features in each sample is constructed via a non-invertible function transformation of the latent factors, where 3 features are of the form $(\sin(h_i) + h_i)$, 3 features of the form $(\cos(h_i) + h_i)$, and 1 is the nonlinear combination $(h_1^2 + h_2^2 + h_3^2)$. Each sample is then associated with 3 binary concepts representing the sign of their corresponding latent variables, i.e. $c_i = (h_i>0)$. In order to make this task Boolean-undecidable from its binary concepts, we modify the downstream task proposed by~\citet{mahinpei2021promises} by assigning each sample a label $y = \mathbb{1}_{(h_1 + h_2) > 0}$ indicating whether $h_1 + h_2$ is positive or not.

\paragraph{Vector dataset} \label{sec:appendix_dot_dataset}
As much as the Trigonometric dataset is designed to highlight that fuzzy concept representations generalize better than Boolean concept representations, we designed the Vector dataset to show the advantage of embedding concept representations over fuzzy concept representations. The Vector dataset is based on four 2-dimensional latent factors from which concepts and task labels are constructed. Two of these four vectors correspond to fixed reference vectors $\mathbf{w}_+$ and $\mathbf{w}_-$ while the remaining two vectors $\{\textbf{v}_i\}_{i=1}^2$ are sampled from a 2-dimensional normal distribution. We then create four input features as the sum and difference of the two factors $\mathbf{v}_i$. From this, we create two binary concepts representing whether or not the latent factors $\mathbf{v}_i$ point in the same direction as the reference vectors $\mathbf{w}_j$ (as determined by their dot products). Finally, we construct the downstream task as determining whether or not vectors $\mathbf{v}_1$ and $\mathbf{v}_2$ point in the same direction (as determined by their dot product).

\paragraph{MNIST Addition} 
In the MNIST addition dataset \cite{manhaeve2018deepproblog}, MNIST images are paired and the pair is labelled with the sum of the two corresponding digits. There are 30000 labelled pairs. The two images are given as two separate inputs to the model (i.e. they are not concateneted). 

\paragraph{Mutagenicity} The \textit{Mutagenicity} dataset \citep{morris2020tudataset} is a labelled graph classification dataset, where a graph represents a molecule. The task is to predict whether the molecule is mutagenic or non-mutagenic. The dataset has 4337 graphs. We use the version available as part of the PyTorch Geometric \citep{fey2019fast} library.

\paragraph{CelebA} 
We use the CelebA dataset to simulate a real-world condition where the set of training and test concepts is not the same, though the embeddings of training and test concepts are still correlated. To this end, we work using pre-trained embeddings generated by a Concept Embedding Model in the setting described by~\citet{zarlenga2022concept}. We then select the $3$ most frequent concepts and train DCR and all the other baseline models on these concepts. However, at test time shift the set of concepts and we use the $3$rd, $4$th, and $5$-th most frequent concept to make predictions. While all the first $5$ concepts are highly correlated being attributes in human face images, the shift in distribution is quite significant. DCR can cope with this shift without any modification. However, usually AI models require a fixed number of features at training and test time. For this reason, we use zero-padding on training and test concepts to allow the other baselines to be trained and tested.

\section{Training details}
\subsection{Deep Concept Reasoner}
For all datasets we train DCR using a Godel t-norm semantics. We also implement the neural modules $\phi$ and $\psi$ as with two-layer MLPs with a number of hidden layers given by the size of the concept embeddings. 

For all synthetic datasets (i.e., \textit{XOR}, \textit{Trig}, \textit{Dot}) and for CelebA we train DCR for $3000$ epochs using a temperature of $\tau=100$. In \emph{Mutagenicity} we train DCR for $7000$ epochs using a temperature of $100$.

\subsection{Concept Embedding Generators}
To generate concept embeddings on synthetic datasets (i.e., \textit{XOR}, \textit{Trig}, \textit{Dot}), we use a Concept Embedding Model~\cite{zarlenga2022concept} implemented as an MLP with hidden layer sizes $\{128, 128\}$ and LeakyReLU activations. When learning concept embedding representations in synthetic datasets, we learn embeddings with $m=128$ activations.

In CelebA, we use a Concept Embedding Model on top of a pretrained ResNet-34 model~\cite{he2016deep} with its last layer modified to output $n_\text{hidden} = m$ activations. In this case, we learn embeddings with $m=16$ activations, smaller than in the synthetic datasets given the larger number of concepts in these tasks.

In \emph{Mutagenicity}, we use a Graph Convolutional Network~\cite{scarselli2008graph,morris2019weisfeiler} to map input graphs to the given task. We then extract concept embeddings using GCExplainer~\cite{magister2021gcexplainer,}, a graph-based variant of the Automated Concept-based Explanation proposed by~\citet{ghorbani2019towards} for image data. We implement the GNN with four layers of graph convolutions with $40$ hidden neurons followed by leaky ReLU activation function each. We then apply mean pooling on node embeddings produced by the preceding graph convolutions and extract predictions via a linear readout function with 10 hidden units. We train these networks for $20$ epochs with a learning rate of $0.001$ and a batch size of $16$ graphs, where we use an 80:20 split for the training and testing set. After training, we run GCExplainer on the node embeddings computed before pooling and extract $30$ concepts using k-Means\citep{forgy1965cluster}, where each concept corresponds to a cluster of graph nodes in the embedding space. We encode these cluster labels as one-hot binary arrays and associate each node with the binary label of the closest cluster. We then obtain the concept truth values of a given graph by aggregating the binary labels of its nodes. To generate concept embeddings, we consider the node embeddings closest to the cluster centroids for active concepts.

\paragraph{Training Hyperparameters} 
In all synthetic tasks, we generate datasets with 3,000 samples and use a traditional 70\%-10\%-20\% random split for training, validation, and testing datasets, respectively. During training, we then set the weight of the concept loss to $\alpha = 1$ across all models. We then train all models for 500 epochs using a batch size of $256$ and a default Adam~\cite{kingma2014adam} optimizer with learning rate $10^{-2}$.


In our CelebA task, we fix the concept loss weight to $\alpha = 1$ in all models and also use a weighted cross entropy loss for concept prediction to mitigate imbalances in concept labels. All models in this task are trained for 200 epochs using a batch size of 512 and an SGD optimizer with $0.9$ momentum and learning rate of $5 \times 10^{-3}$.

In all models and tasks, we use a weight decay factor of $4e-05$ and scale the learning rate during training by a factor of $0.1$ if no improvement has been seen in validation loss for the last $10$ epochs. Furthermore, all models are trained using an early stopping mechanism monitoring validation loss and stopping training if no improvement has been seen for 15 epochs.

\subsection{Hyperparameter search for benchmark classifiers}
xWe run a grid search using an internal 3-fold cross-validation to find the optimal settings for benchmark classifiers. The parameter grid we use is:
\begin{itemize}
    \item decision tree
    \begin{itemize}
        \item max depth: [2, 4, 10, all leaves pure]
        \item min\_samples\_split: [2, 4, 10]
        \item min\_samples\_leaf: [1, 2, 5, 10]
    \end{itemize}
    \item logistic regression
    \begin{itemize}
        \item penalty: [l1, l2, elasticnet]
    \end{itemize}
    \item XGBoost
    \begin{itemize}
        \item booster: [tree, linear, dart]
    \end{itemize}
\end{itemize}

\section{Mutagenicity: Extracted Concepts}
\label{app:mutag}
\begin{figure}[H]
    \centering
    \includegraphics[width=0.45\textwidth]{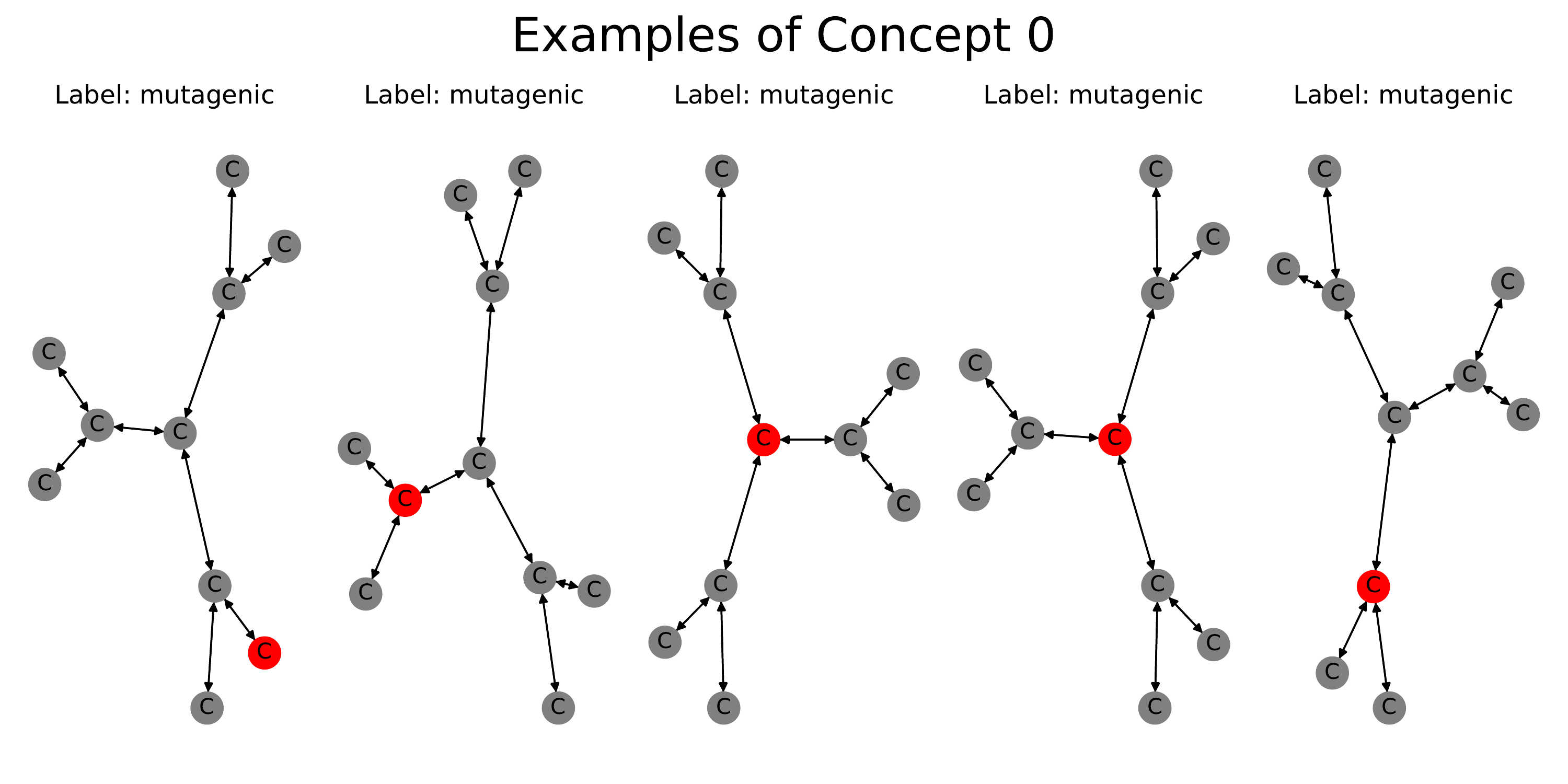}
    \includegraphics[width=0.45\textwidth]{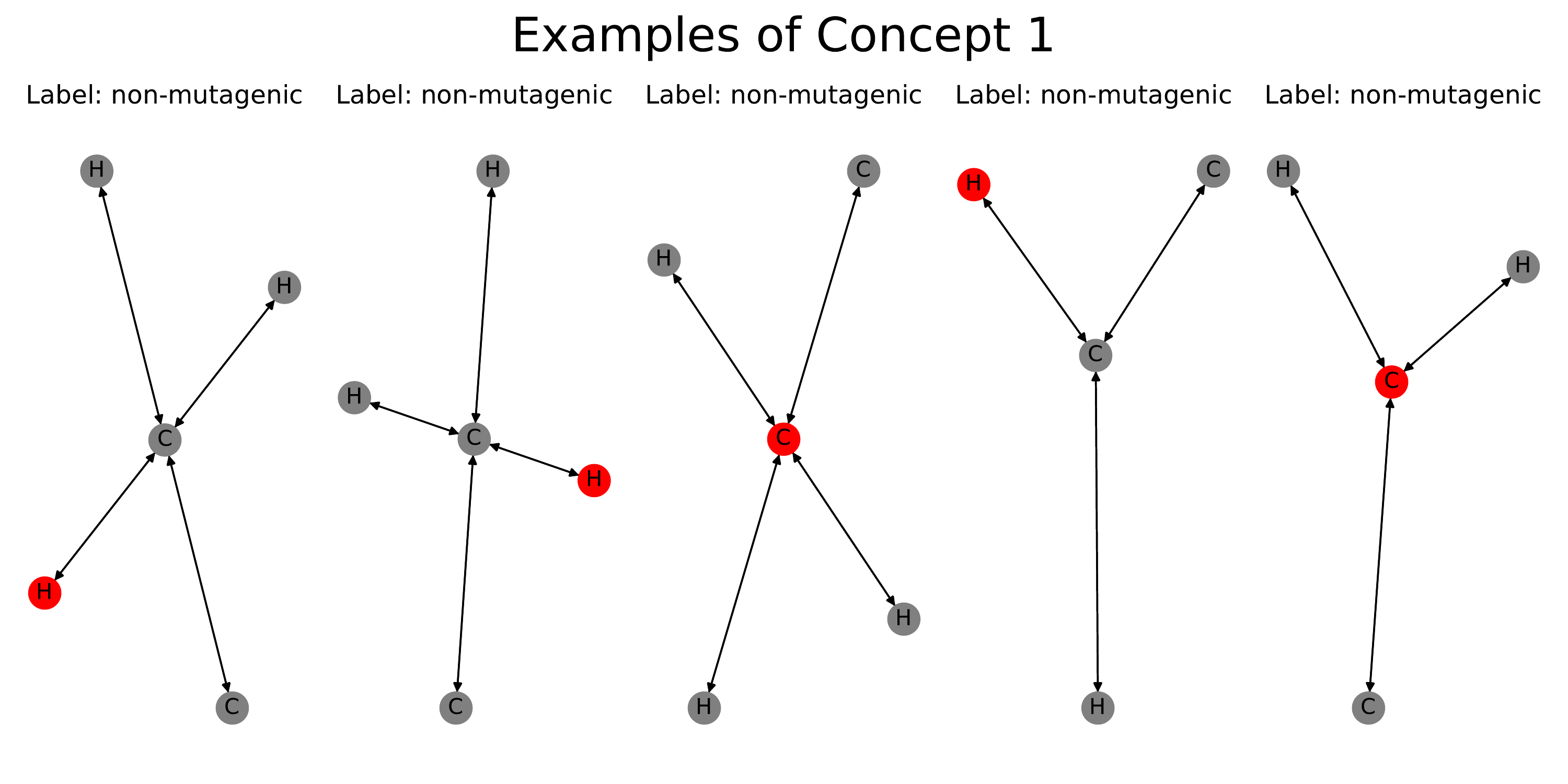}\\
    \includegraphics[width=0.45\textwidth]{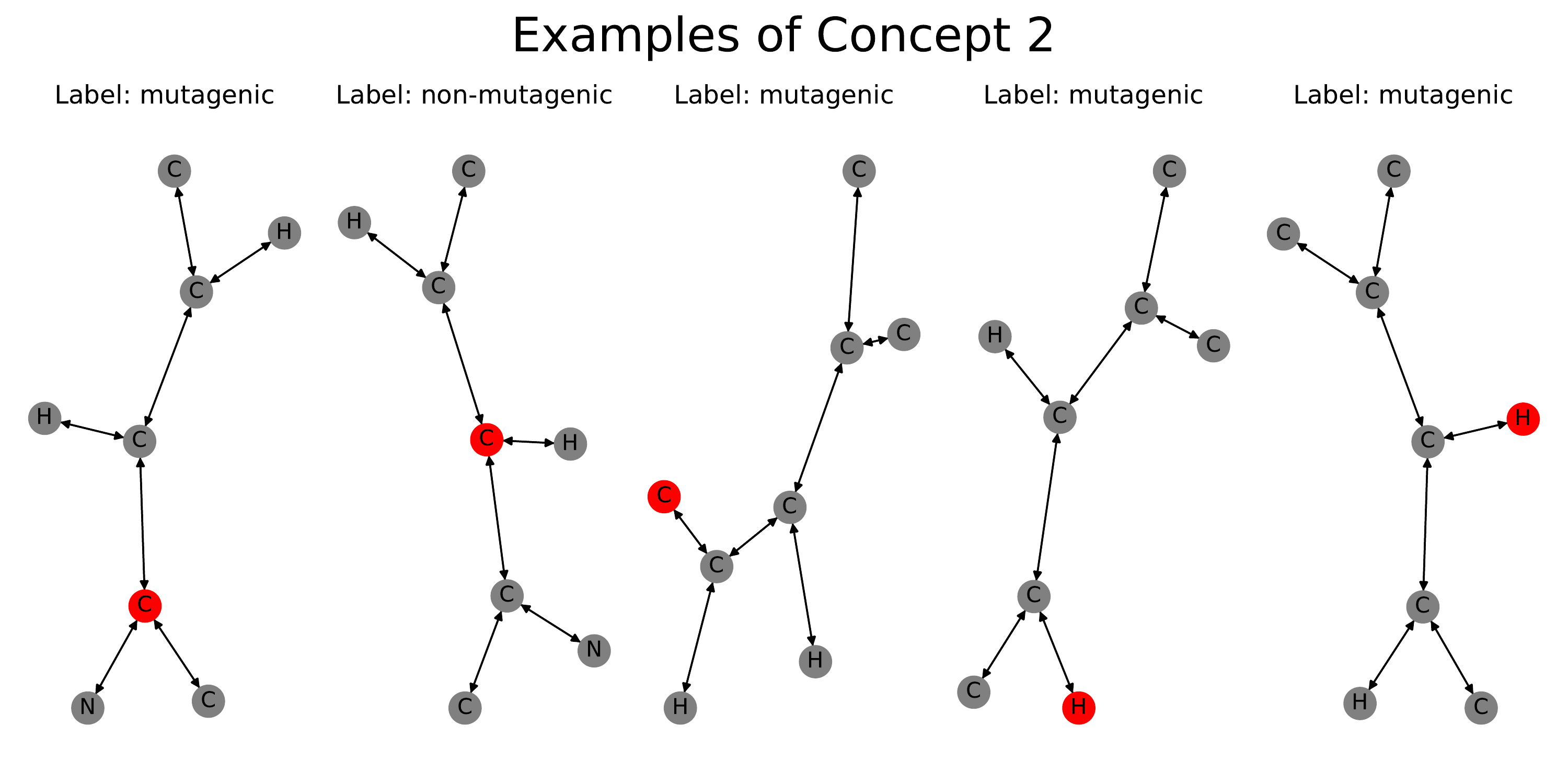}
    \includegraphics[width=0.45\textwidth]{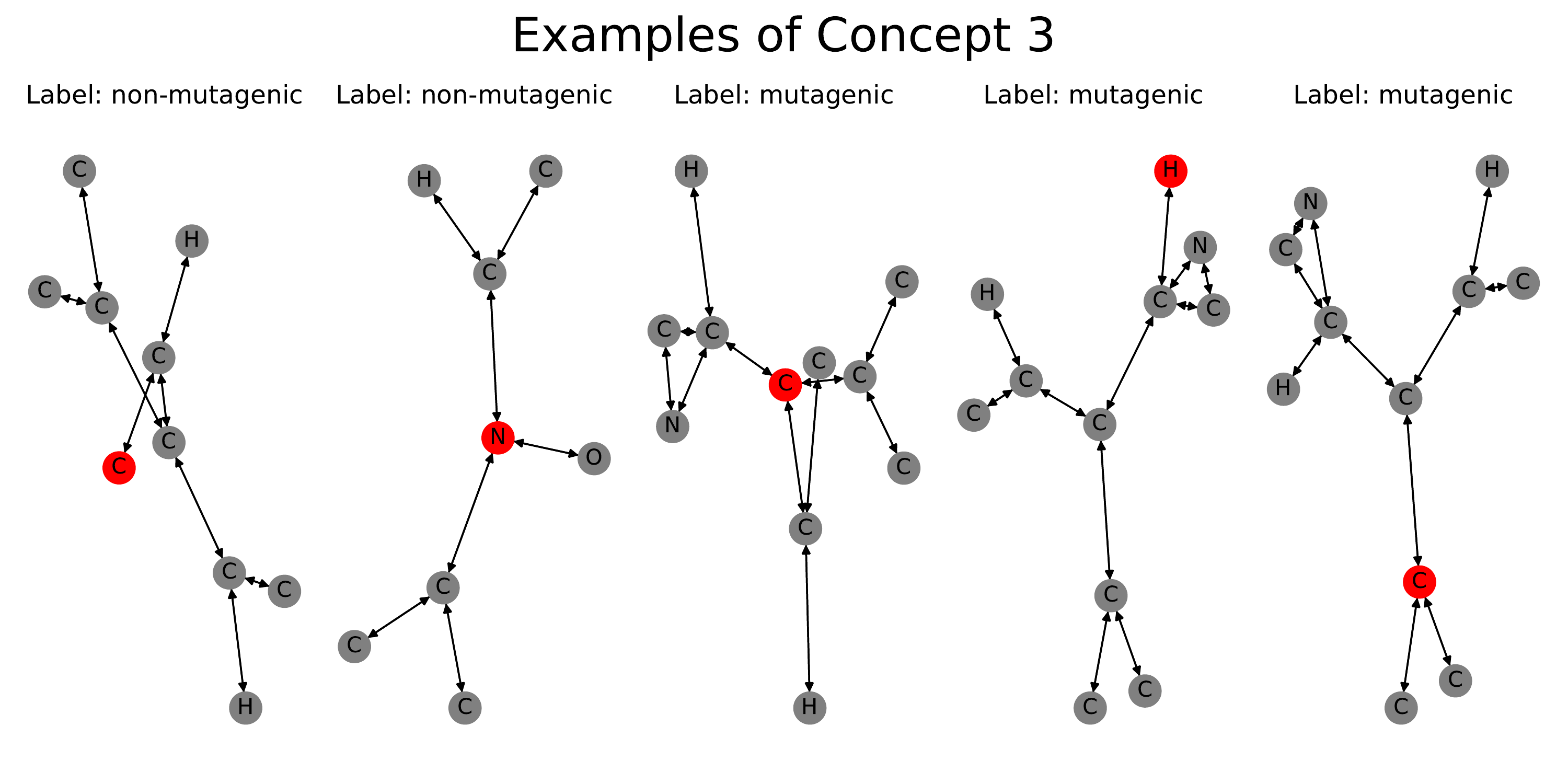}\\
    \includegraphics[width=0.45\textwidth]{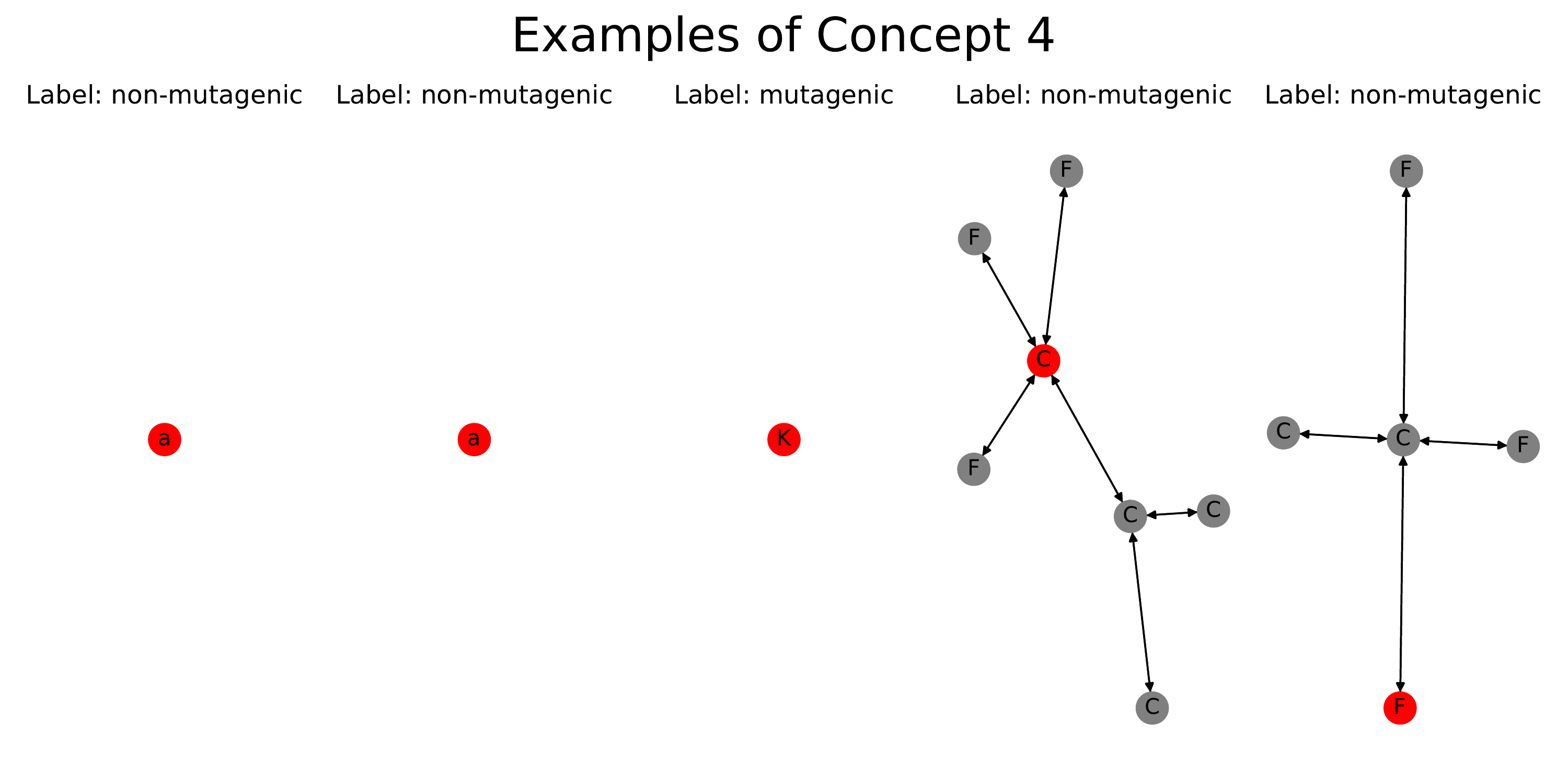}
    \includegraphics[width=0.45\textwidth]{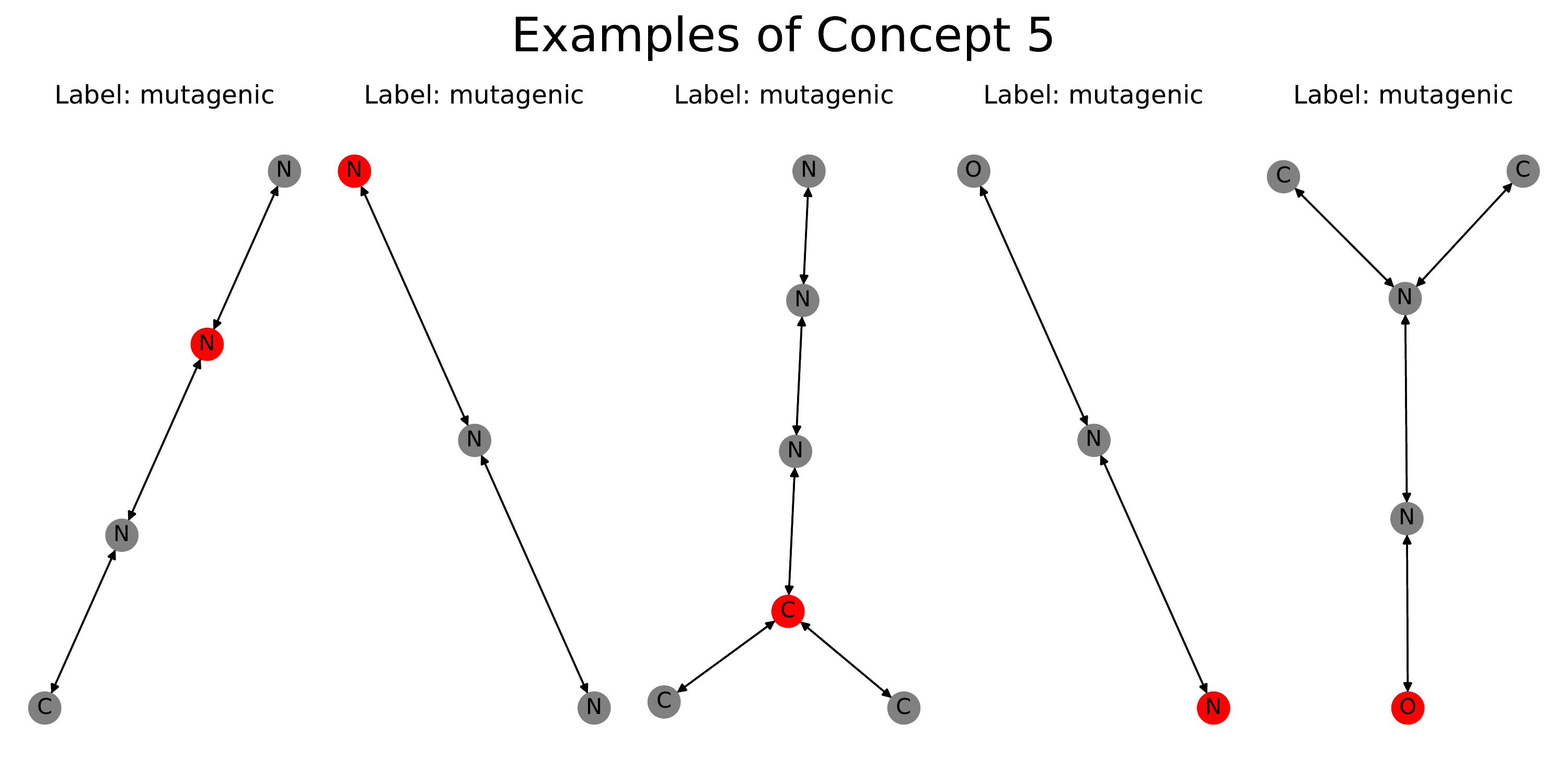}\\
    \includegraphics[width=0.45\textwidth]{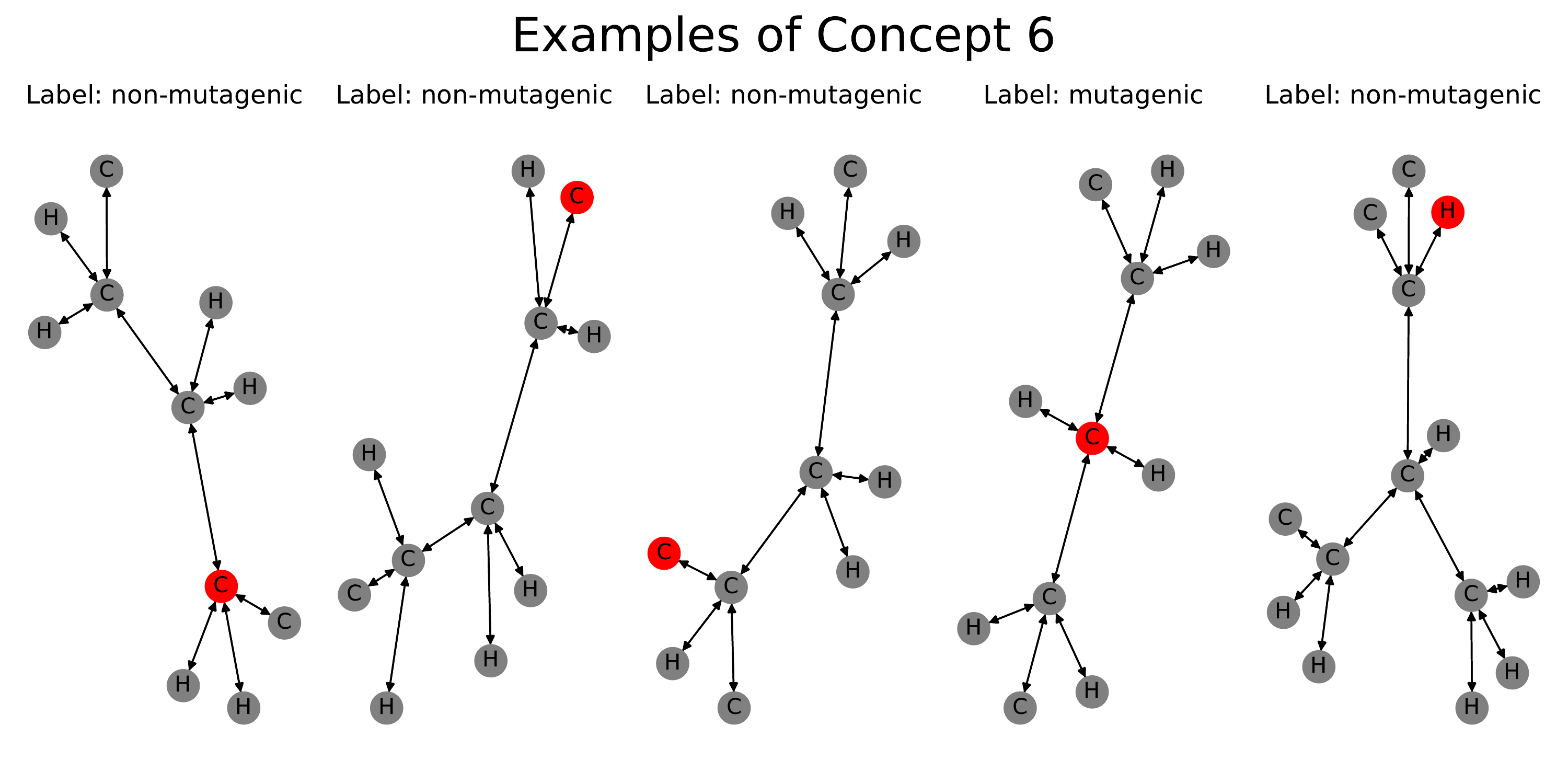}
    \includegraphics[width=0.45\textwidth]{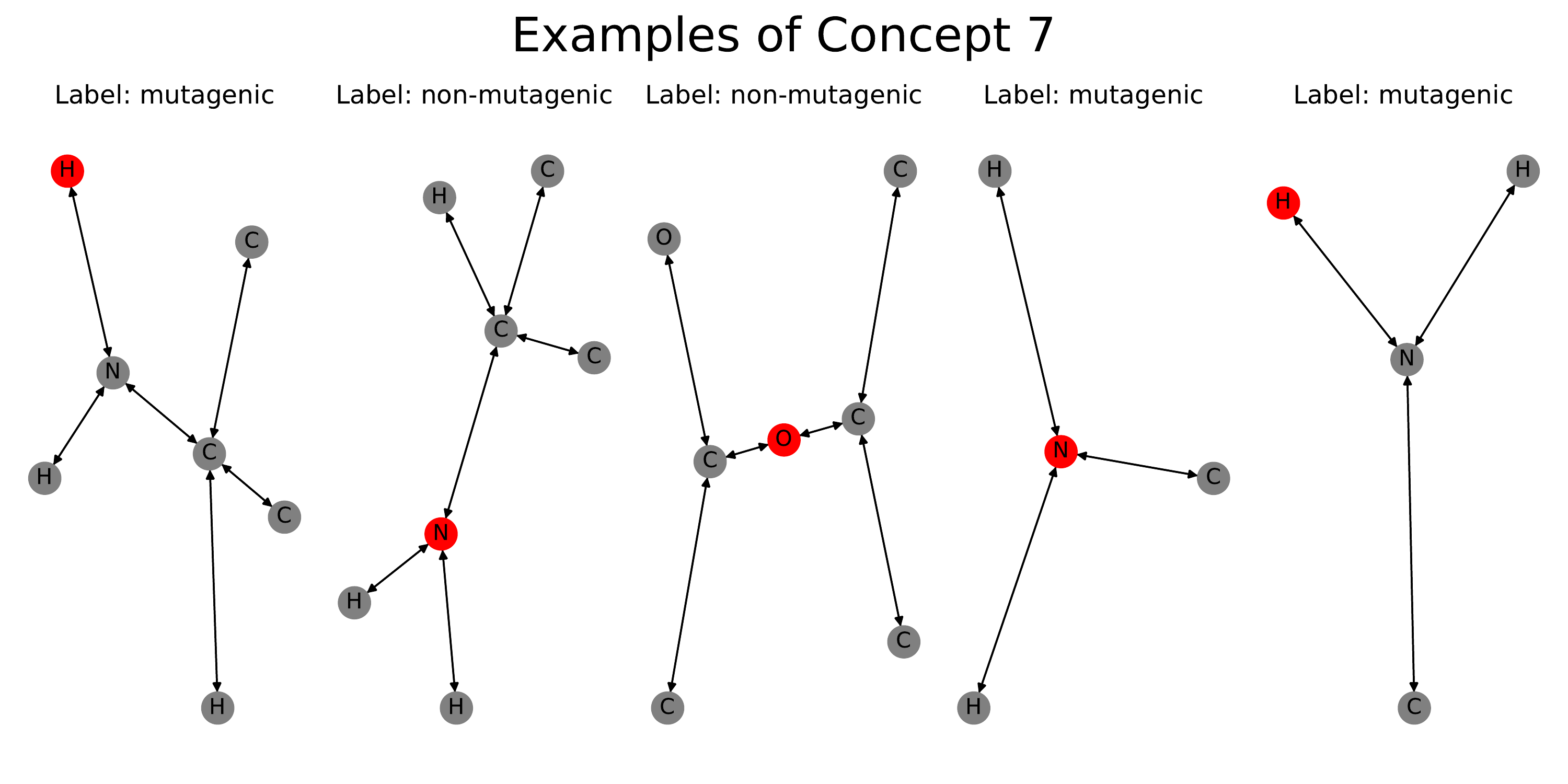}\\
    \includegraphics[width=0.45\textwidth]{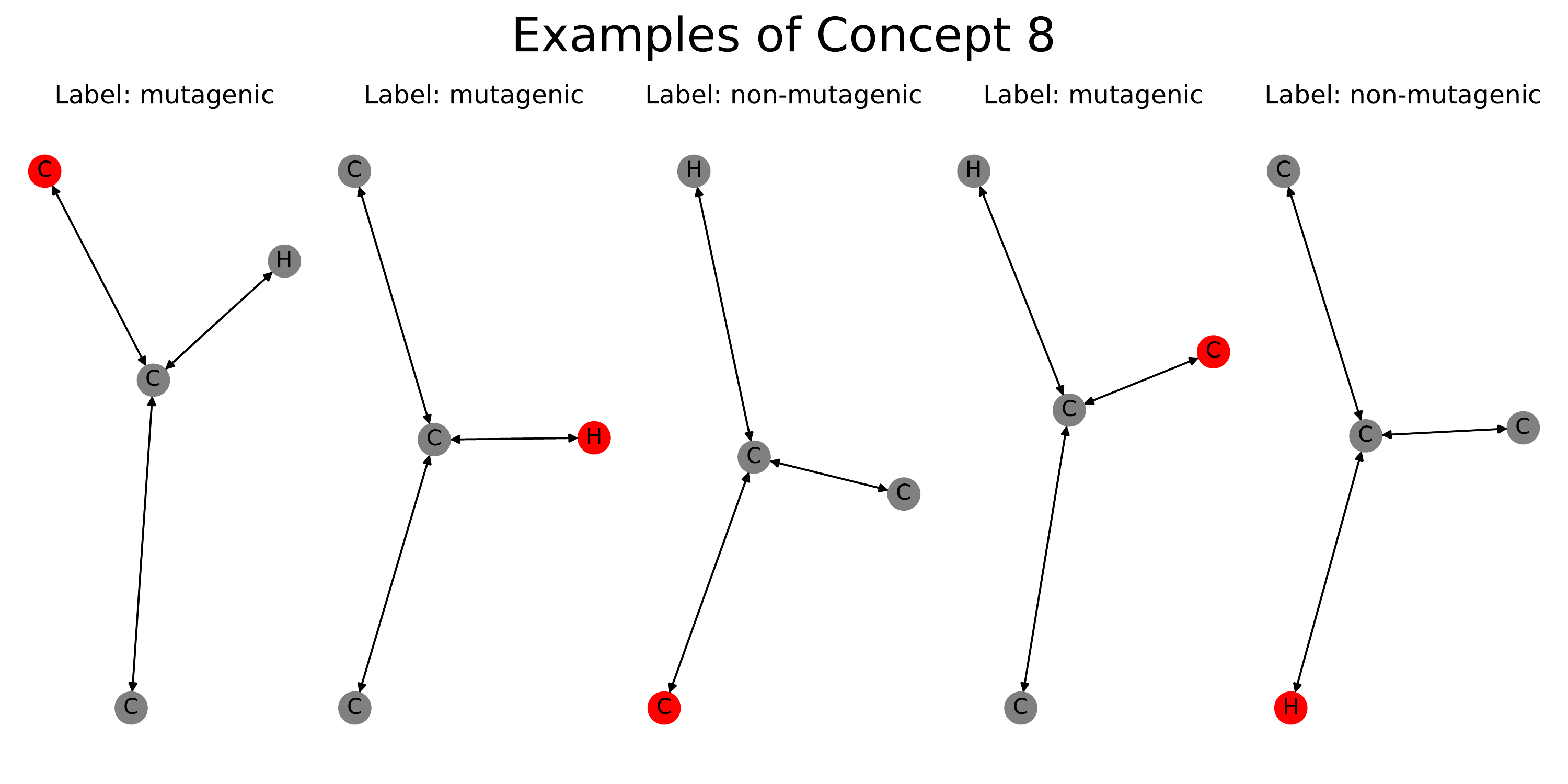}
    \includegraphics[width=0.45\textwidth]{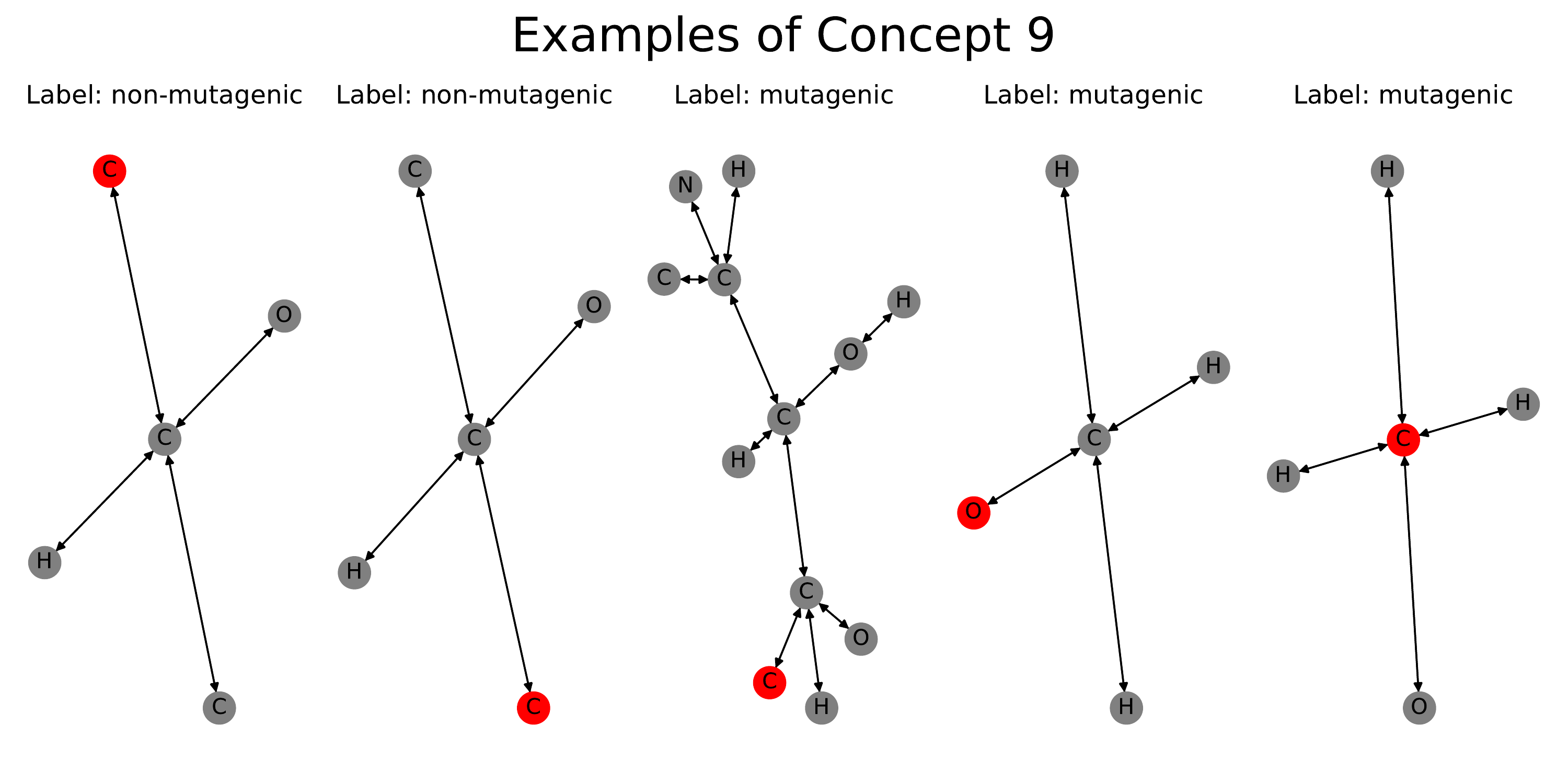}   
    \caption{Concept discovered by the graph concept explainer. Part I.}
    \label{fig:mutag1}
\end{figure}

\begin{figure}[H]
    \centering
    \includegraphics[width=0.45\textwidth]{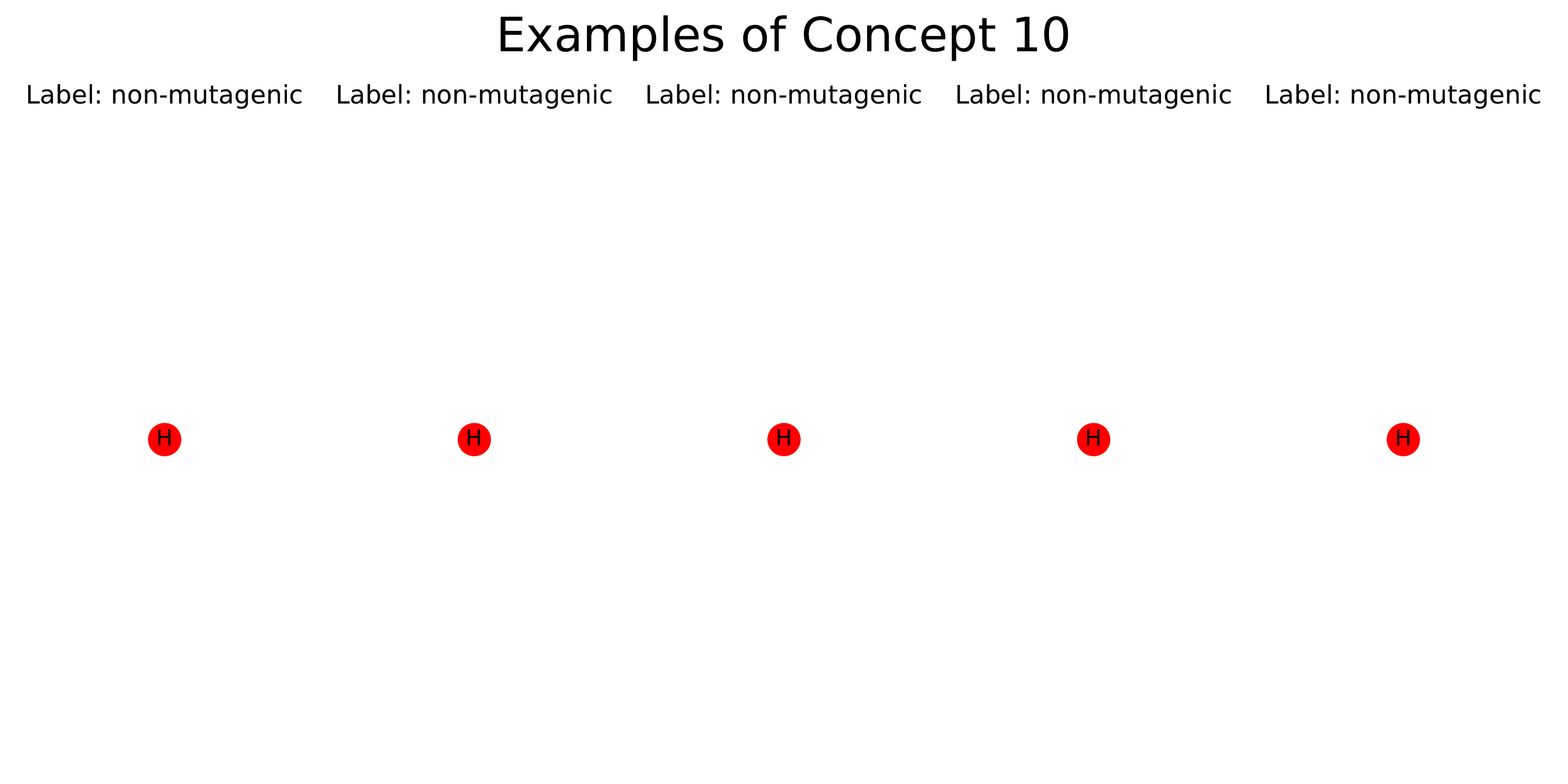}
    \includegraphics[width=0.45\textwidth]{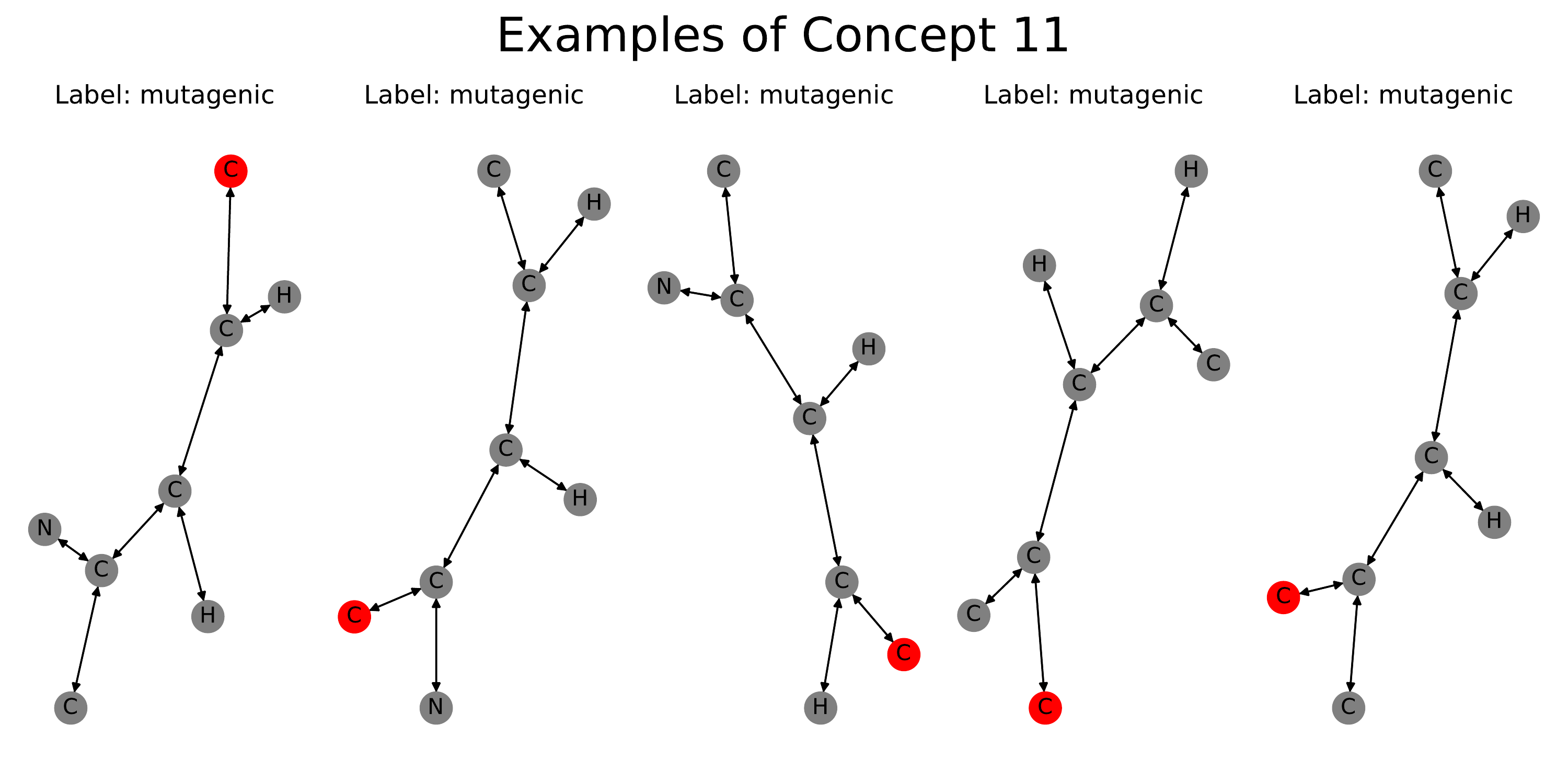}\\
    \includegraphics[width=0.45\textwidth]{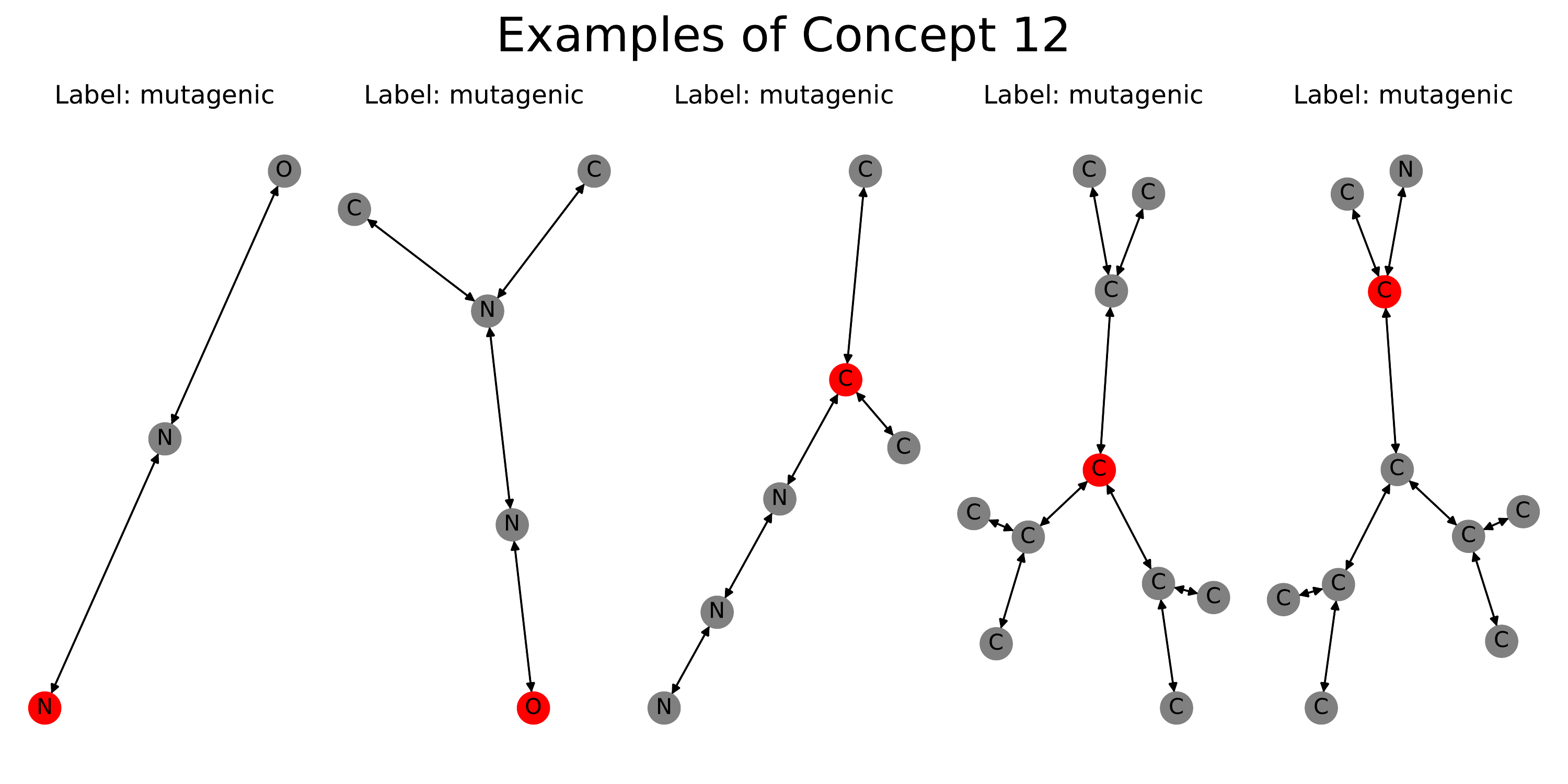}
    \includegraphics[width=0.45\textwidth]{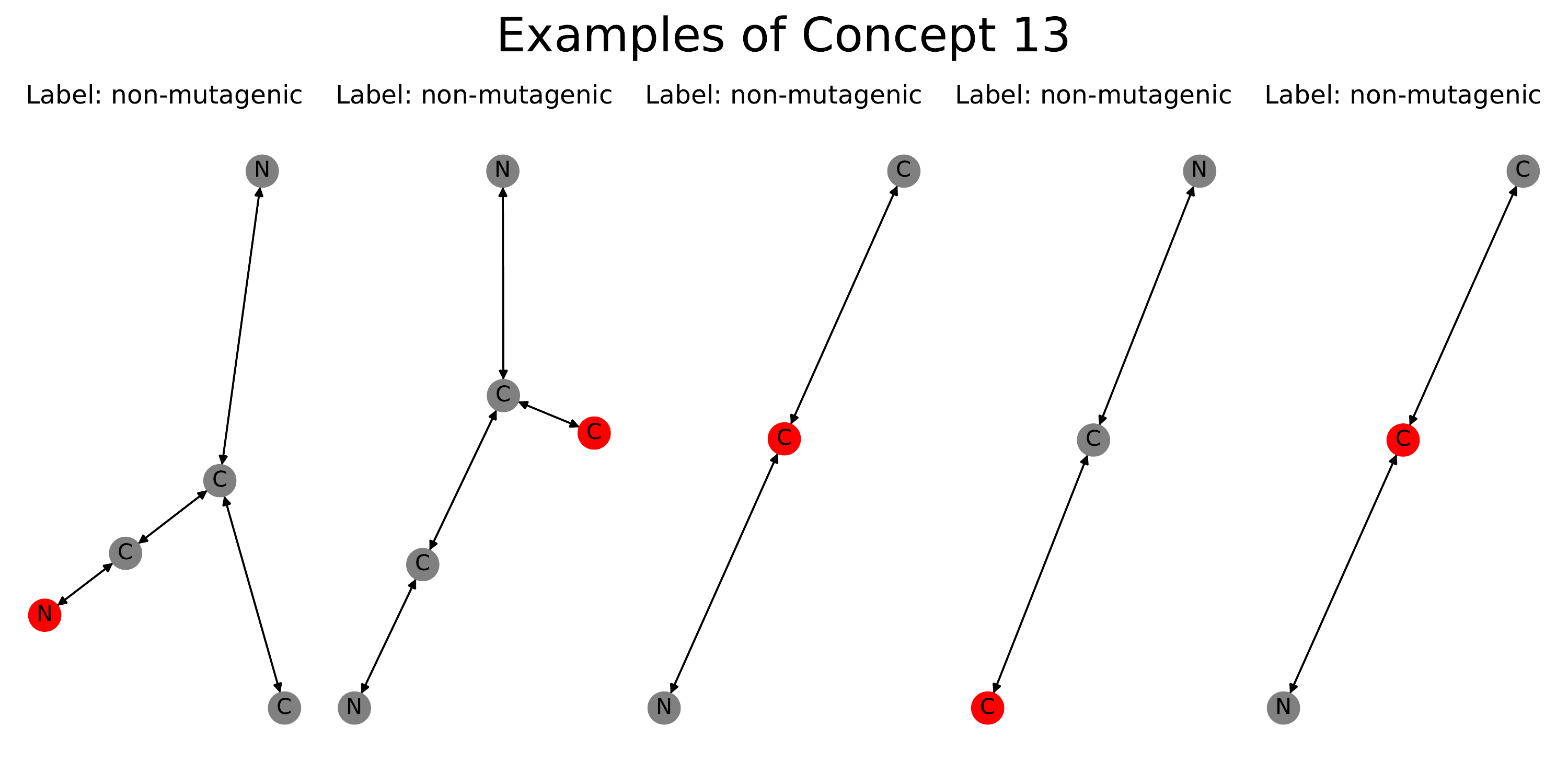}\\
    \includegraphics[width=0.45\textwidth]{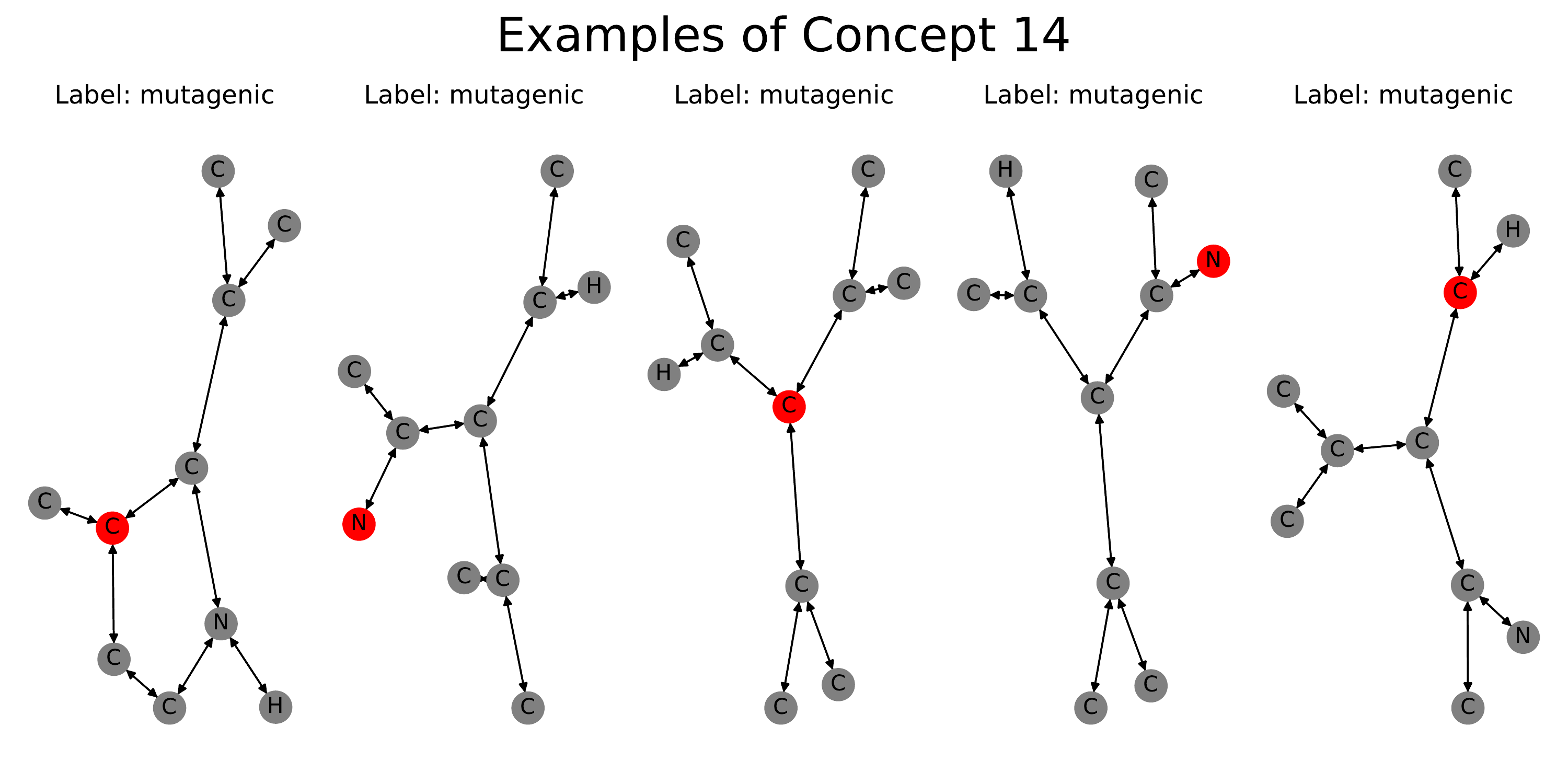}
    \includegraphics[width=0.45\textwidth]{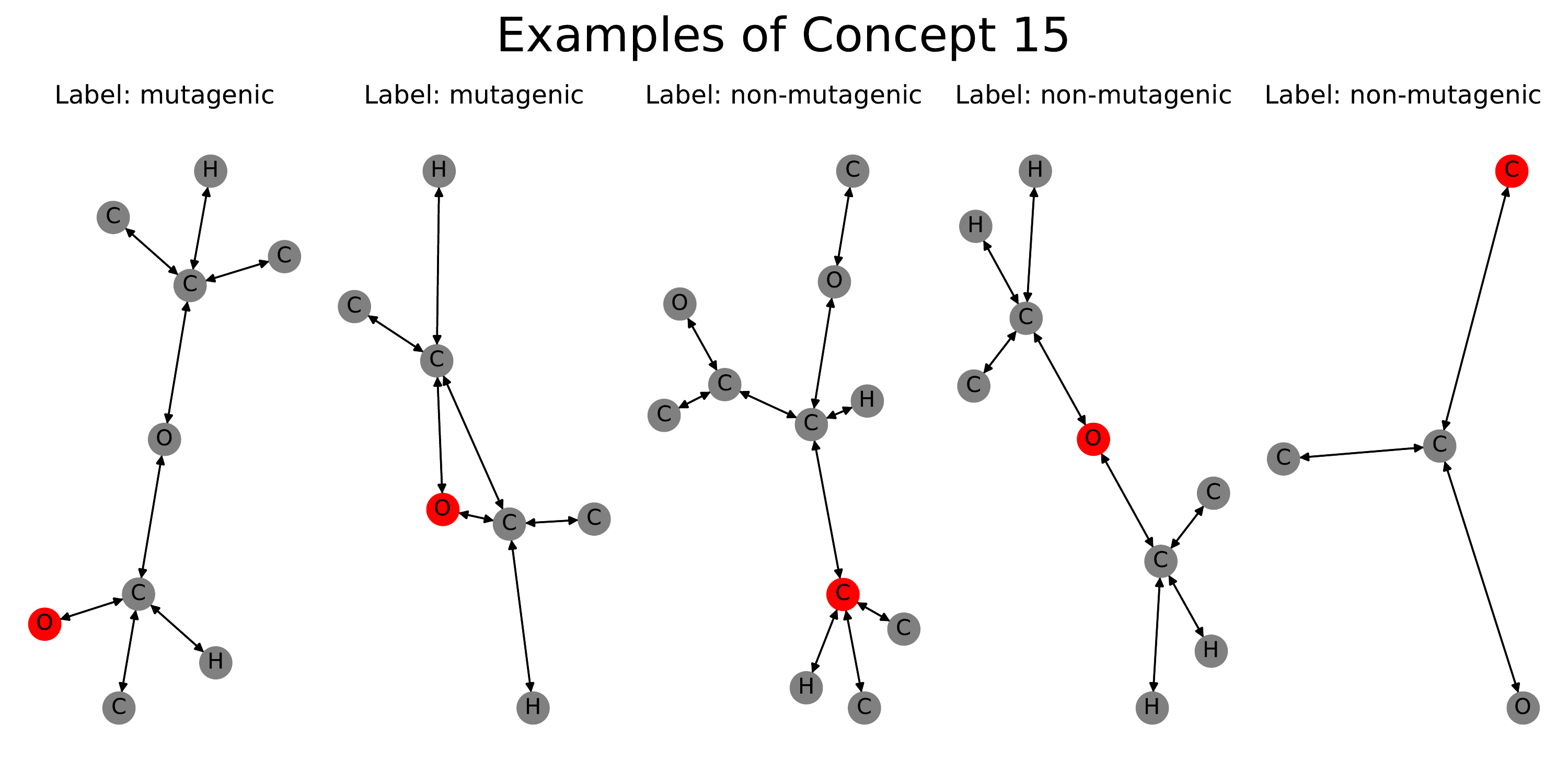}\\
    \includegraphics[width=0.45\textwidth]{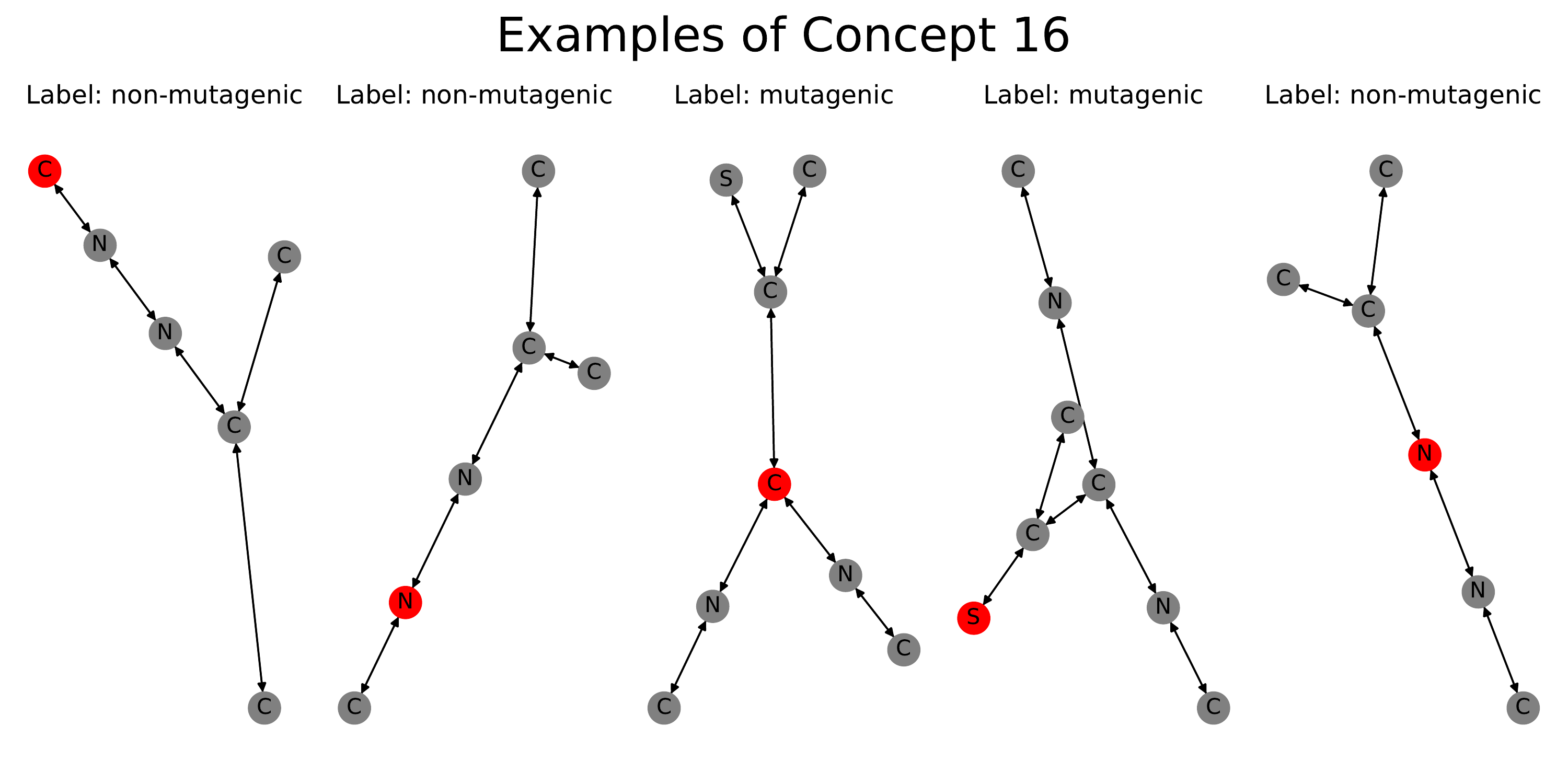}
    \includegraphics[width=0.45\textwidth]{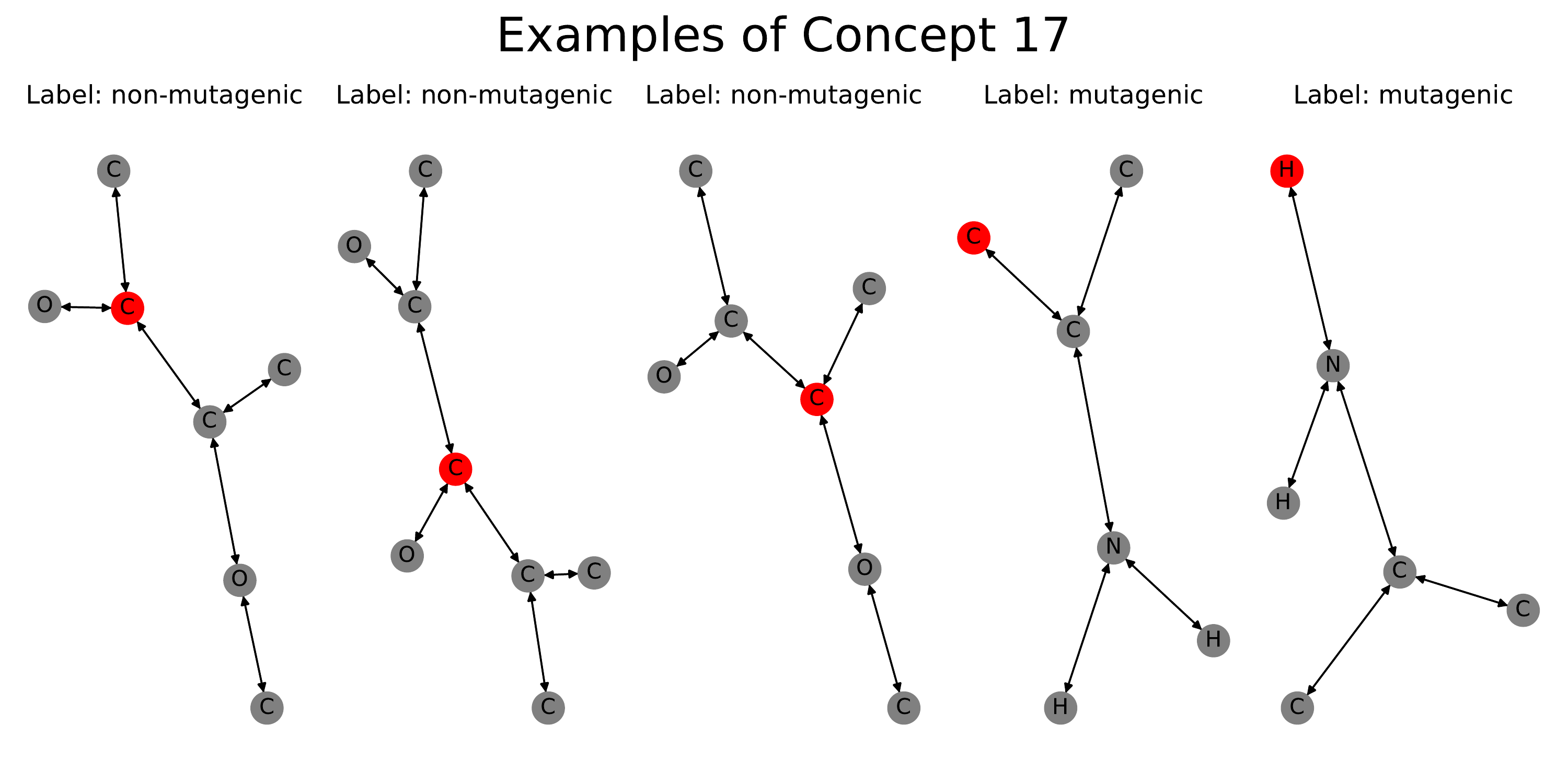}\\
    \includegraphics[width=0.45\textwidth]{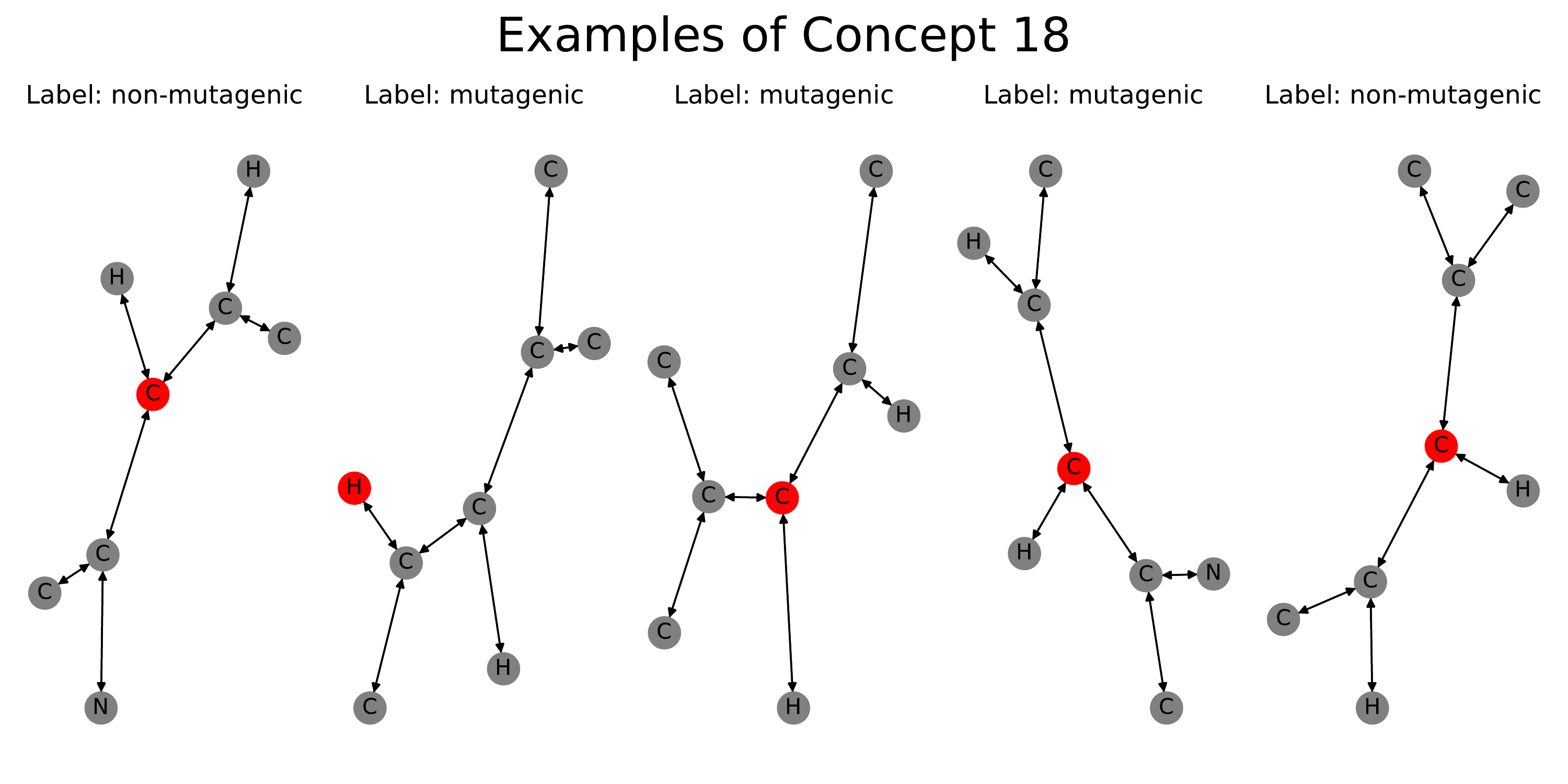}
    \includegraphics[width=0.45\textwidth]{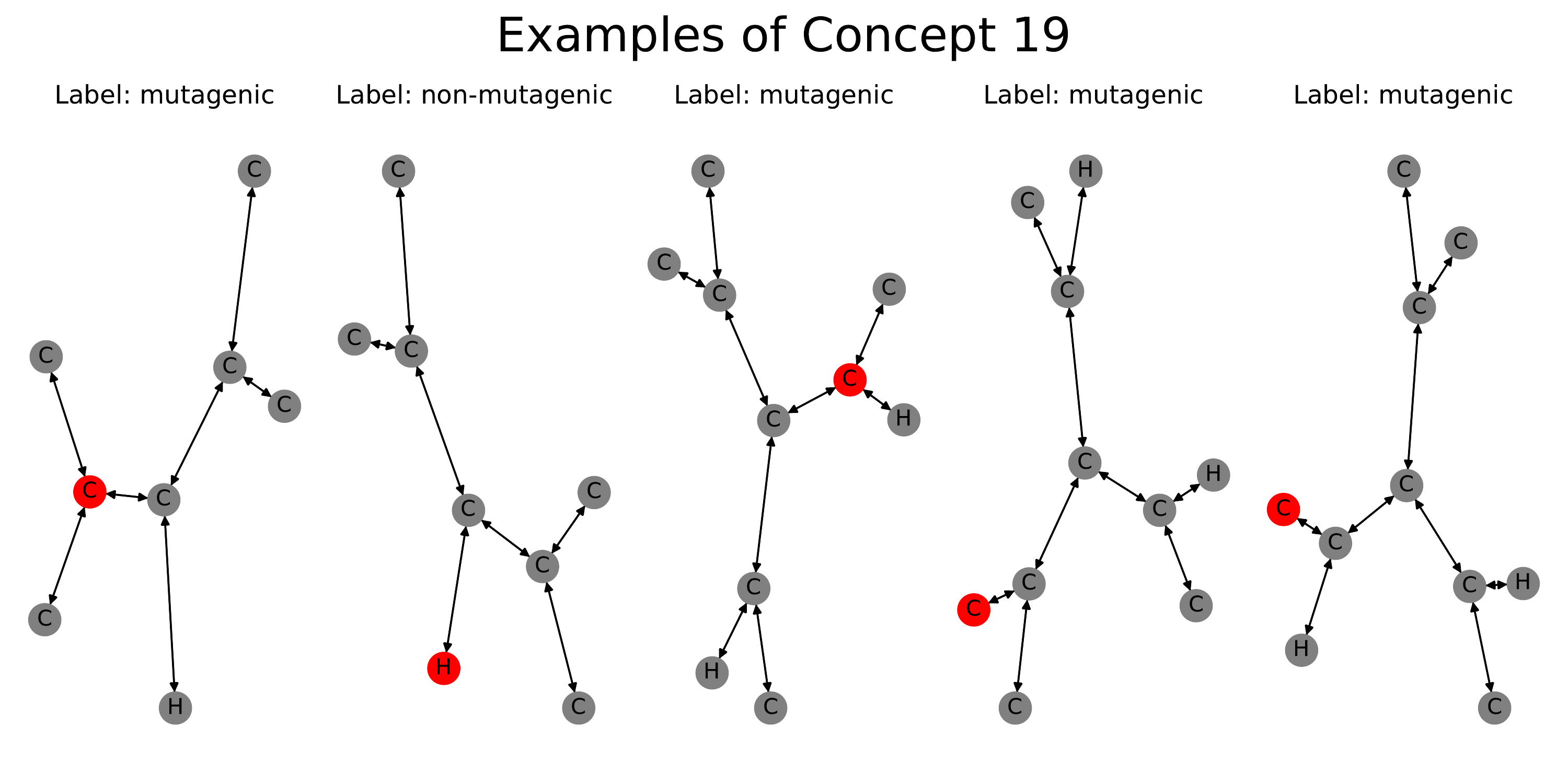}\\
    \caption{Concept discovered by the graph concept explainer. Part II.}
    \label{fig:mutag2}
\end{figure}

\begin{figure}[H]
    \centering
    \includegraphics[width=0.45\textwidth]{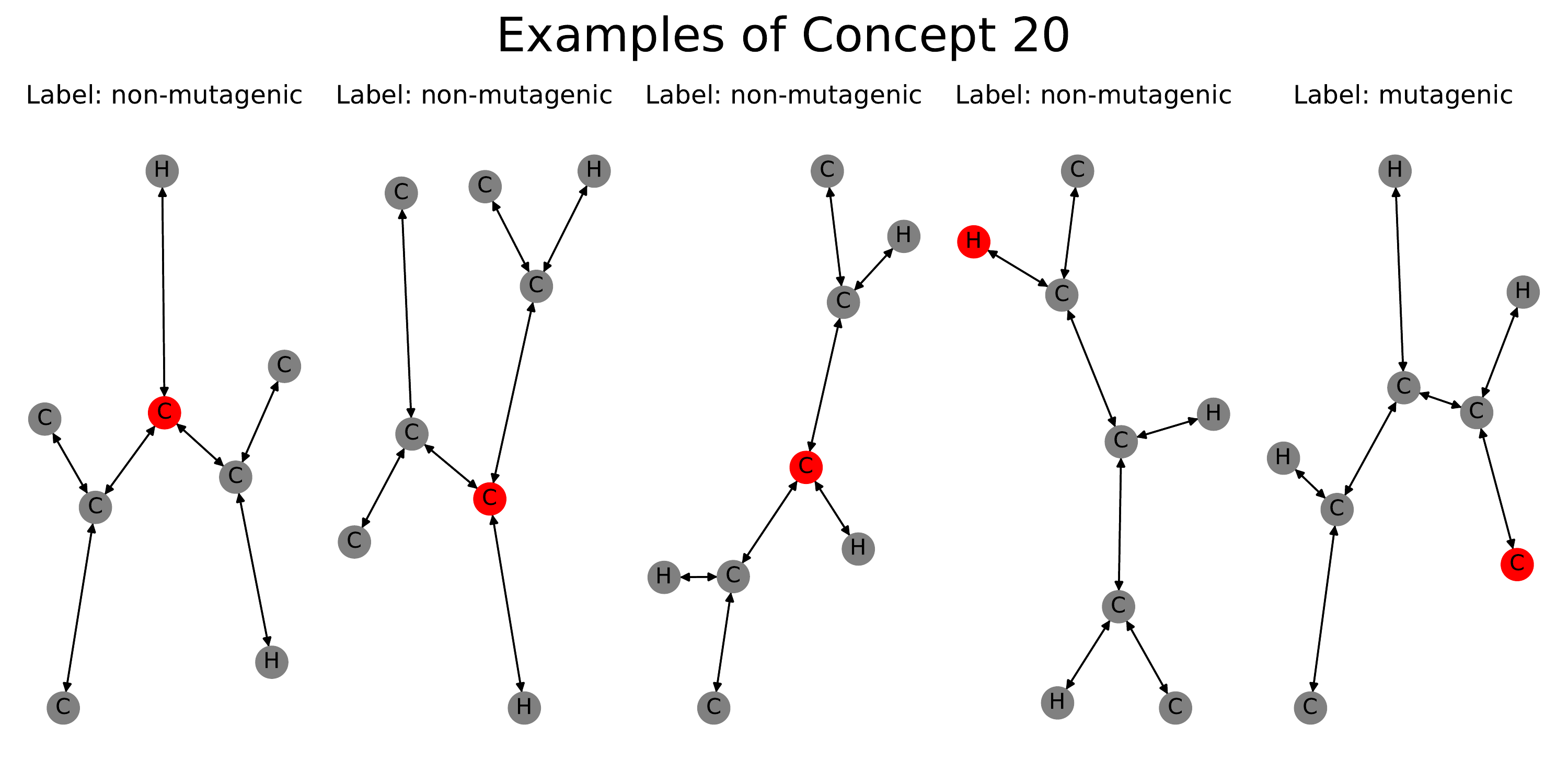}
    \includegraphics[width=0.45\textwidth]{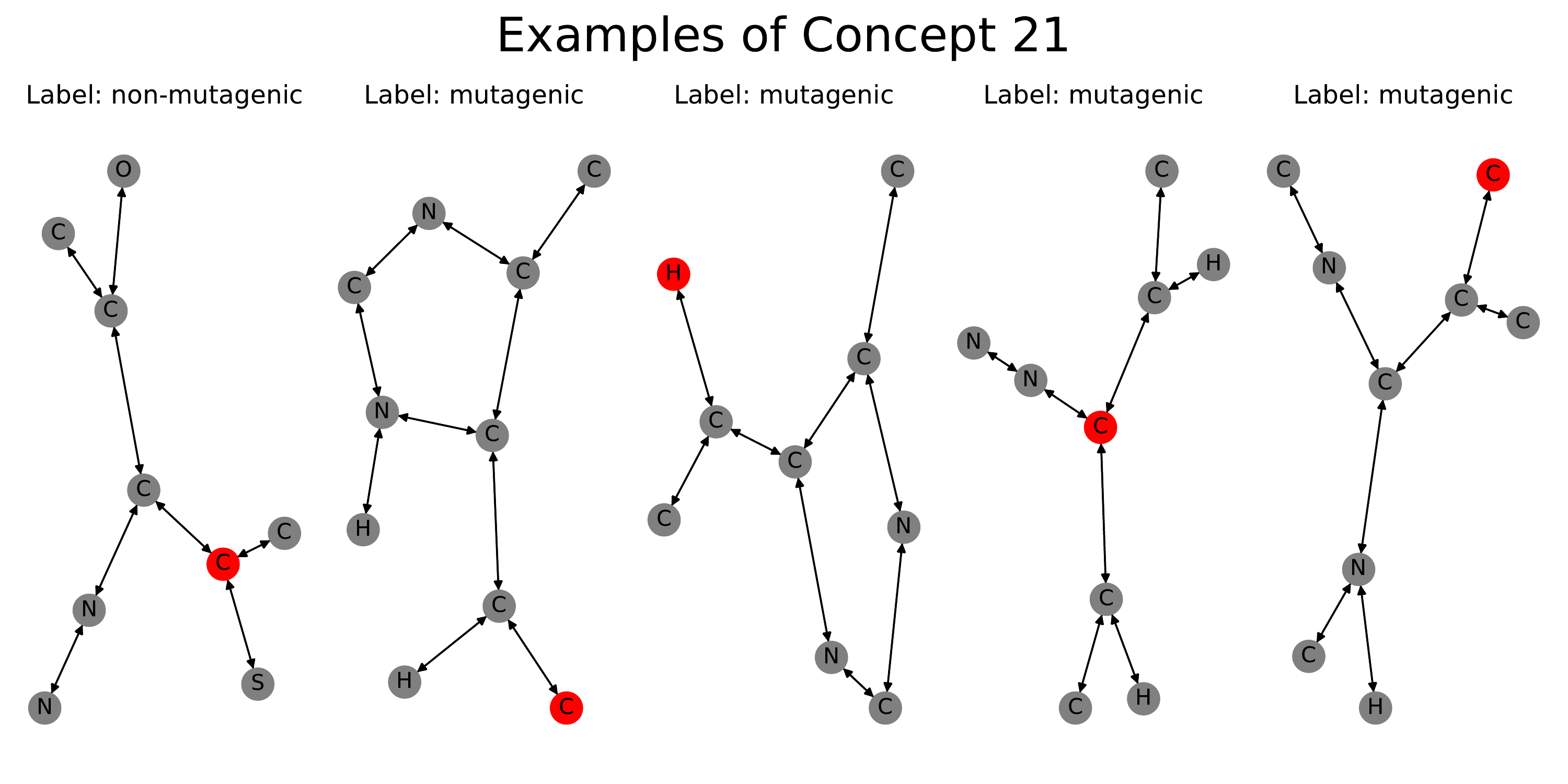}\\
    \includegraphics[width=0.45\textwidth]{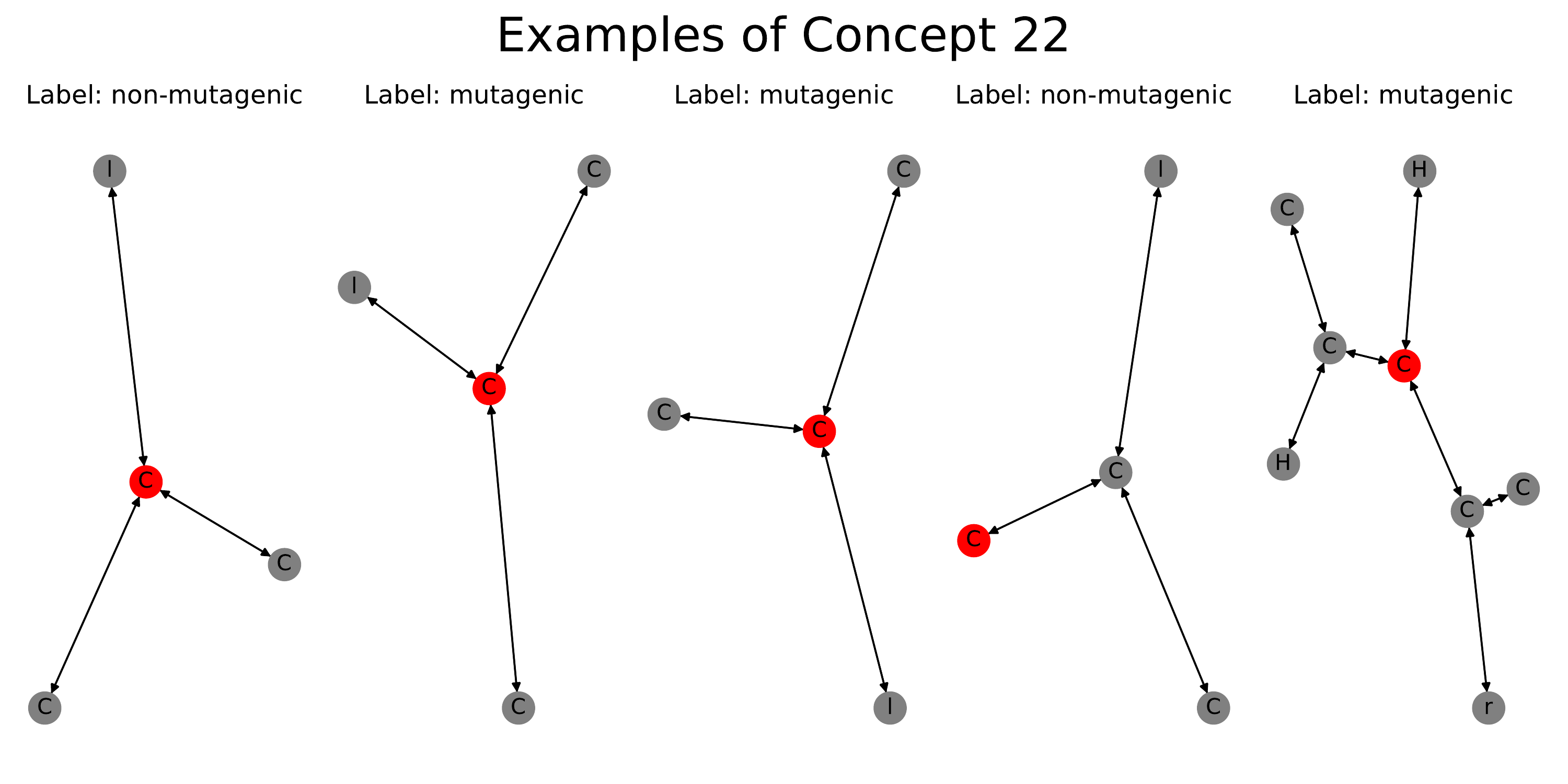}
    \includegraphics[width=0.45\textwidth]{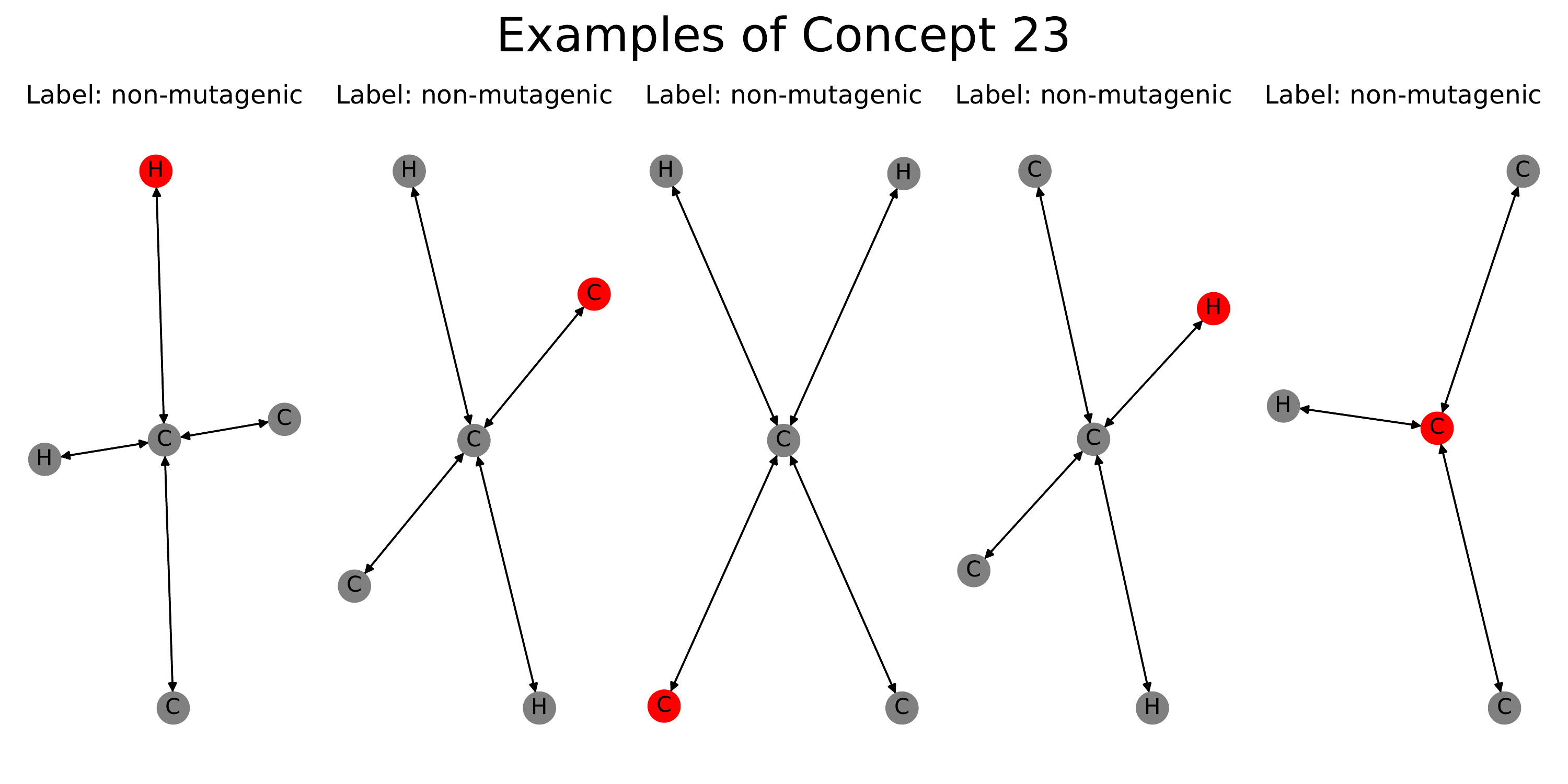}\\
    \includegraphics[width=0.45\textwidth]{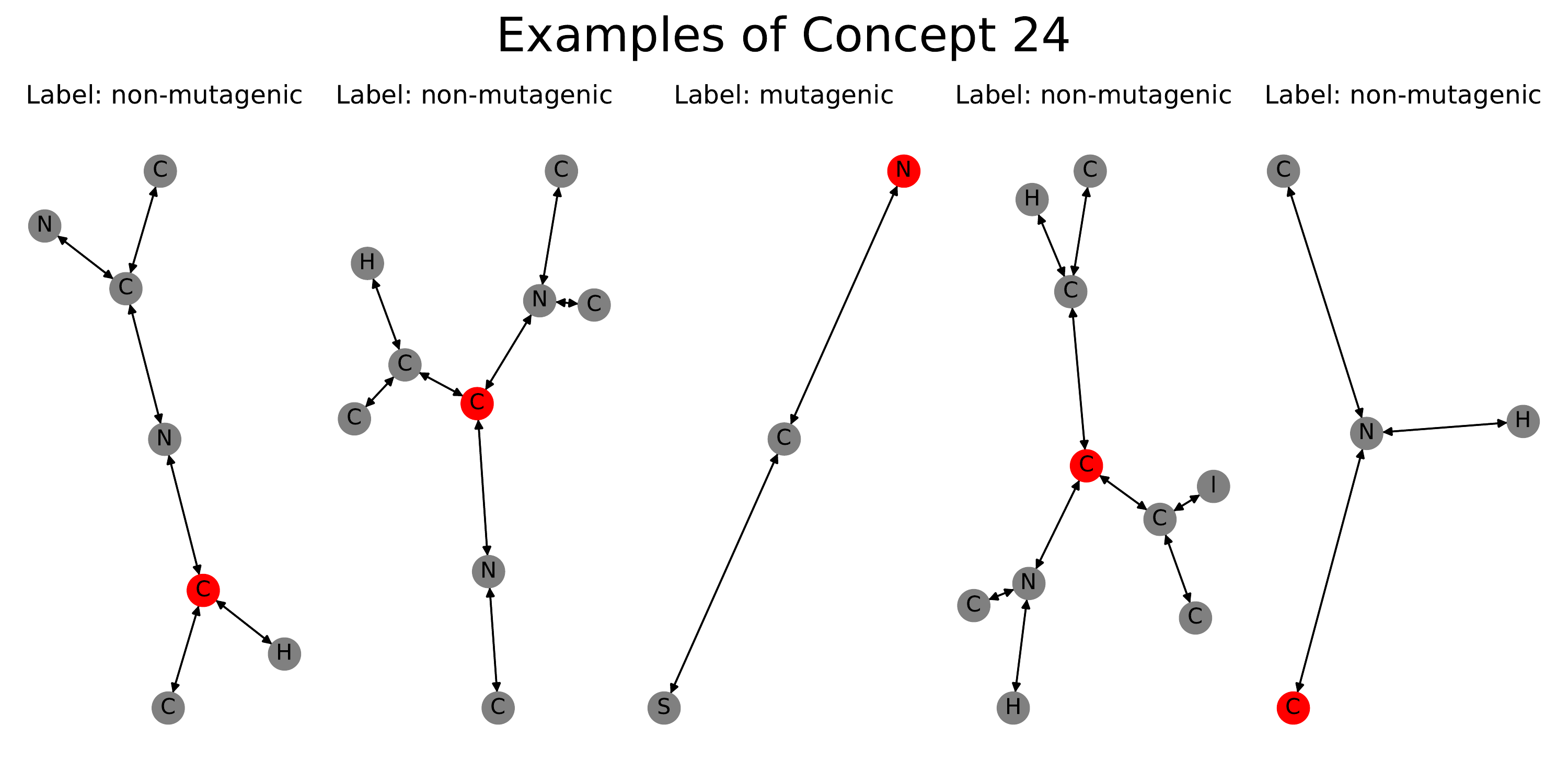}
    \includegraphics[width=0.45\textwidth]{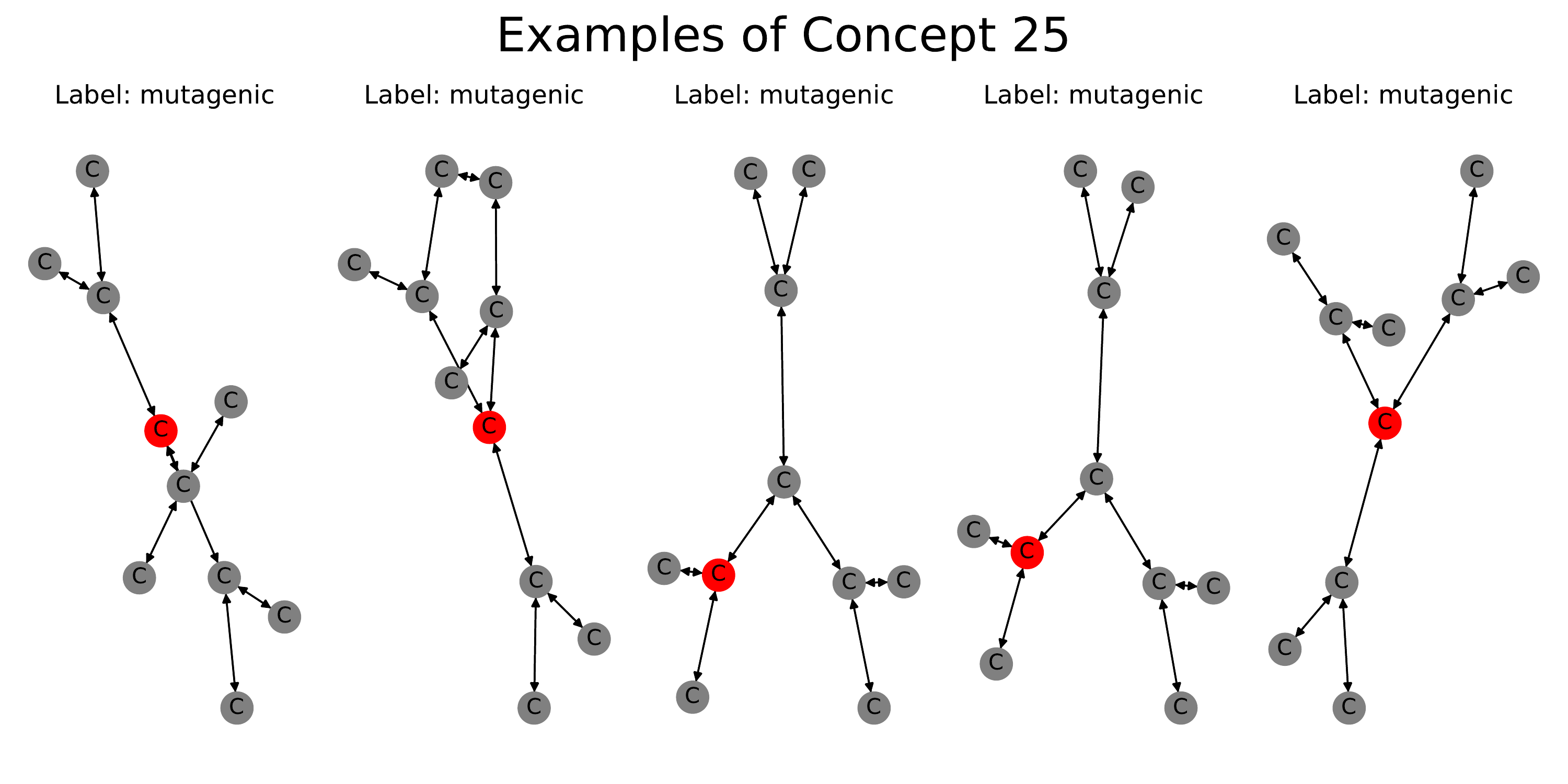}\\
    \includegraphics[width=0.45\textwidth]{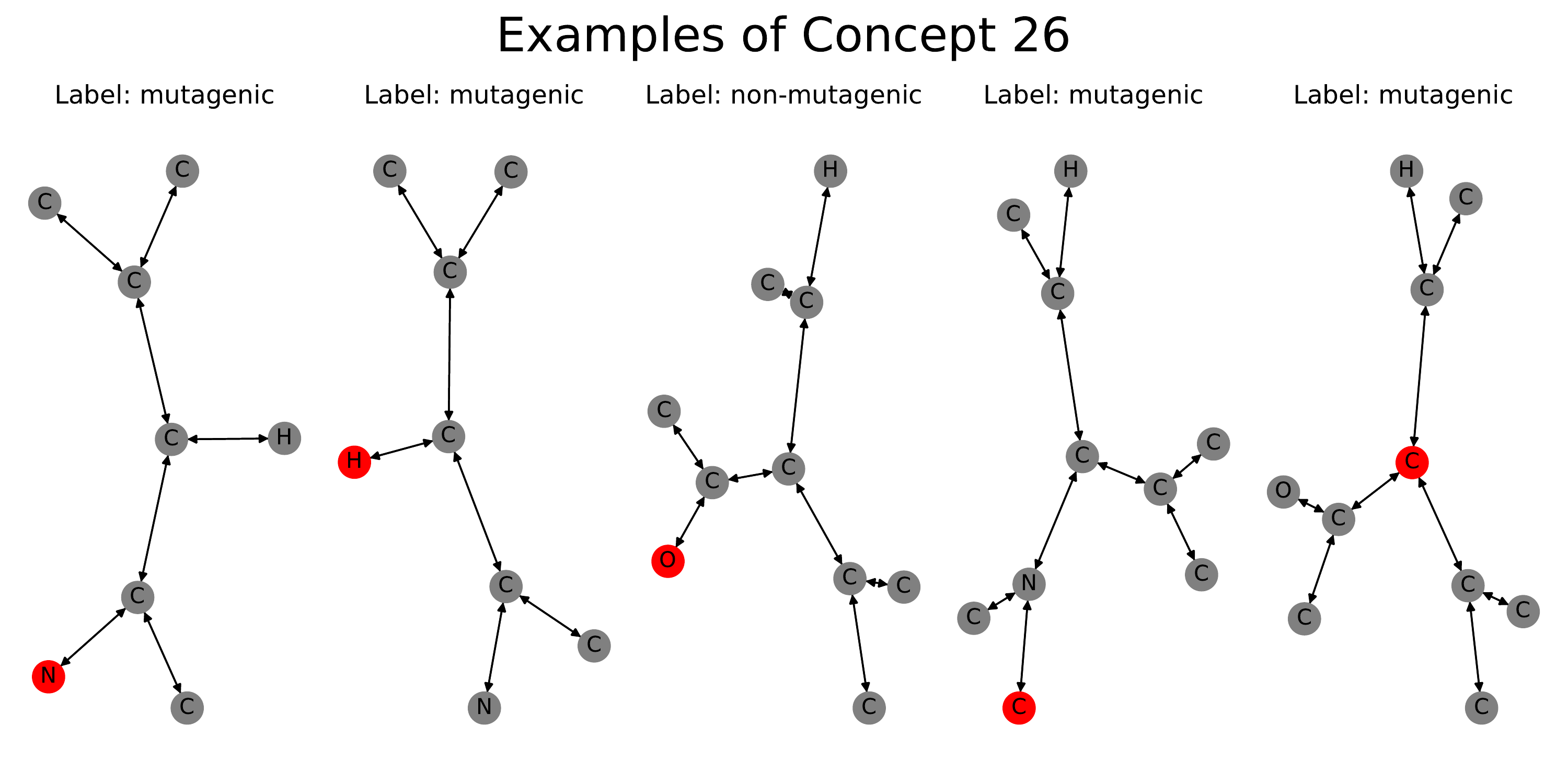}
    \includegraphics[width=0.45\textwidth]{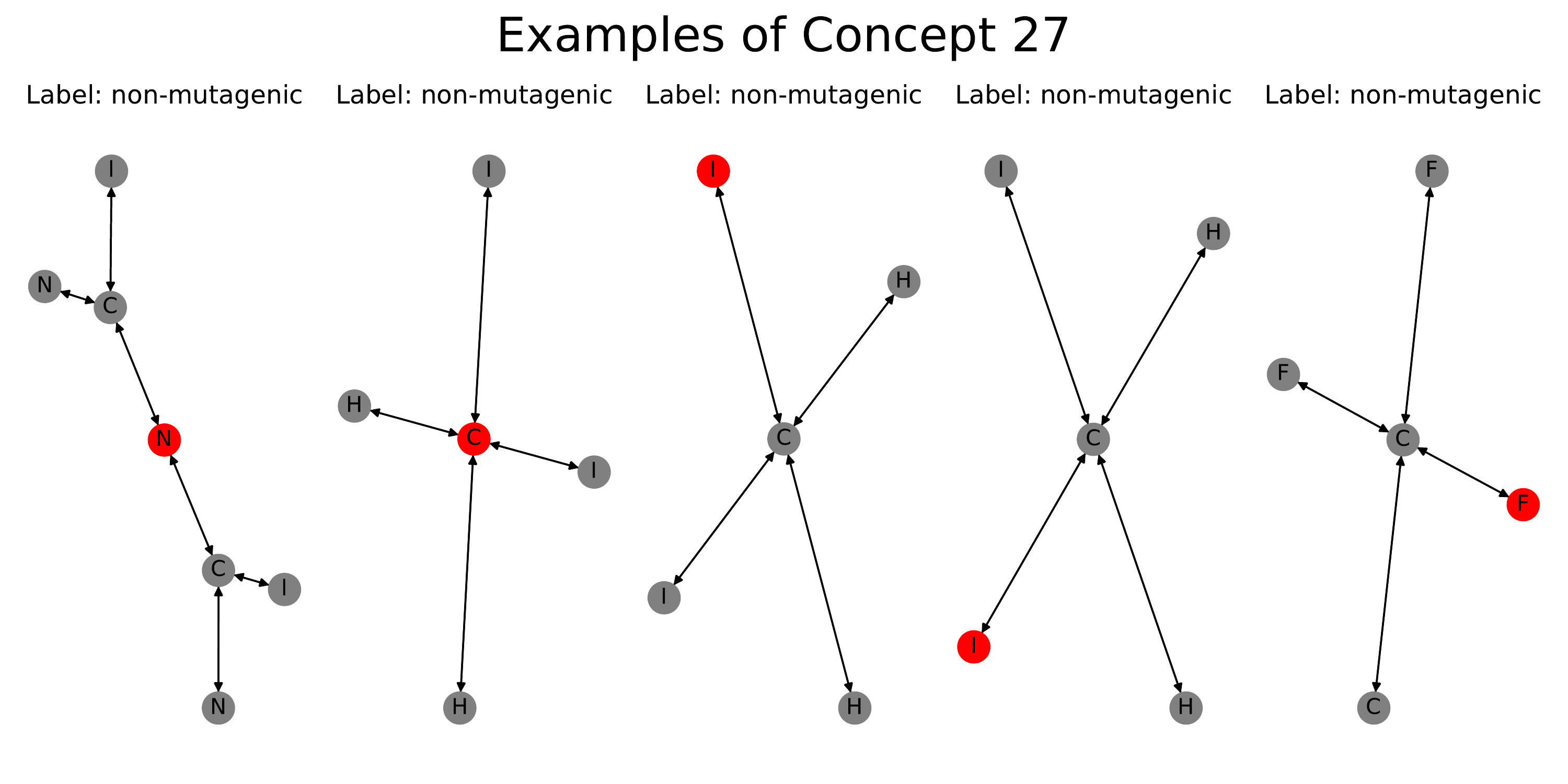}\\
    \includegraphics[width=0.45\textwidth]{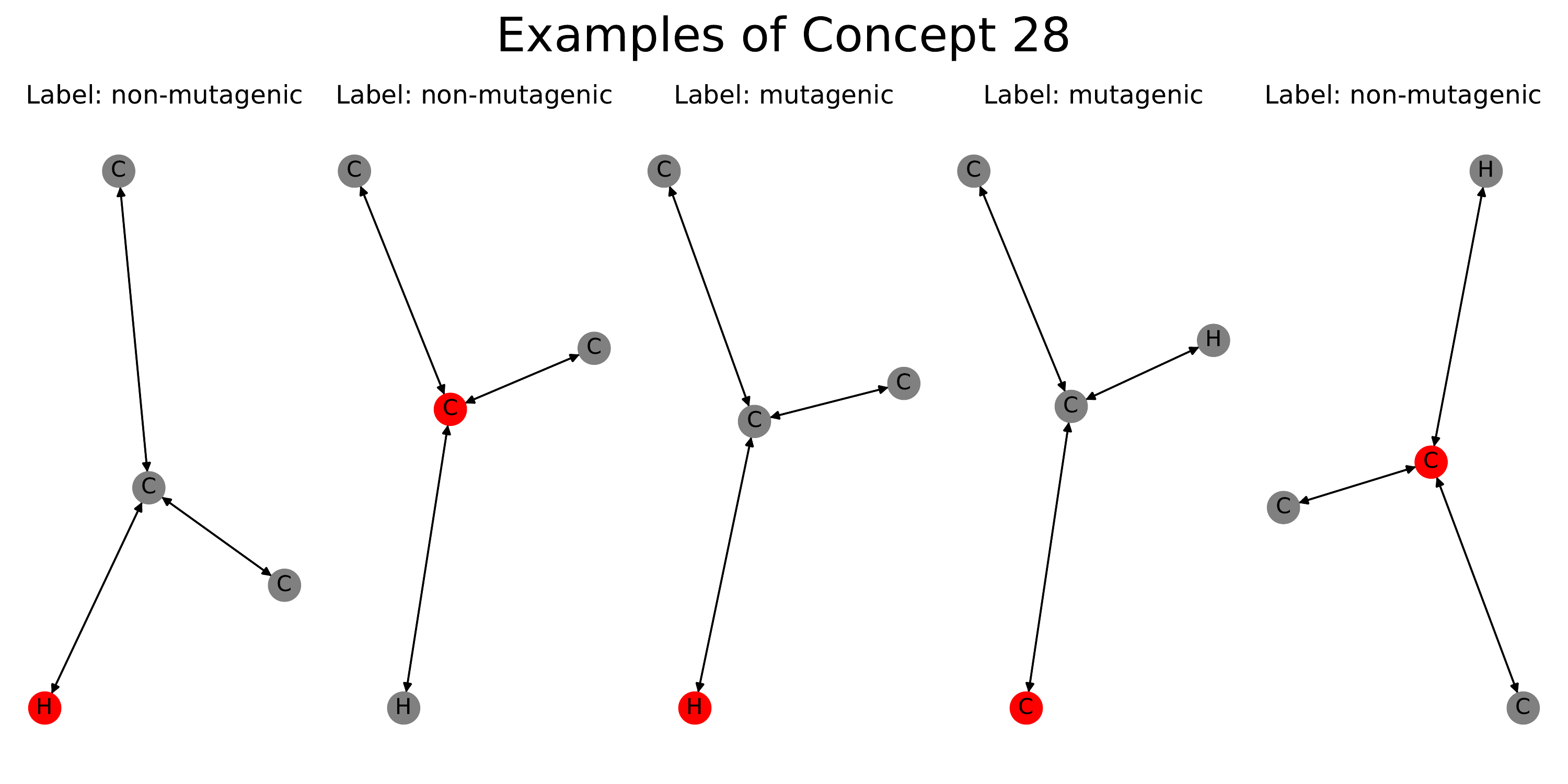}
    \includegraphics[width=0.45\textwidth]{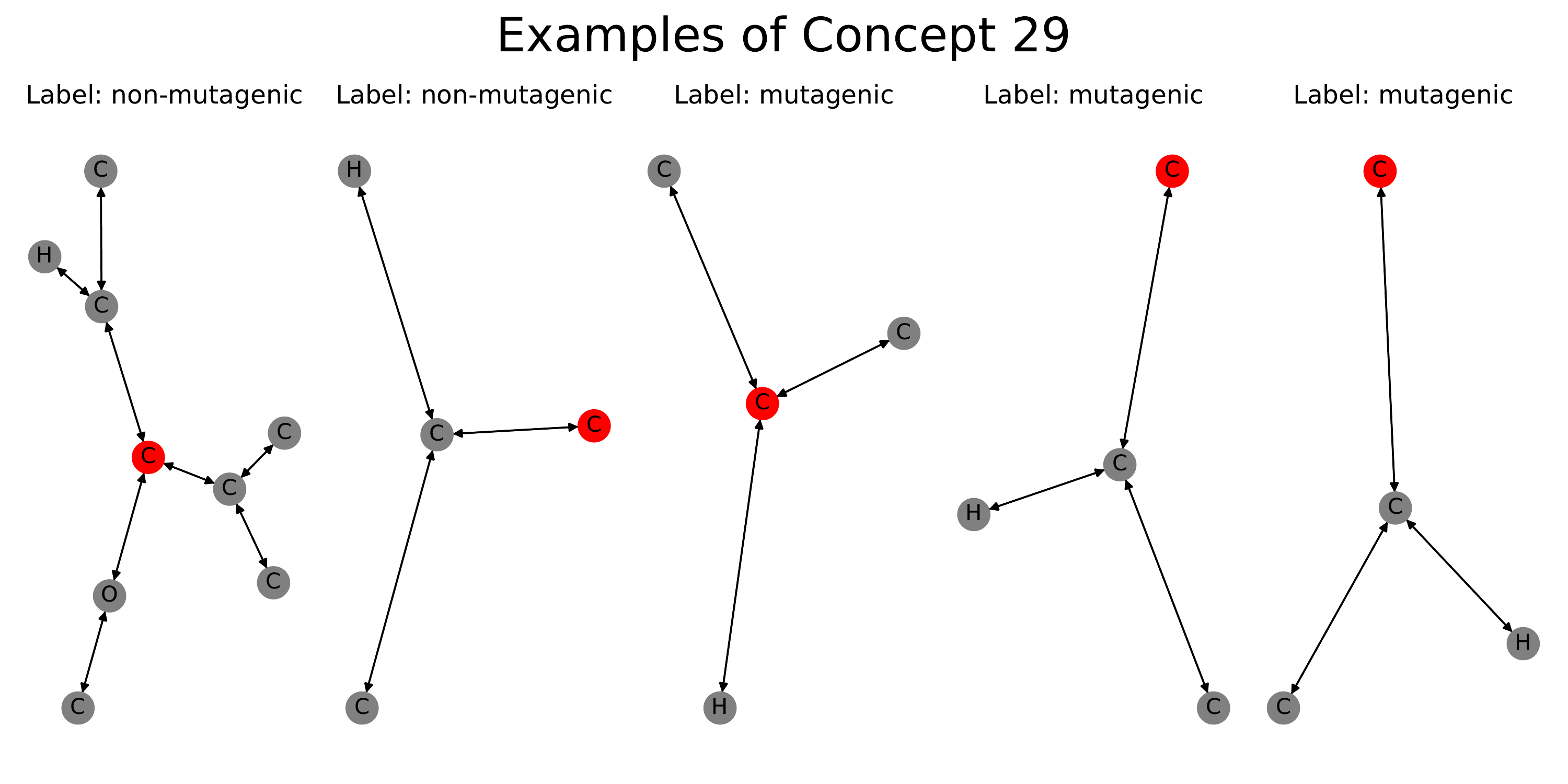}
    \caption{Concept discovered by the graph concept explainer. Part III.}
    \label{fig:mutag3}
\end{figure}

\begin{figure}[H]
    \centering
    \includegraphics[width=0.33\textwidth]{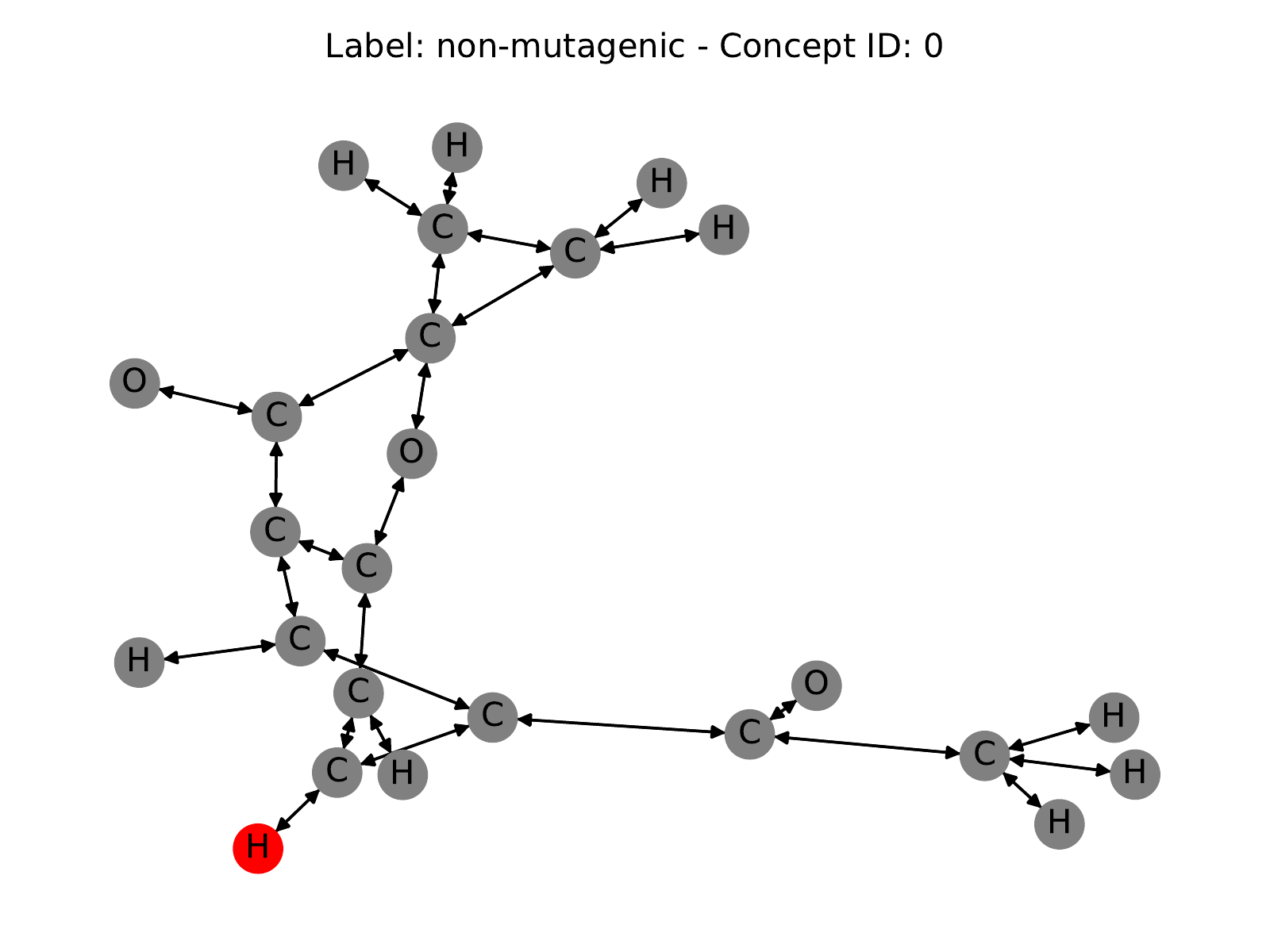}
    \includegraphics[width=0.33\textwidth]{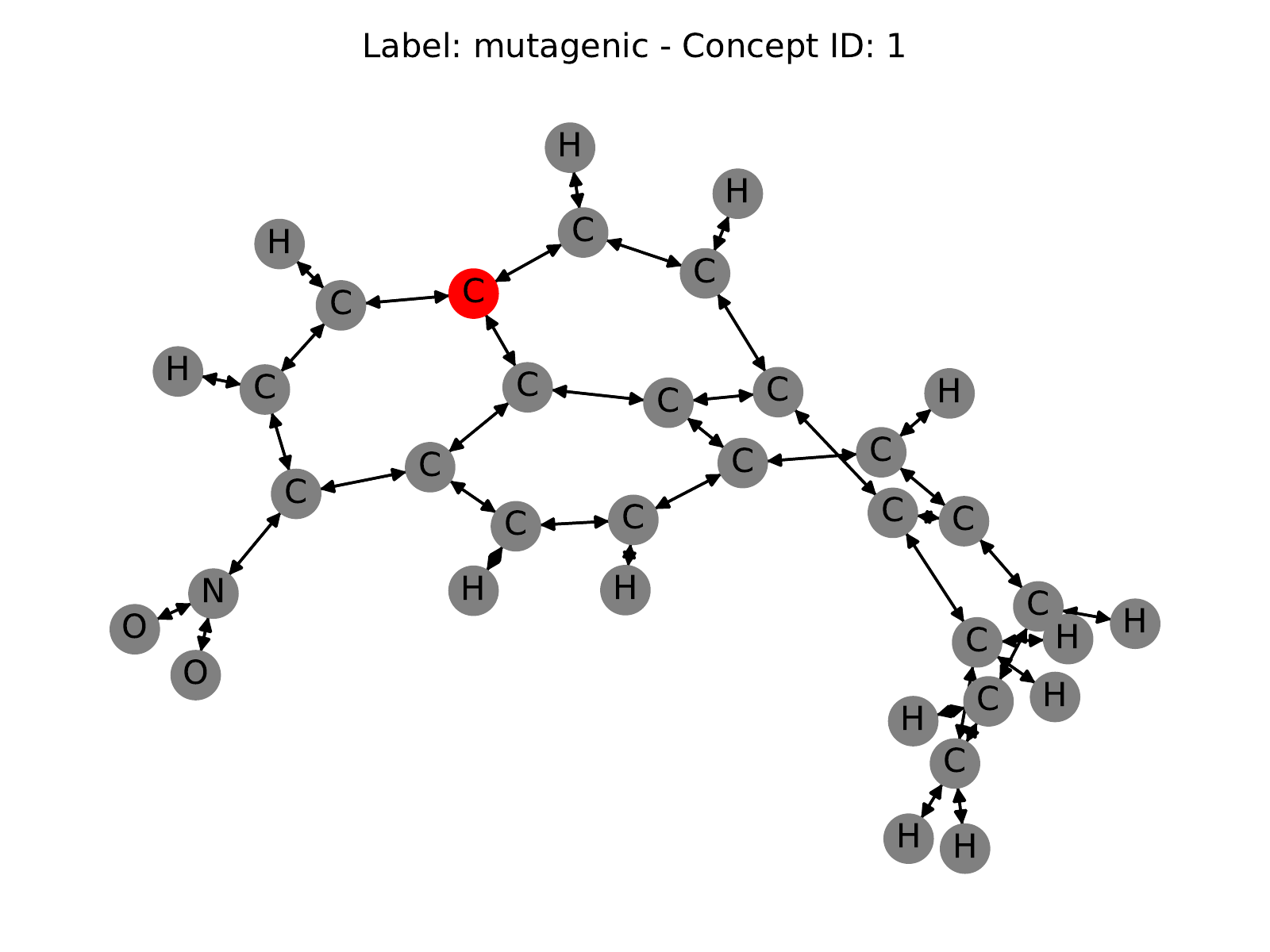}
    \includegraphics[width=0.33\textwidth]{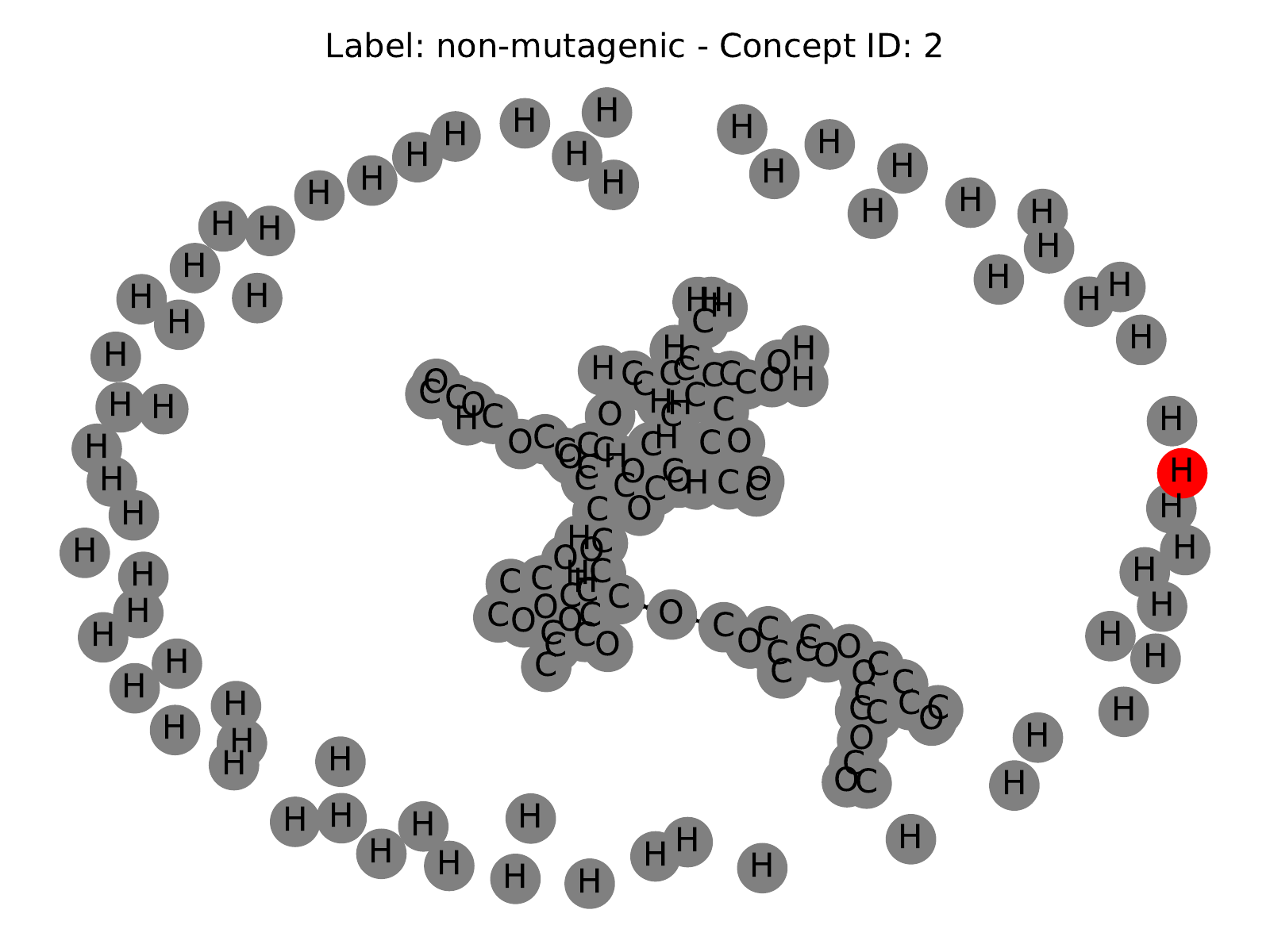}\\
    \includegraphics[width=0.33\textwidth]{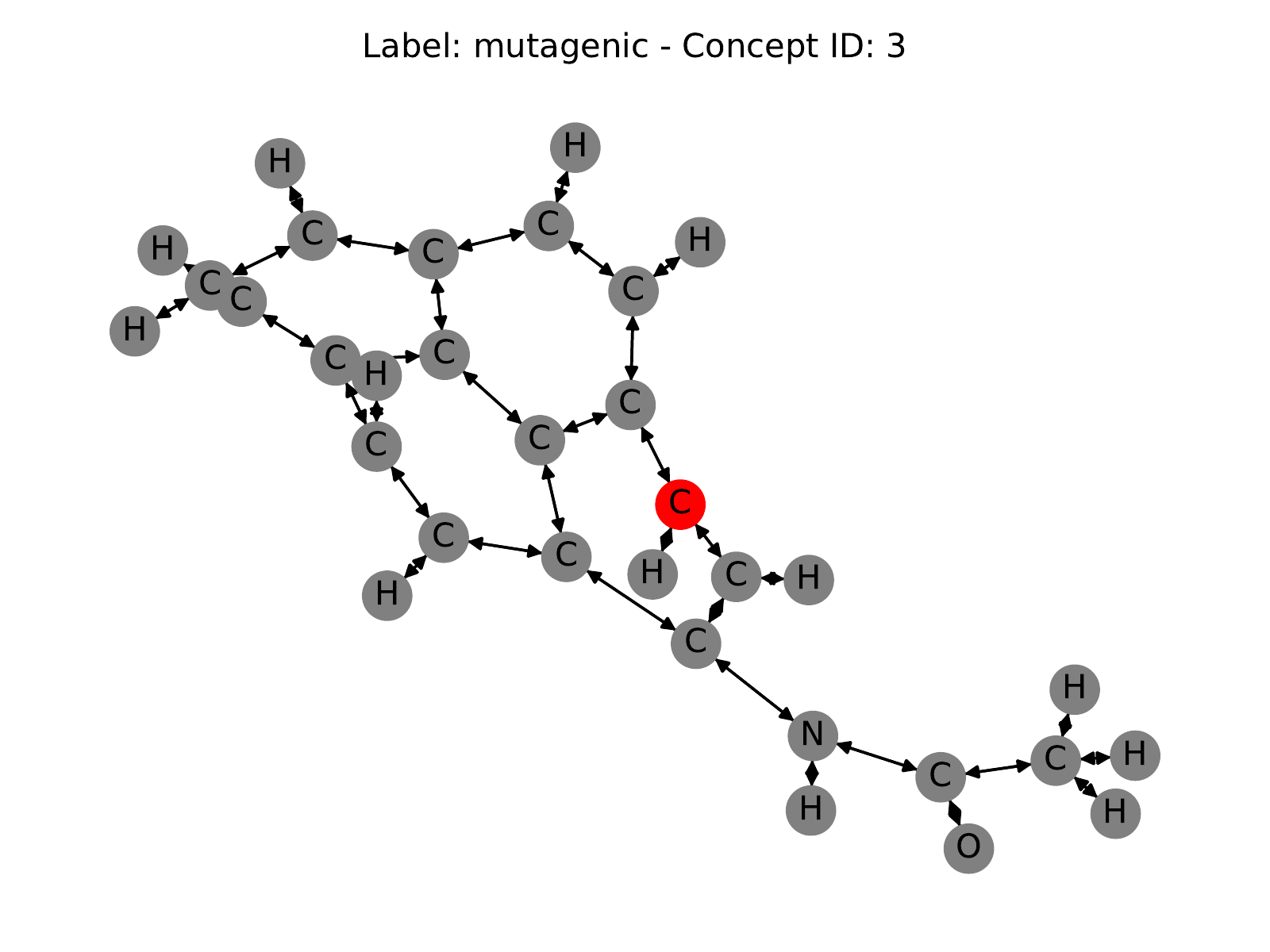}
    \includegraphics[width=0.33\textwidth]{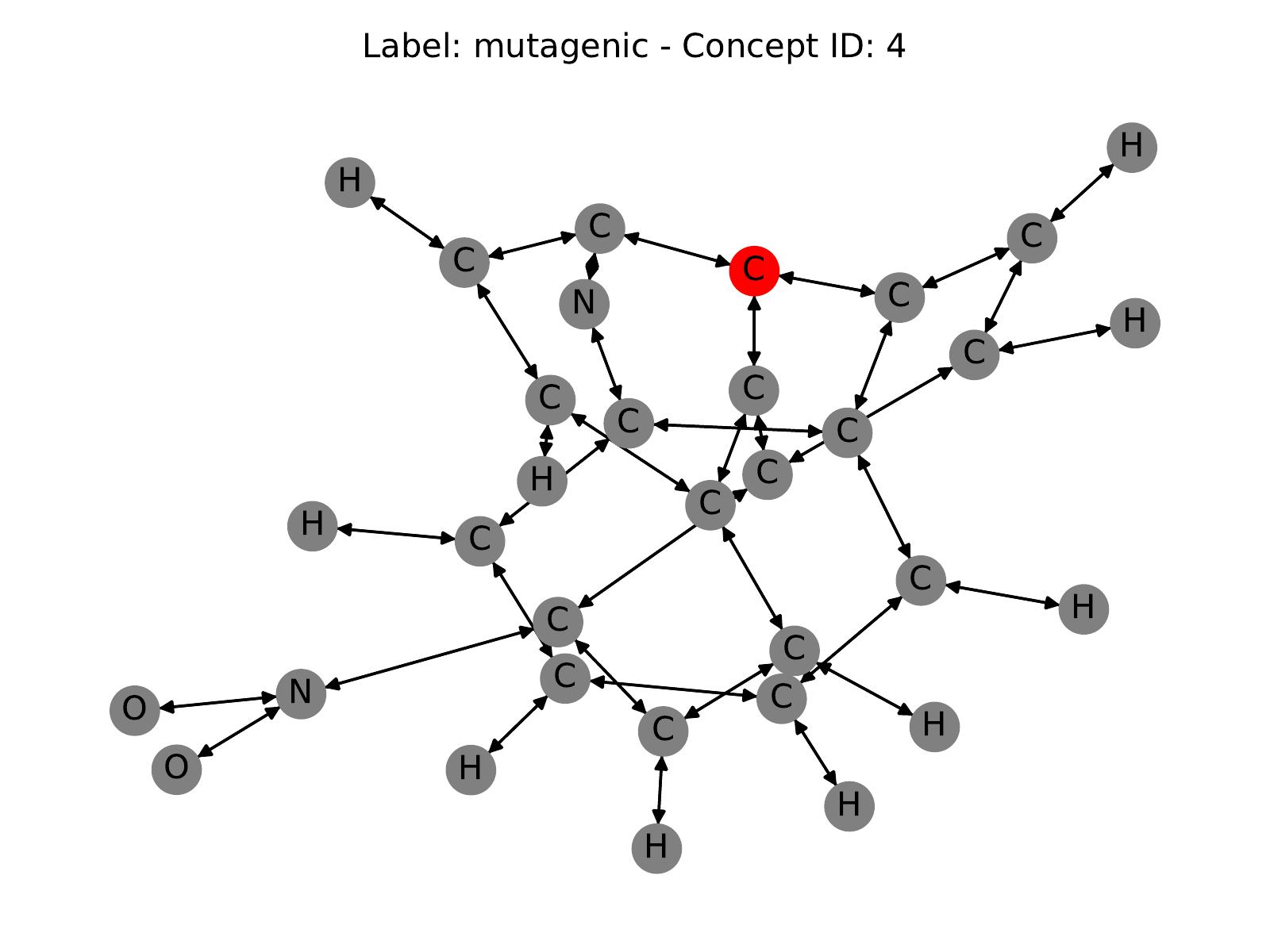}
    \includegraphics[width=0.33\textwidth]{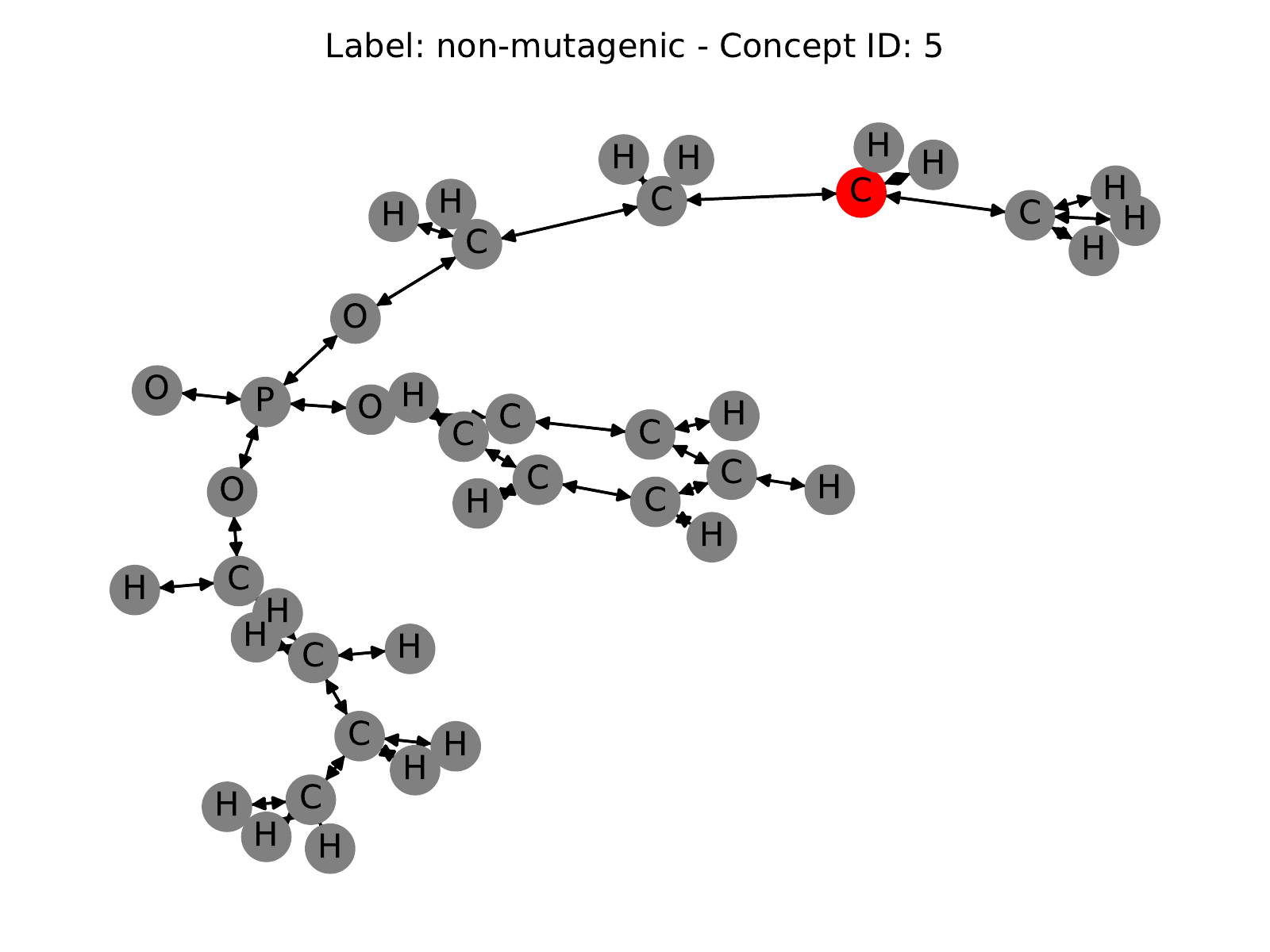}\\
    \includegraphics[width=0.33\textwidth]{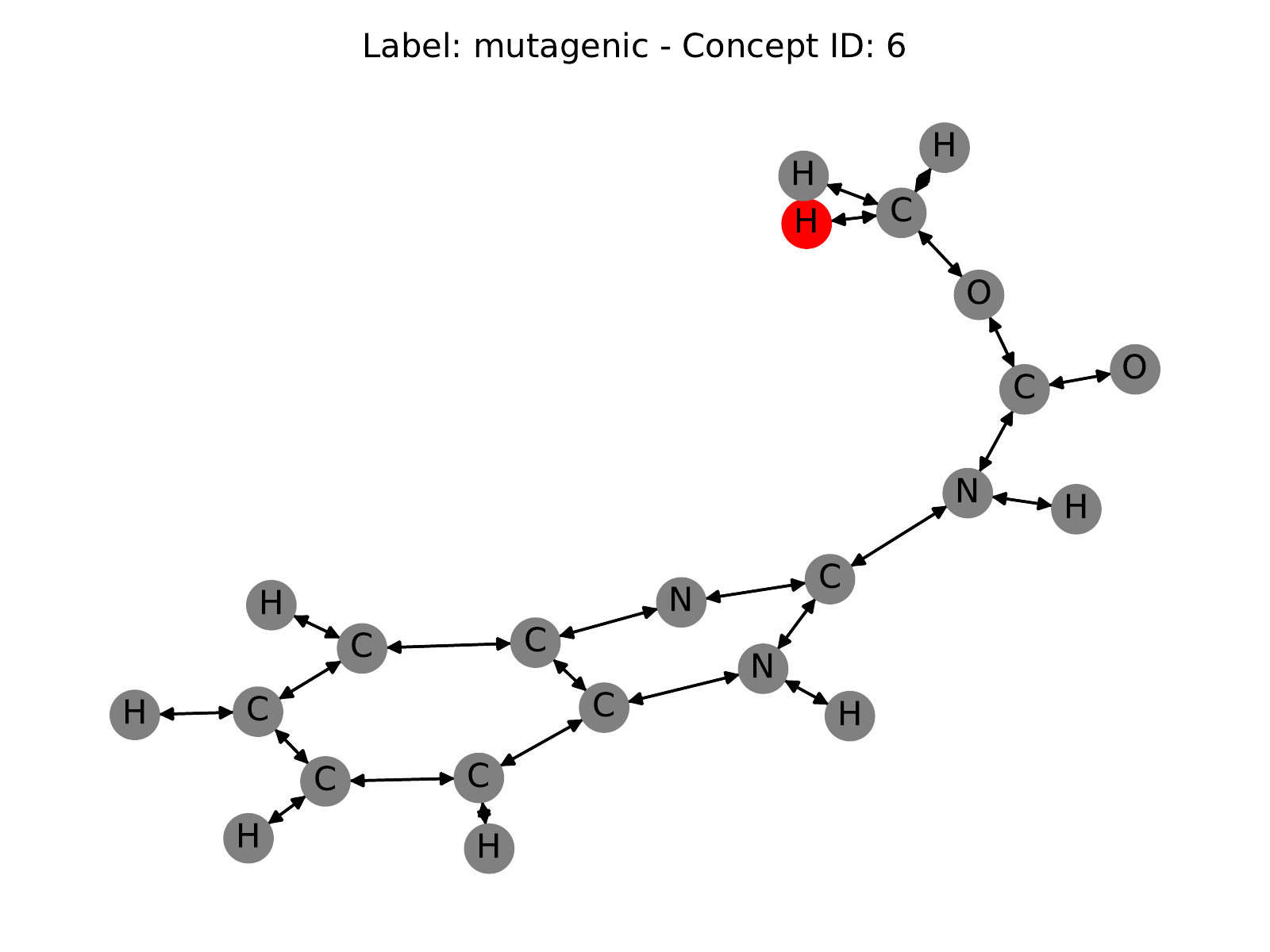}
    \includegraphics[width=0.33\textwidth]{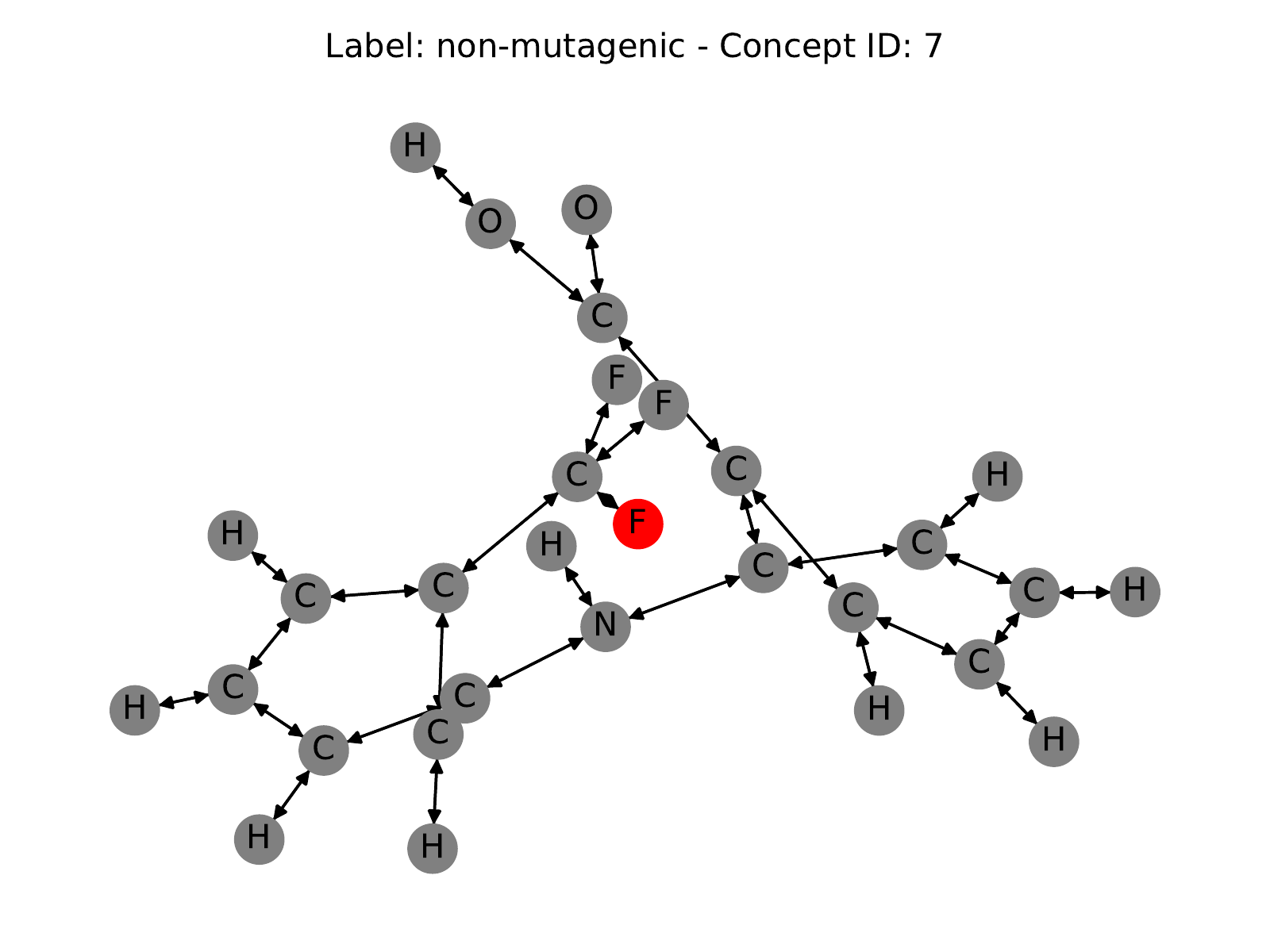}
    \includegraphics[width=0.33\textwidth]{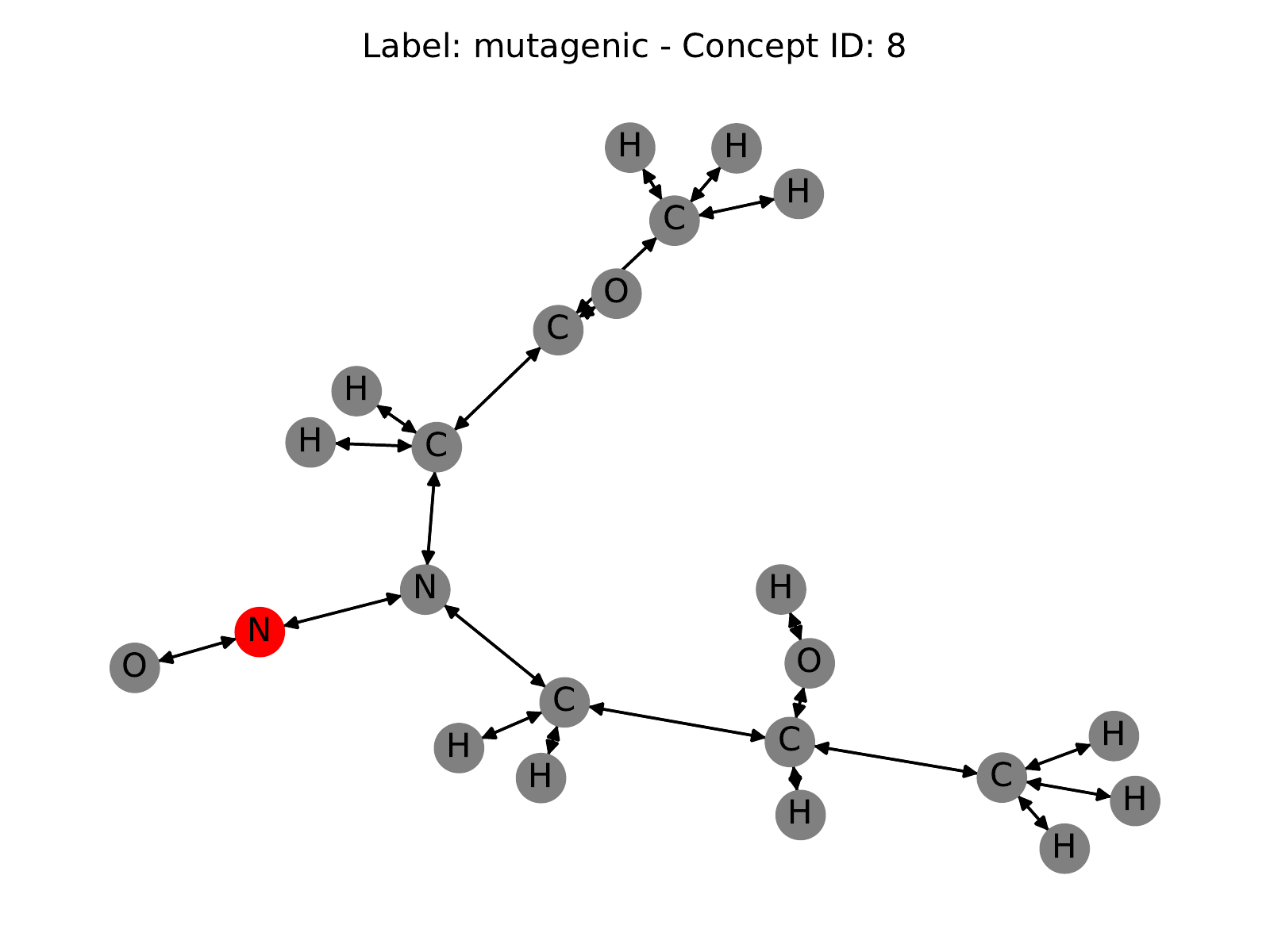}\\
    \includegraphics[width=0.33\textwidth]{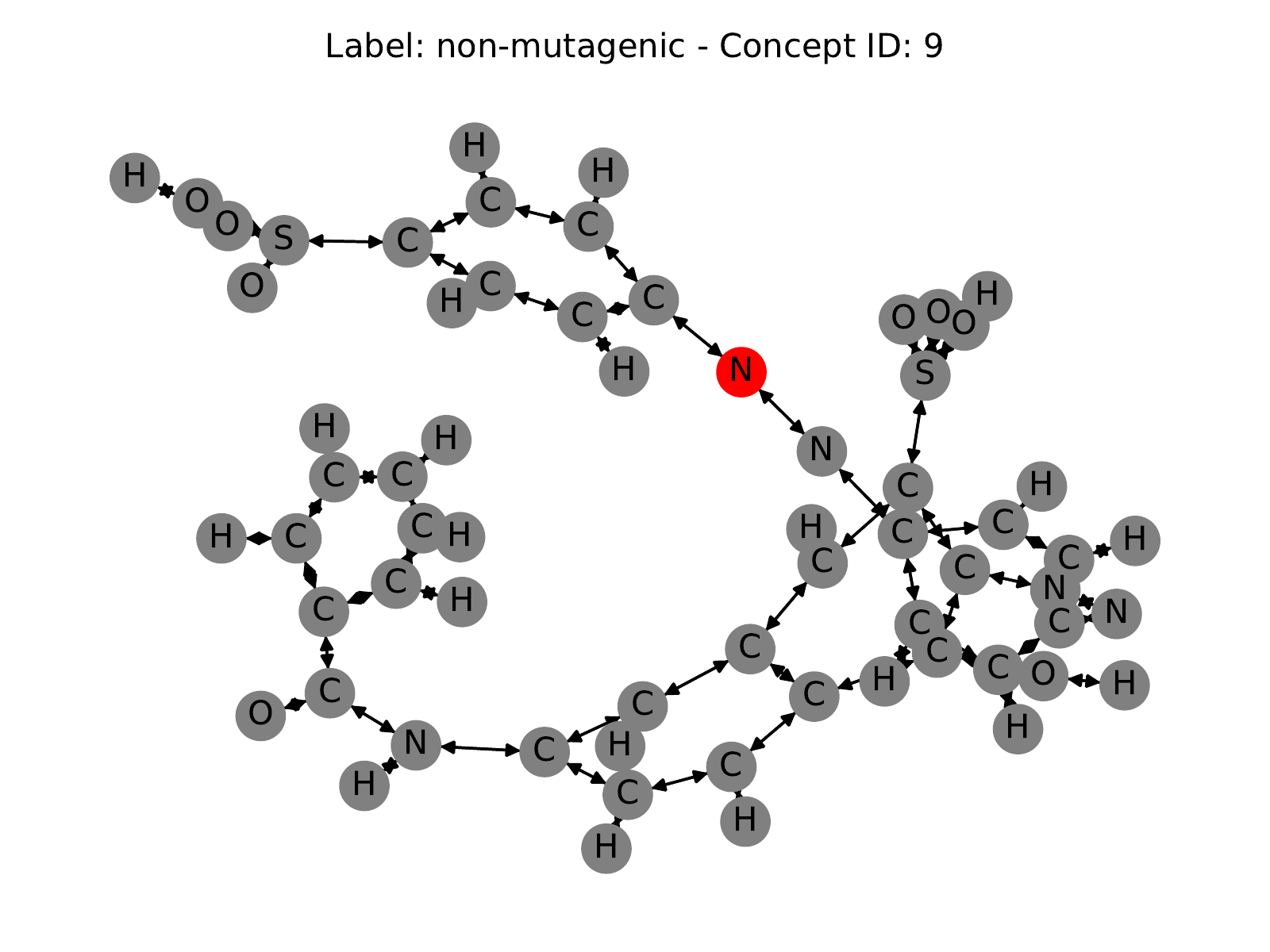}
    \includegraphics[width=0.33\textwidth]{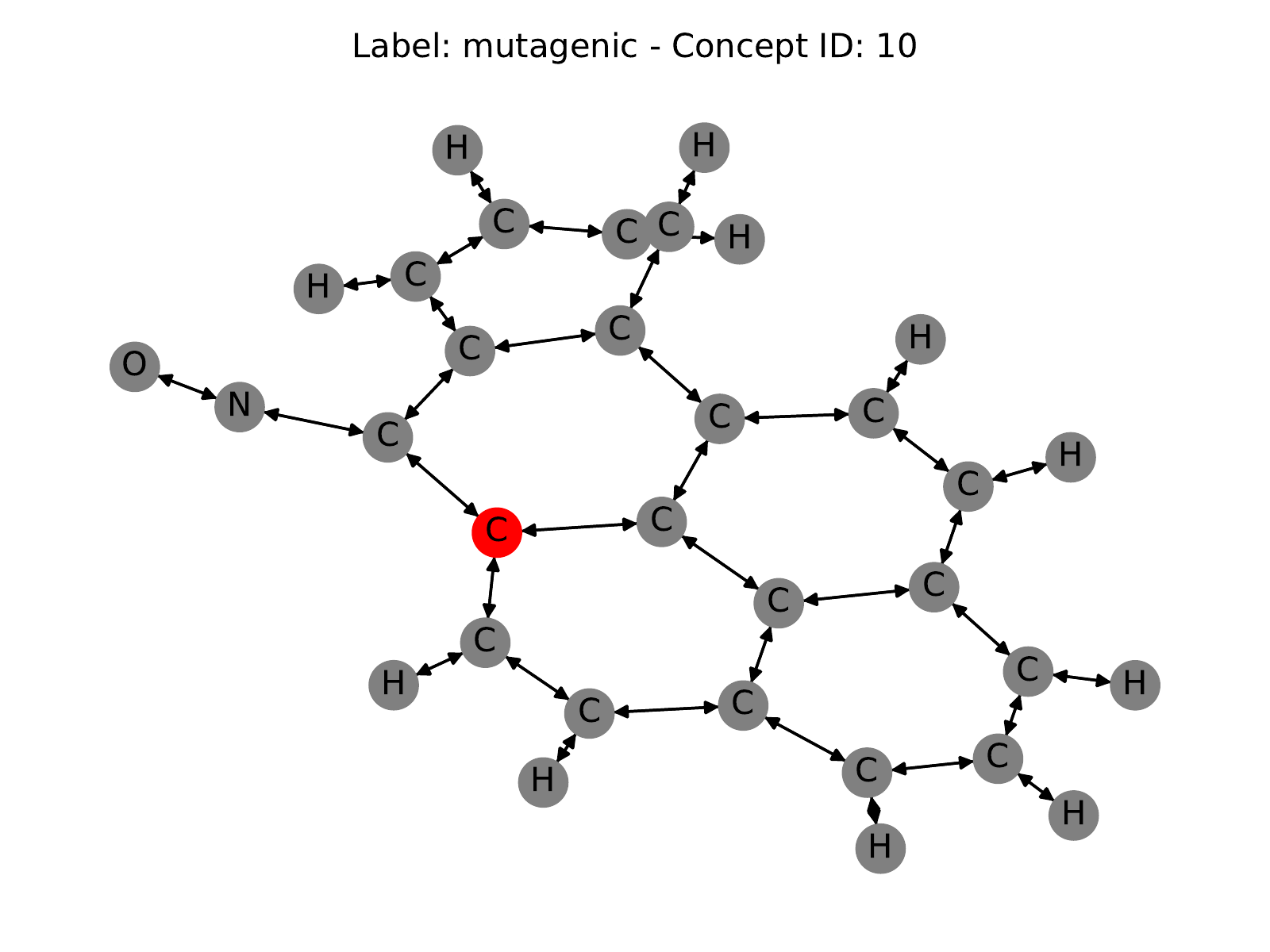}
    \includegraphics[width=0.33\textwidth]{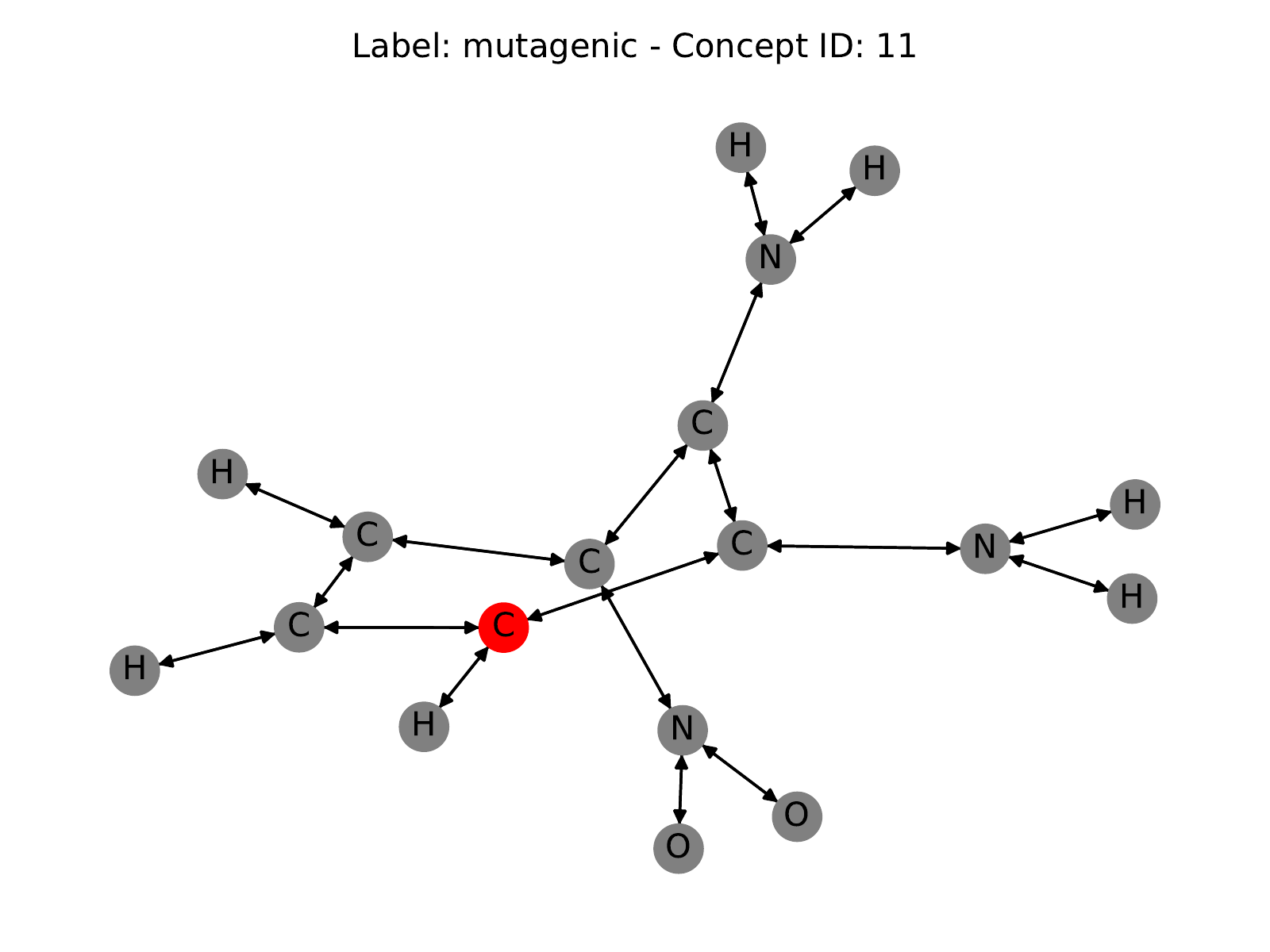}\\
    \includegraphics[width=0.33\textwidth]{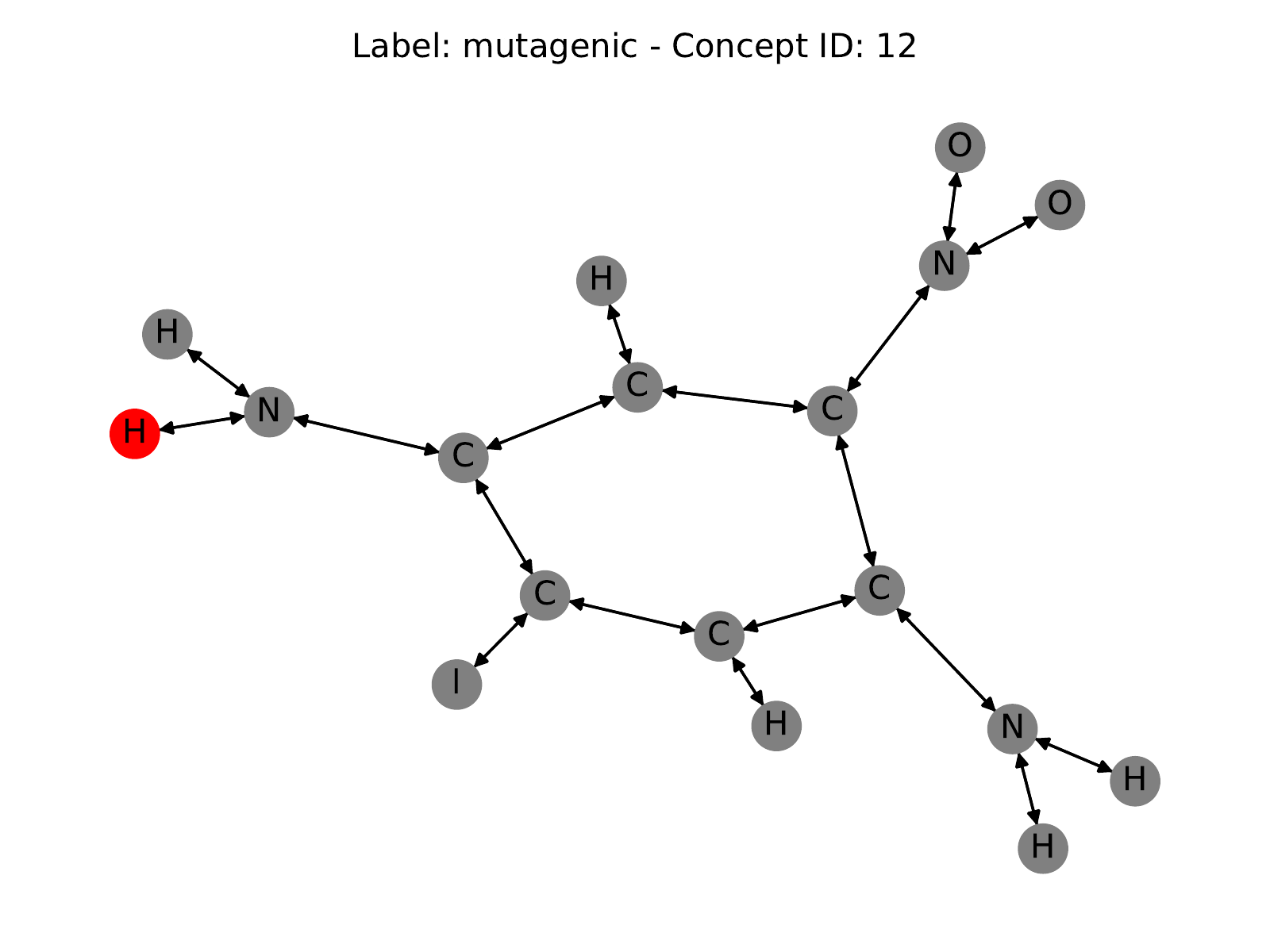}
    \includegraphics[width=0.33\textwidth]{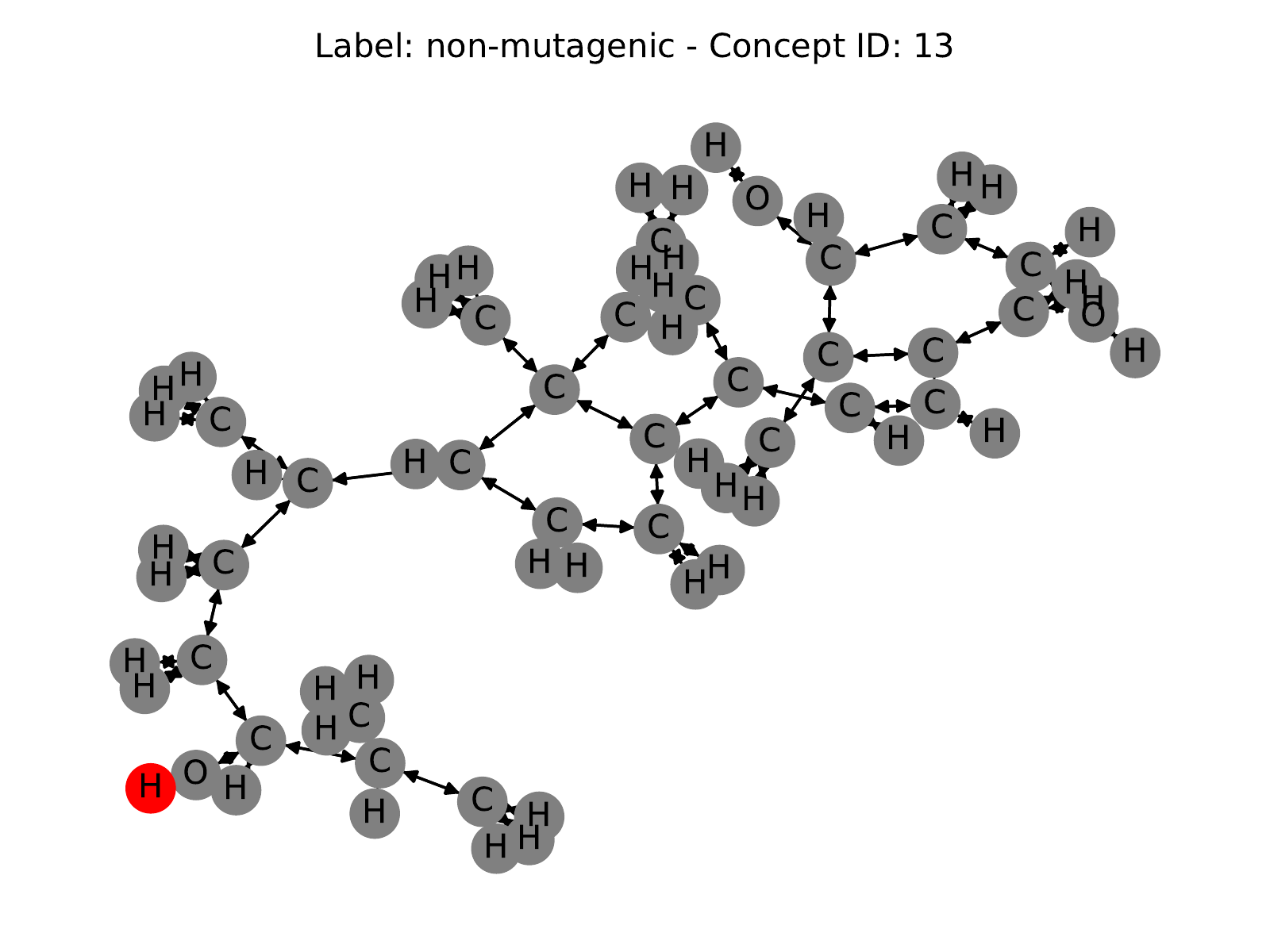}
    \includegraphics[width=0.33\textwidth]{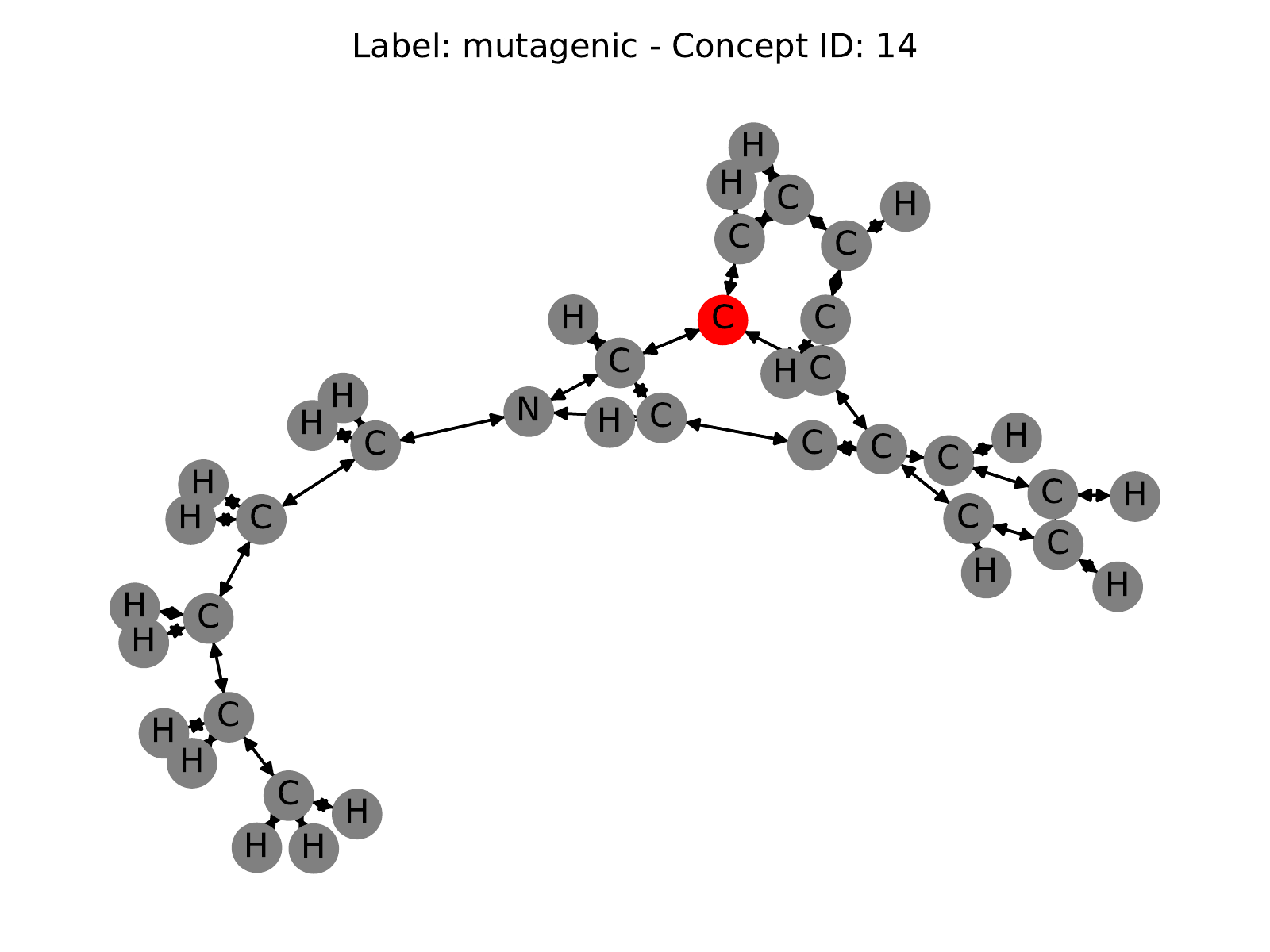}
    \caption{Full molecule corresponding to the closest node embedding to the concept centroid. Part I.}
    \label{fig:mutag5}
\end{figure}

\begin{figure}[H]
    \centering
    \includegraphics[width=0.33\textwidth]{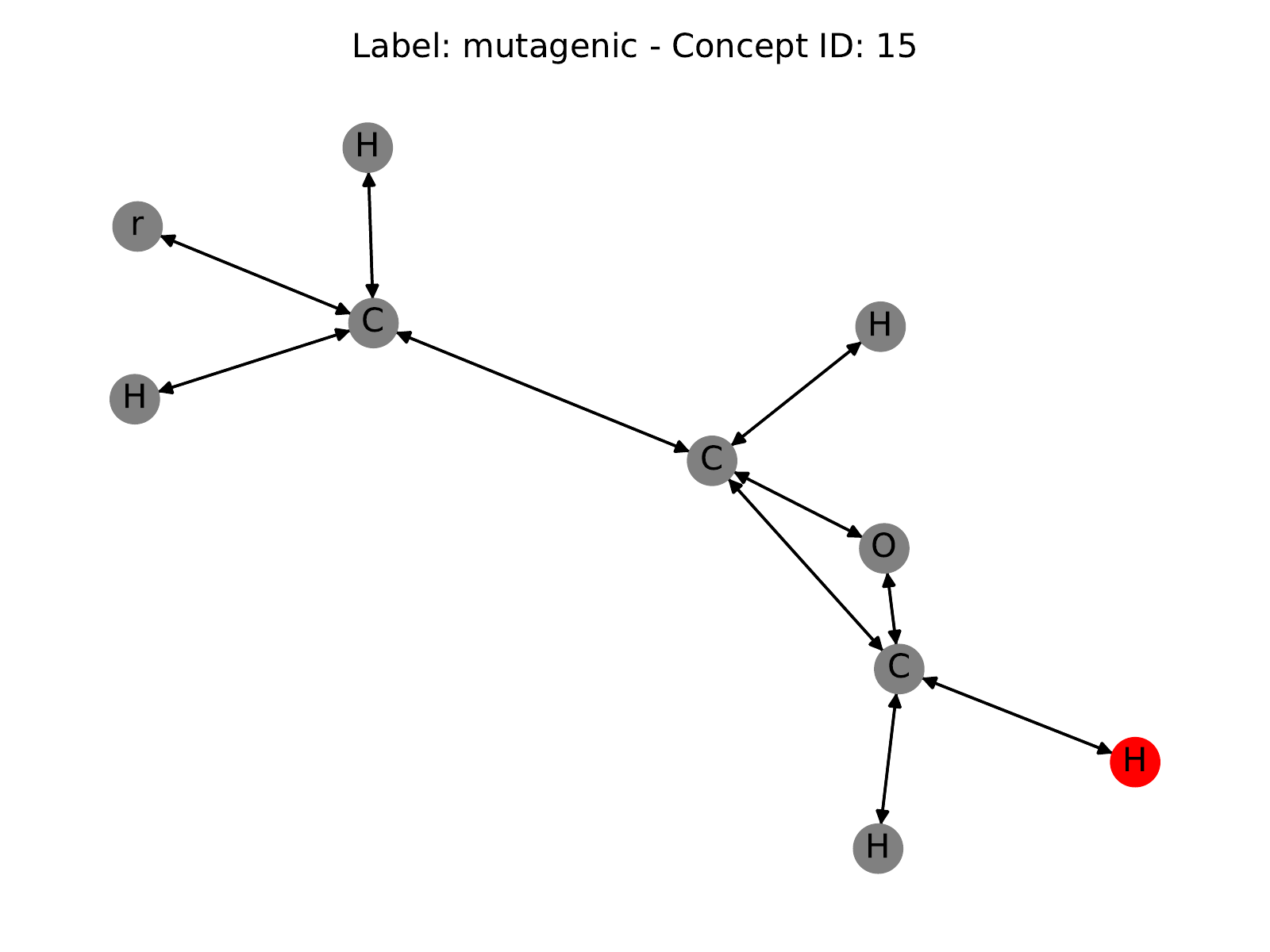}
    \includegraphics[width=0.33\textwidth]{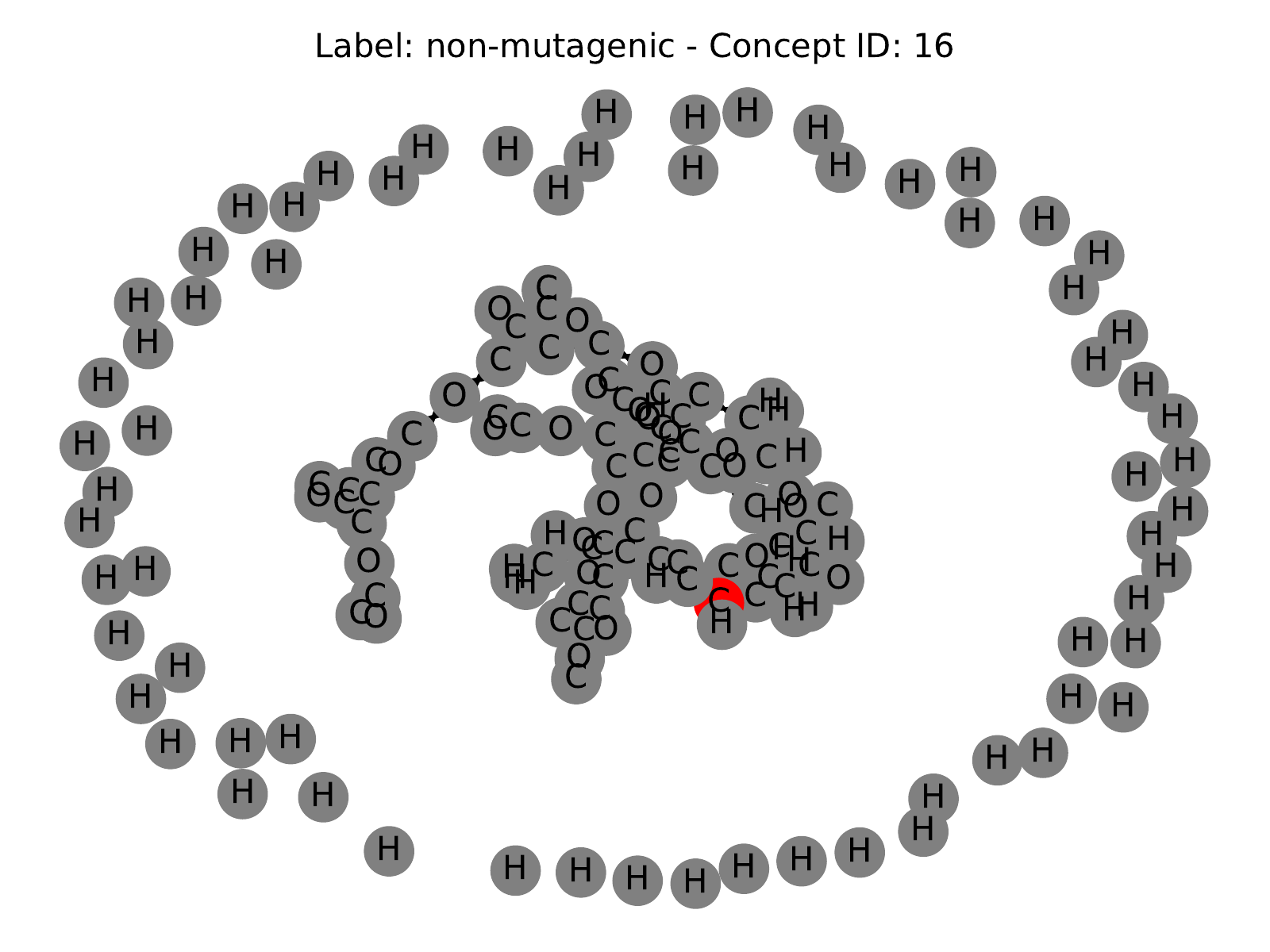}
    \includegraphics[width=0.33\textwidth]{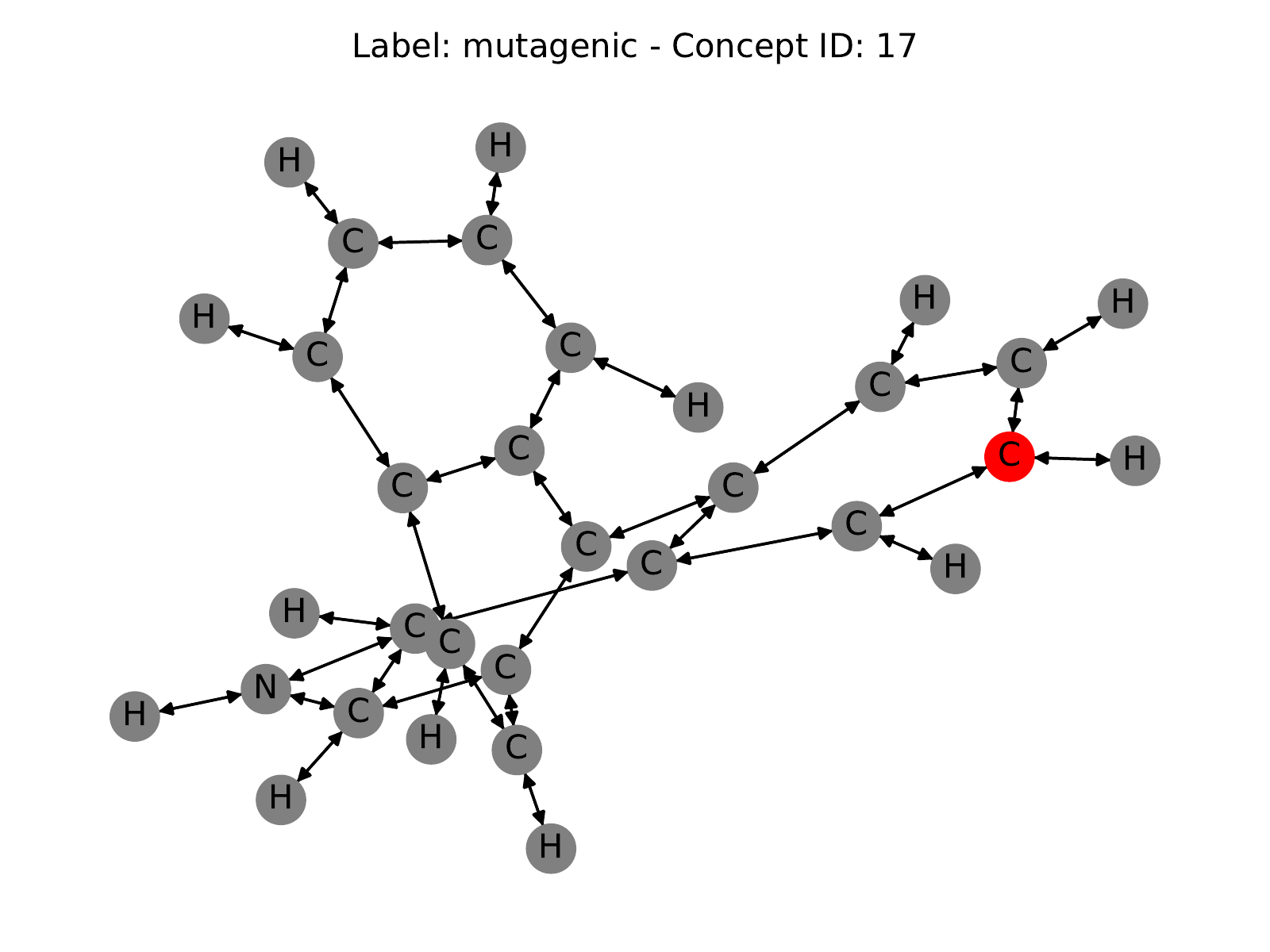}\\
    \includegraphics[width=0.33\textwidth]{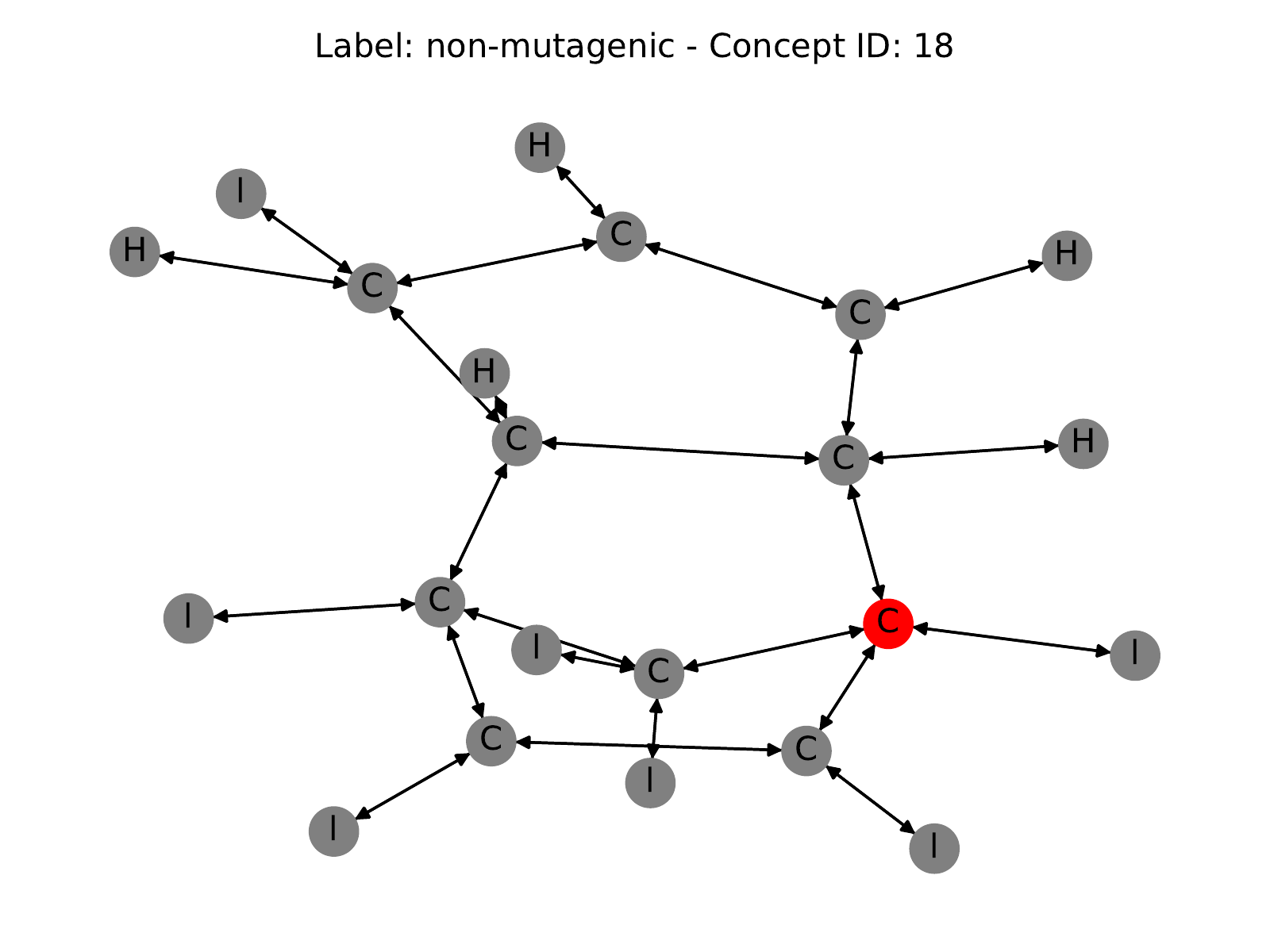}
    \includegraphics[width=0.33\textwidth]{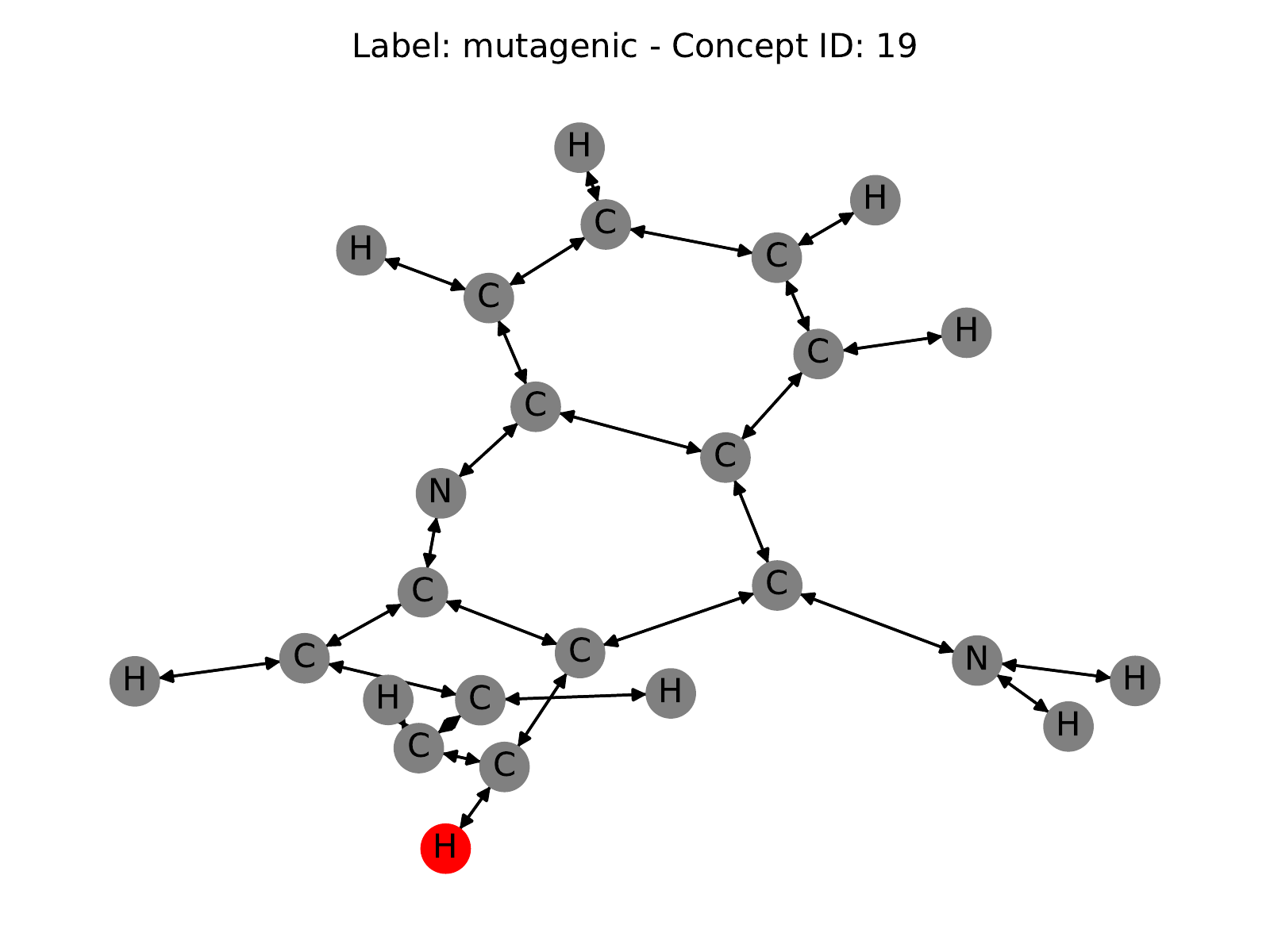}
    \includegraphics[width=0.33\textwidth]{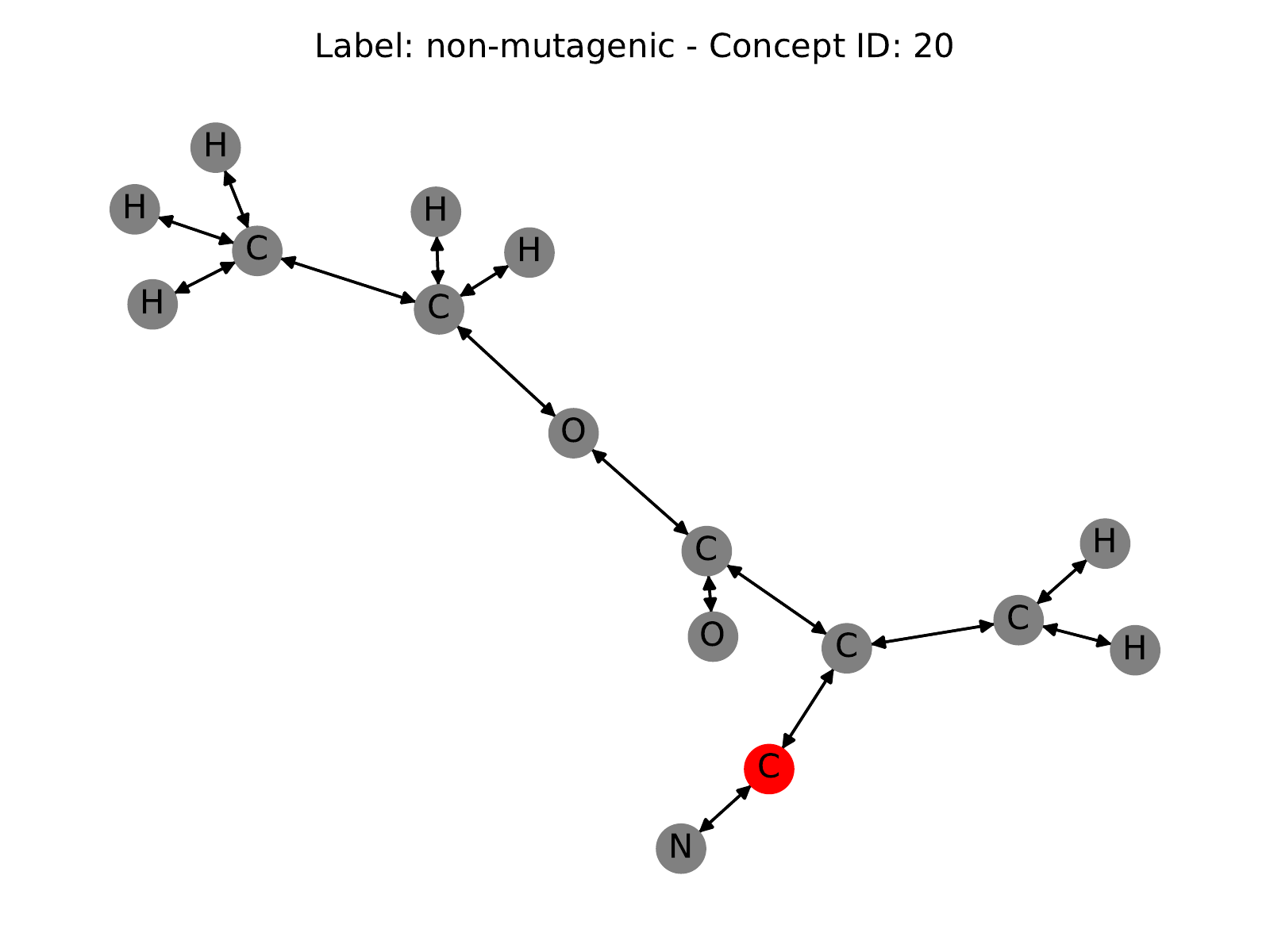}\\
    \includegraphics[width=0.33\textwidth]{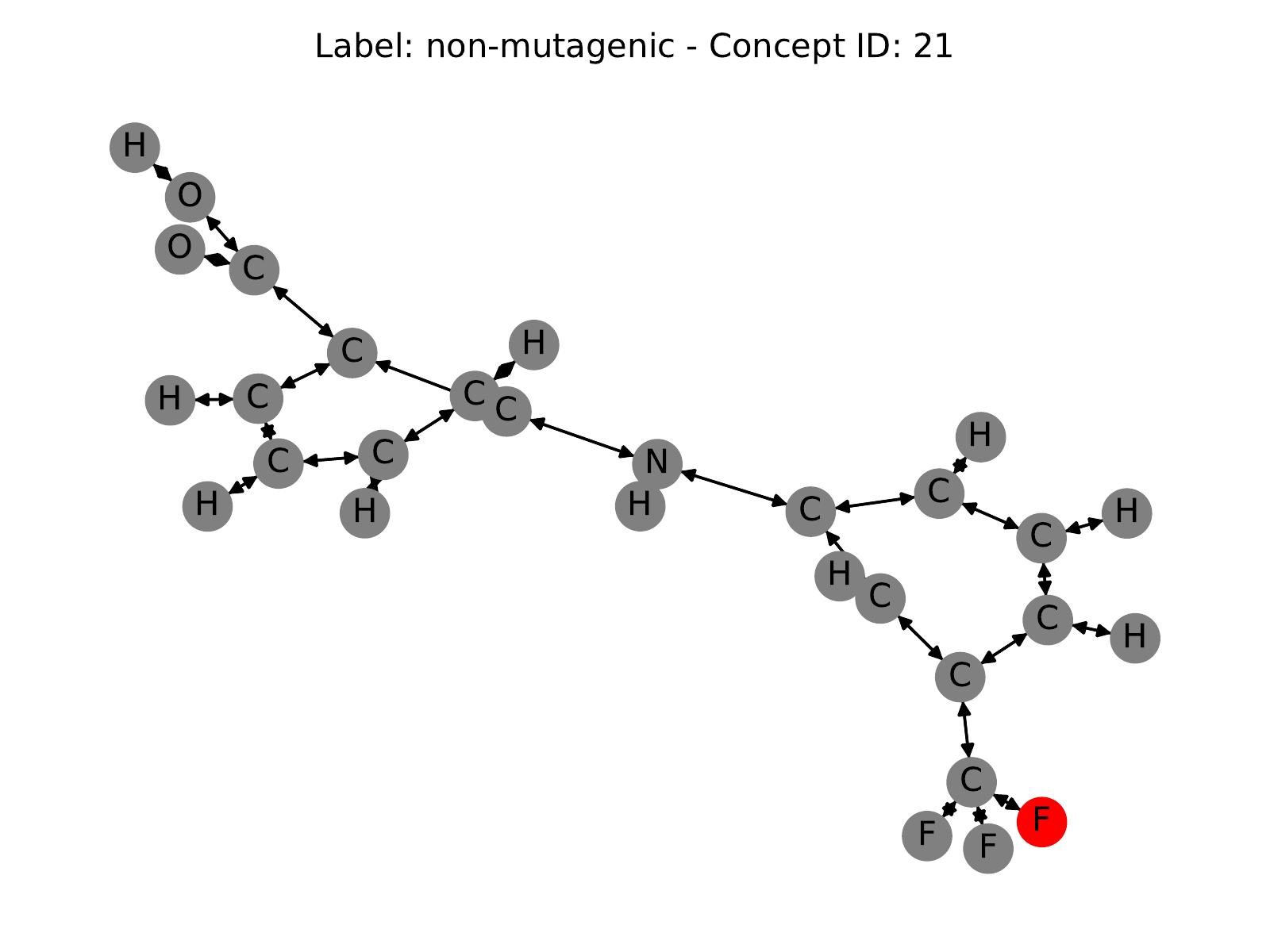}
    \includegraphics[width=0.33\textwidth]{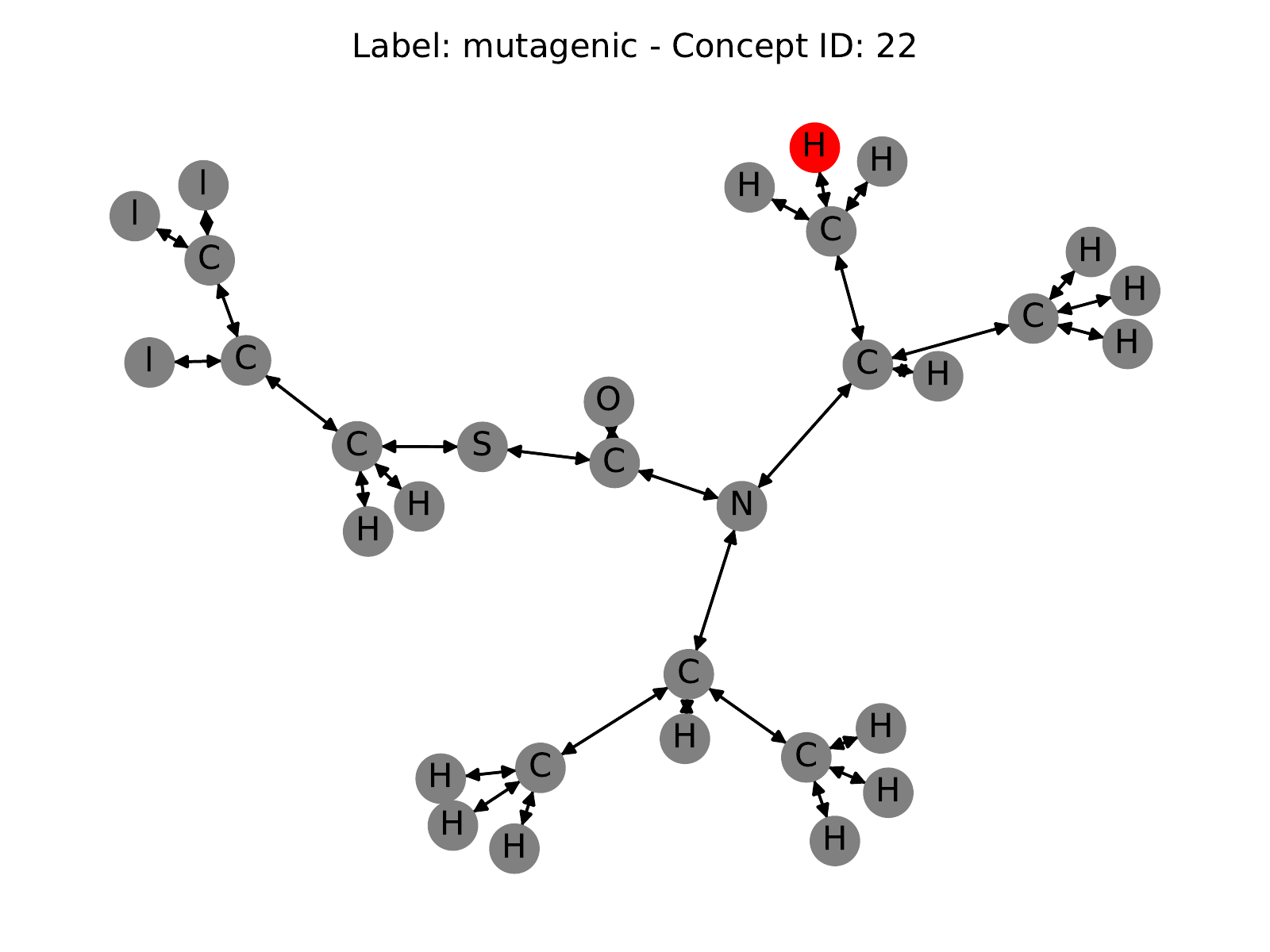}
    \includegraphics[width=0.33\textwidth]{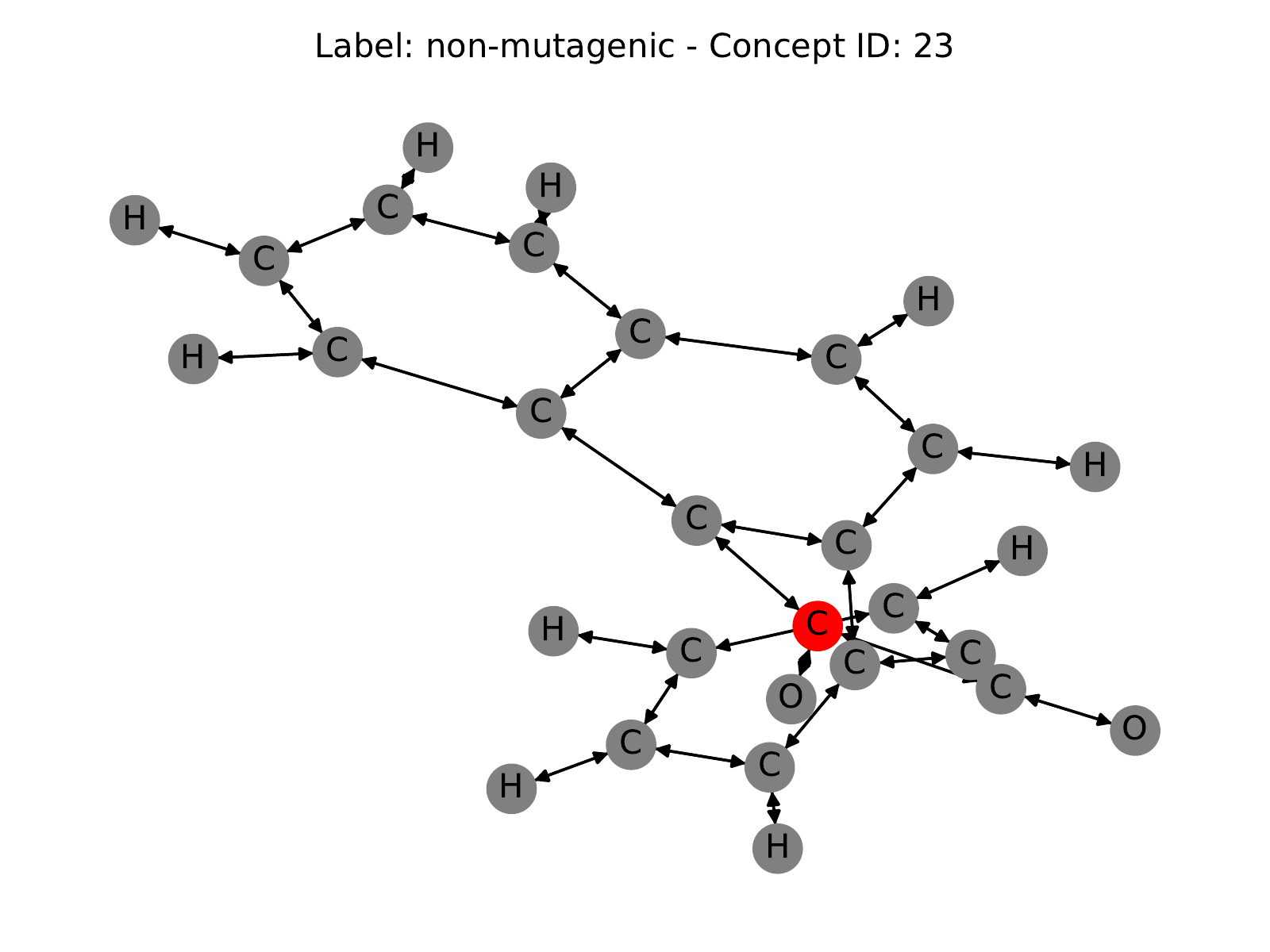}\\
    \includegraphics[width=0.33\textwidth]{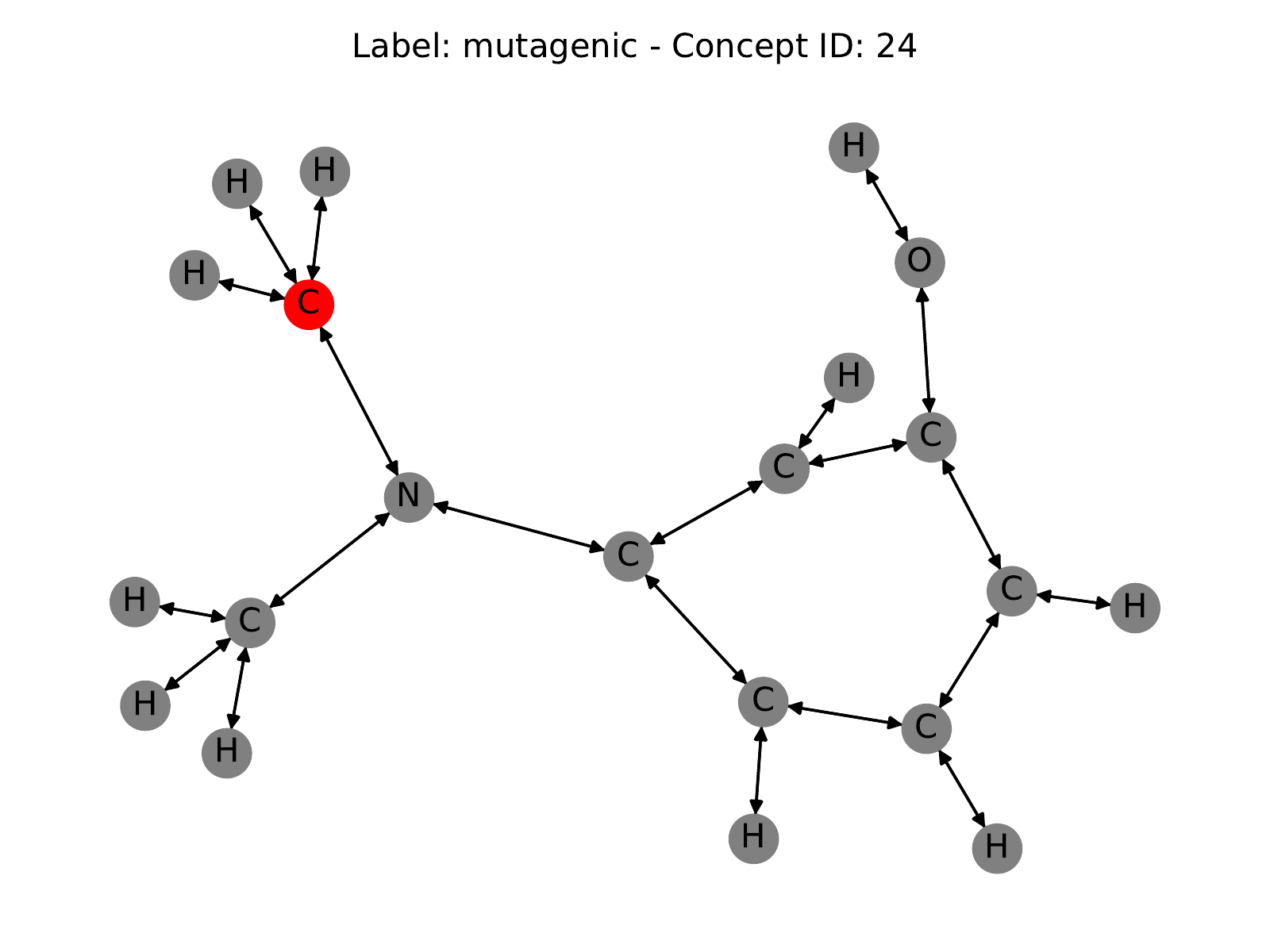}
    \includegraphics[width=0.33\textwidth]{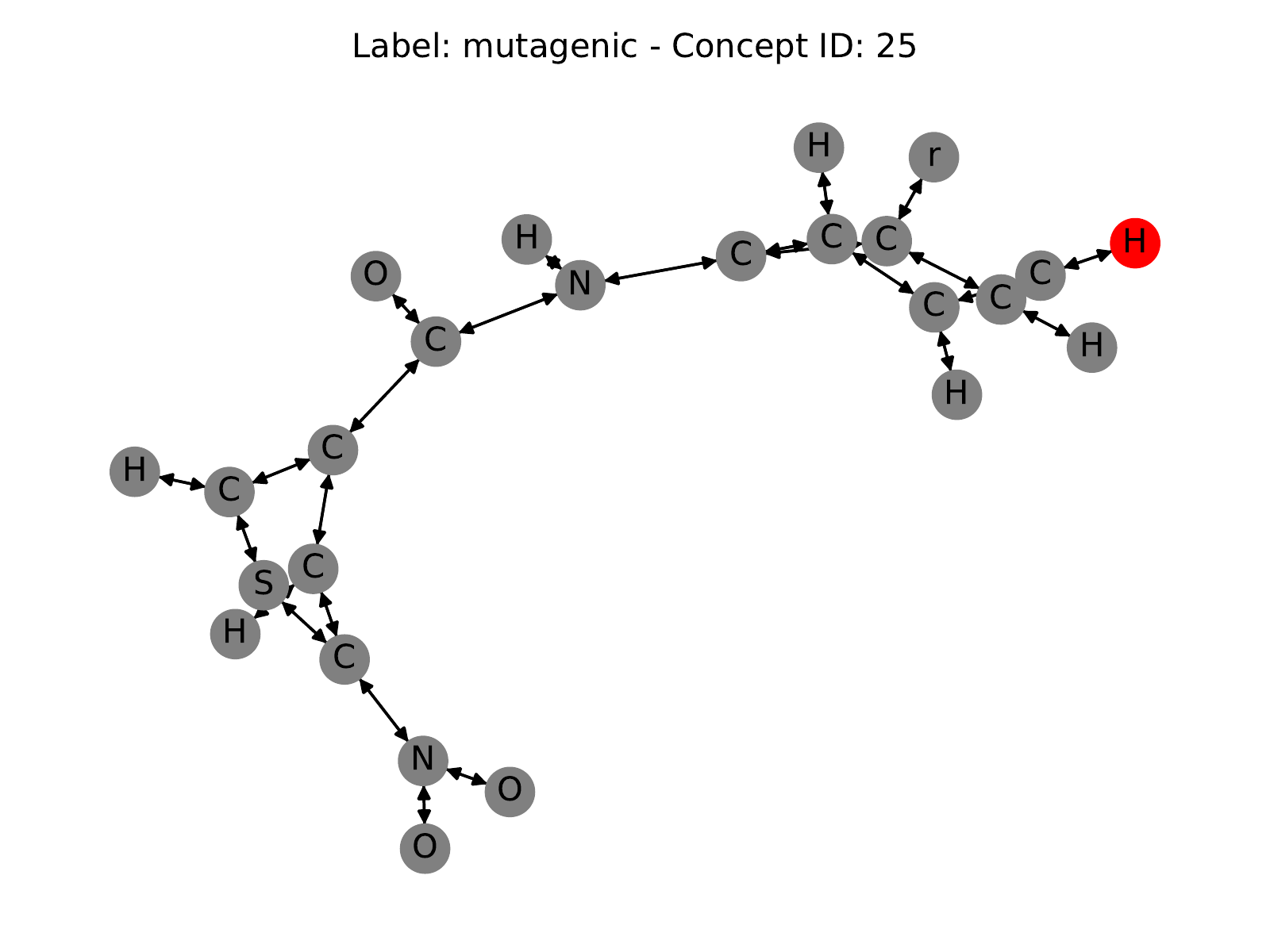}
    \includegraphics[width=0.33\textwidth]{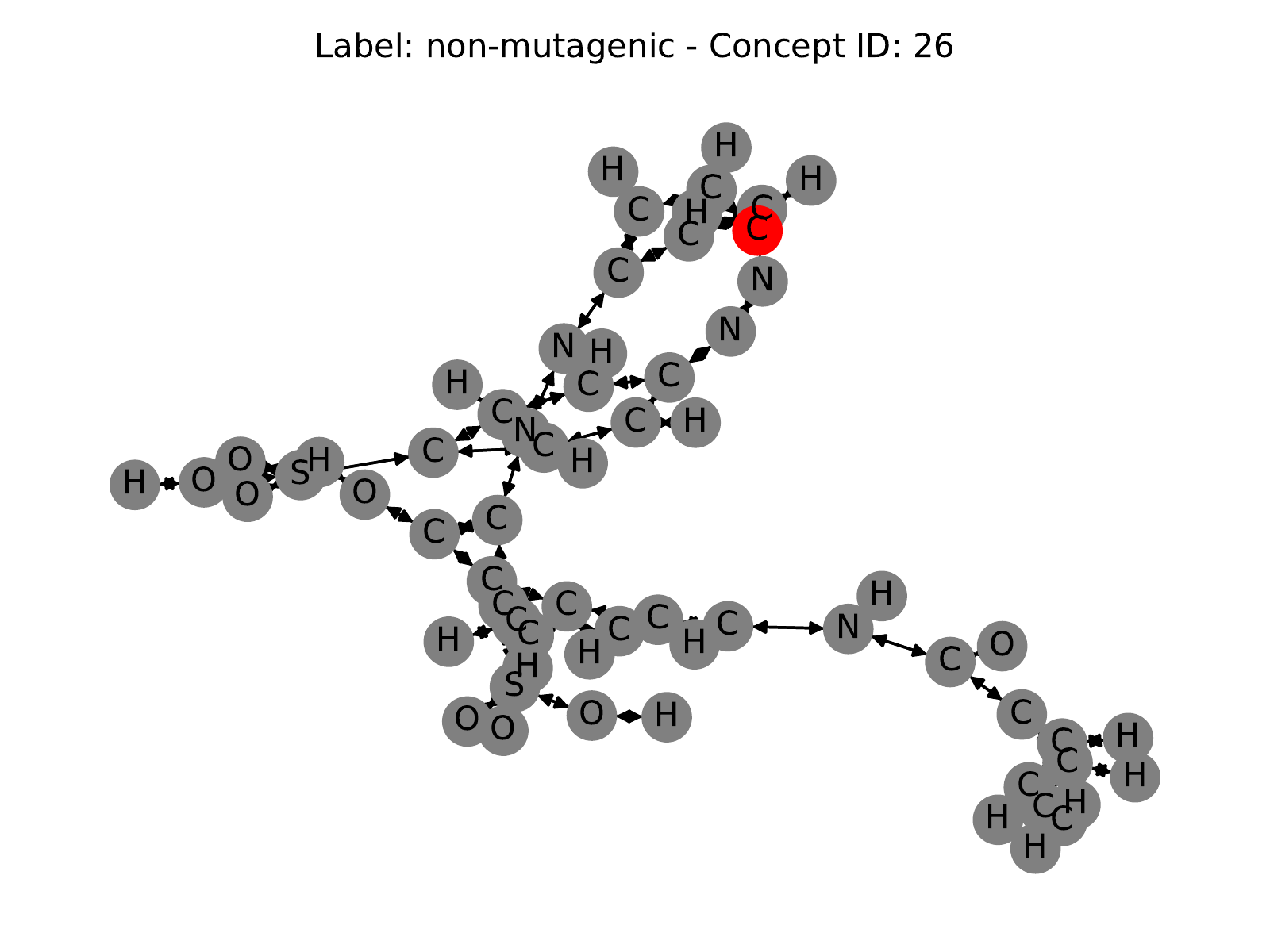}\\
    \includegraphics[width=0.33\textwidth]{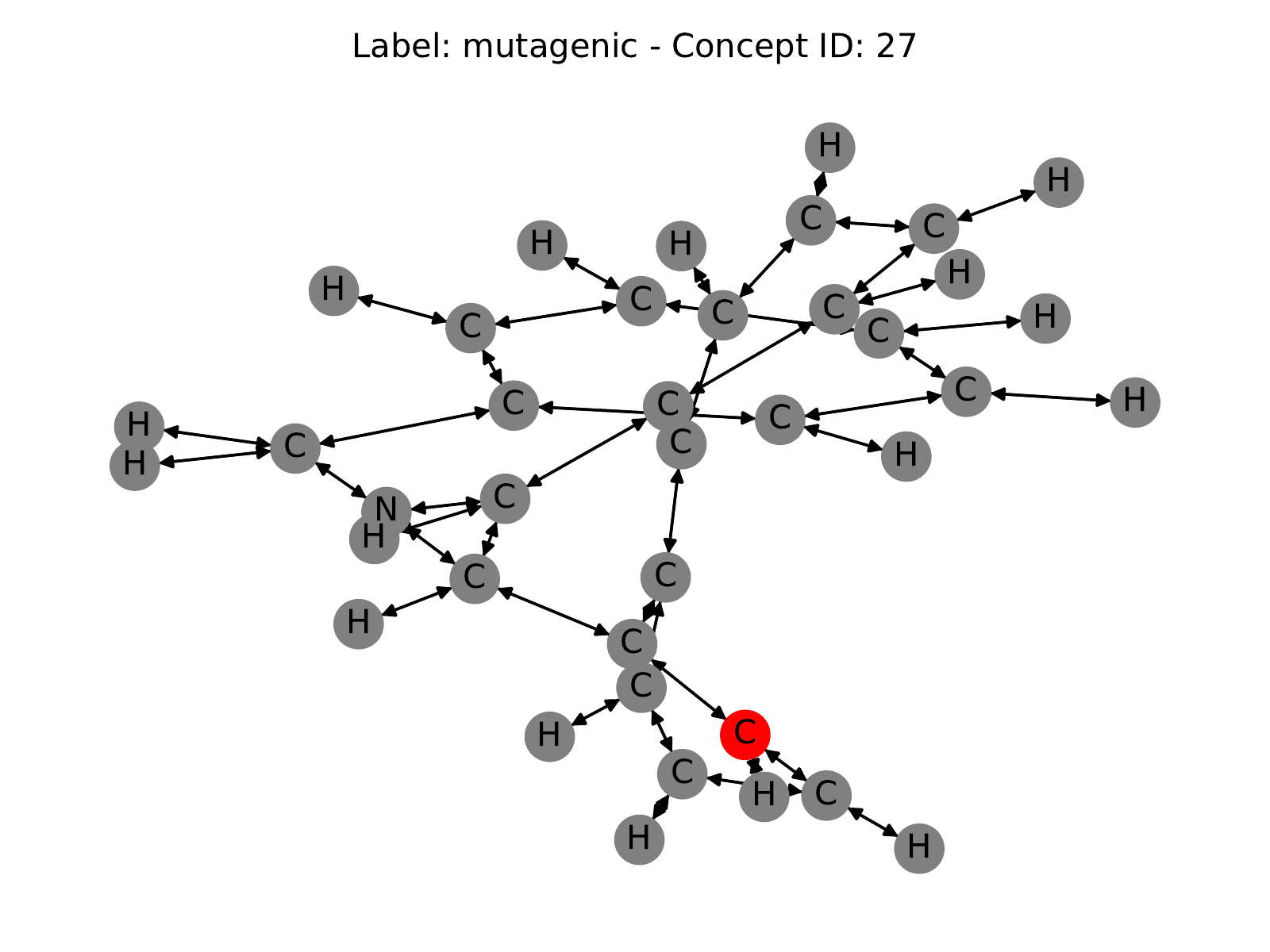}
    \includegraphics[width=0.33\textwidth]{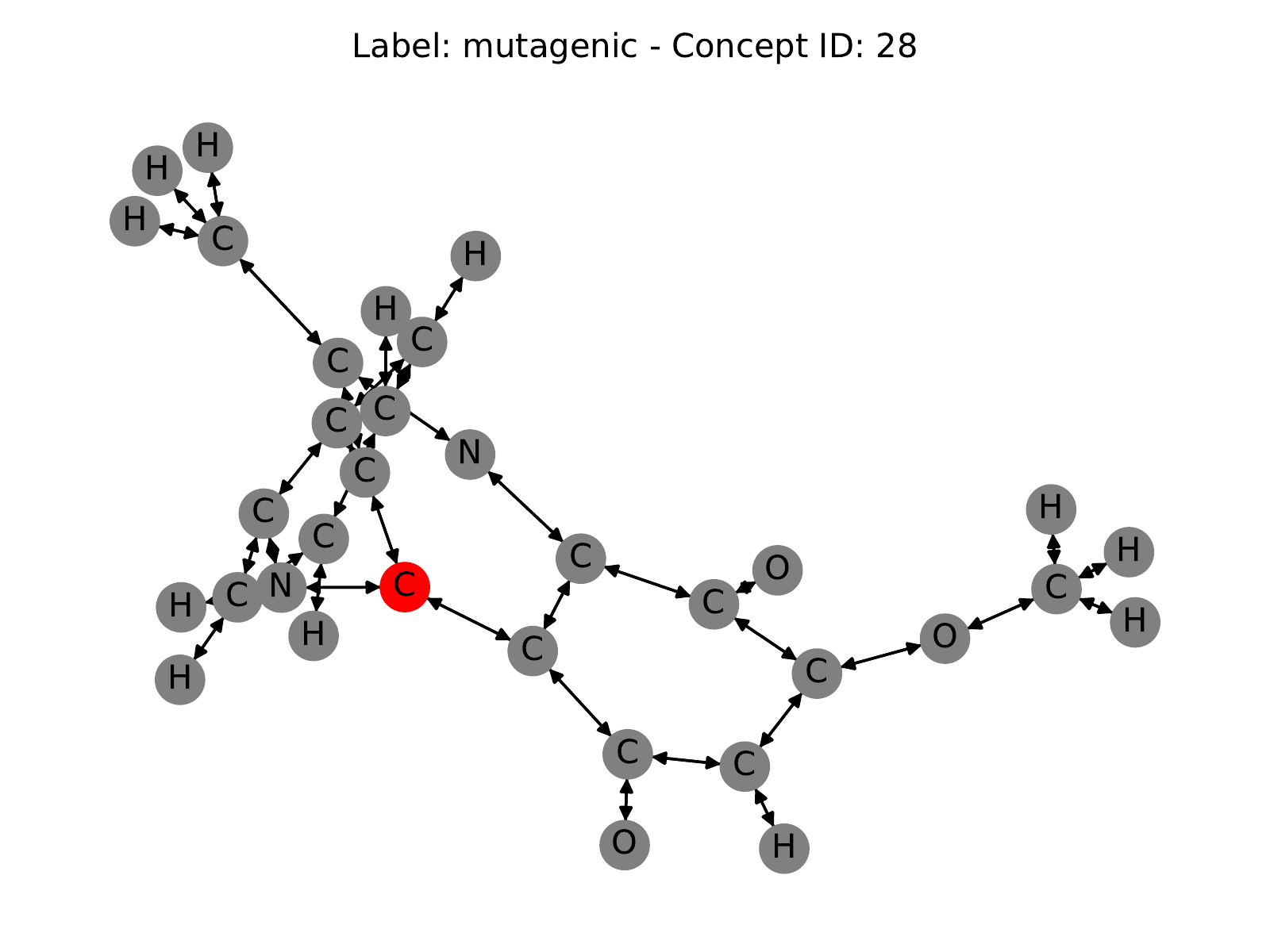}
    \includegraphics[width=0.33\textwidth]{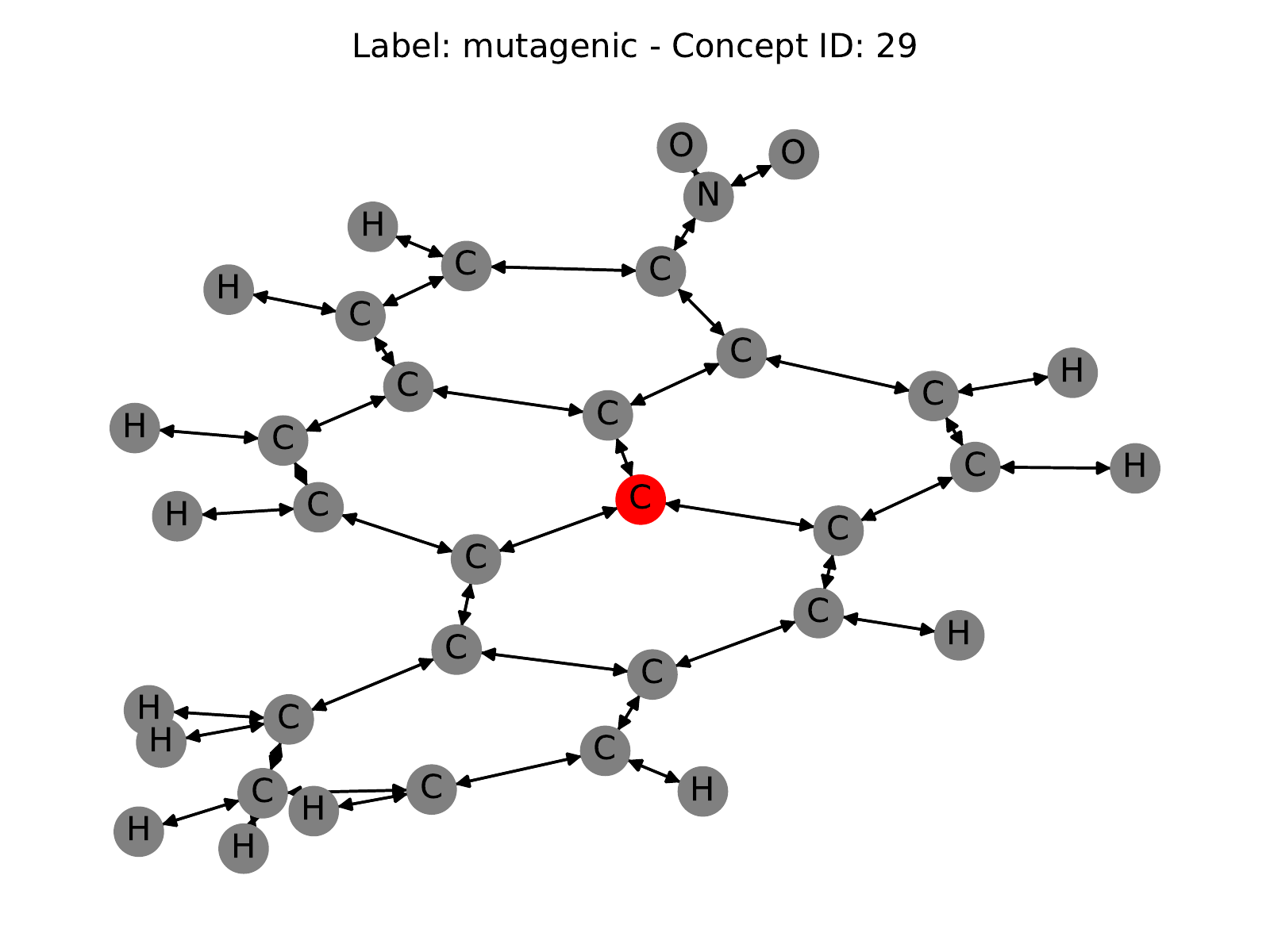}
    \caption{Full molecule corresponding to the closest node embedding to the concept centroid. Part II.}
    \label{fig:mutag6}
\end{figure}

Here we report the visualization of the concepts extracted in \emph{Mutagenicity} by GCExplainer. Following~\citet{magister2021gcexplainer} we represent the concept of a node by expanding and visualizing its $p$-hop neighborhood. In this experiment we set $p=4$ as we used four graph convolutional layers. Figures \ref{fig:mutag1} - \ref{fig:mutag3} show the 30 concepts extracted using GCExplainer when $k = 30$ in k-Means, where the red nodes are the nodes clustered together for a given concept. A human can identify the concept present by reasoning about which features and structures are repeated across the five sample subgraphs, representative of a concept. Using this approach, a number of concepts can be clearly identified. For example, concept 0 (Figure \ref{fig:mutag1}, highlights the importance of the Carbon atom for the prediction that the molecule is mutagenic. In contrast, concepts 8 (Figure \ref{fig:mutag1}) and 28 (Figure \ref{fig:mutag3}) highlight the importance of the star structure in both the prediction of the molecule being mutagenic and non-mutagenic. Concept 11 clearly identified a complex structure of carbon, nitrogen and hydrogen atoms for predicting the label 'mutagenic'. For a complete overview, we visualise the full molecule of the medoids of each cluster in Figures \ref{fig:mutag5} and \ref{fig:mutag6}, highlighting in red the node corresponding to the closest concept. This highlights the size and variety of the molecules classified as different concepts.



\section{Softmax temperature effect on relevant concepts}
We perform an ablation study on the temperature hyperparameter of DCR. This hyperparameter controls the number of concepts selected by DCR to generate rules in the activation function of Equation~\ref{eq:rel}. A low temperature biases DCR towards simpler rules composing fewer concepts, while a high temperature biases DCR towards more complex rules composing many concepts. To assess this we train DCR on the embeddings of a pre-trained Concept Embedding Model on the Caltech-UCSD Birds-200-2011 dataset~\cite{wah2011caltech} as it contains a large number of concepts. We test 7 temperature ranges $\tau \in [0.1, 10]$ and we train DCR using $5$ different initialization seeds.
\begin{figure}[H]
    \centering
    \includegraphics[width=0.45\textwidth]{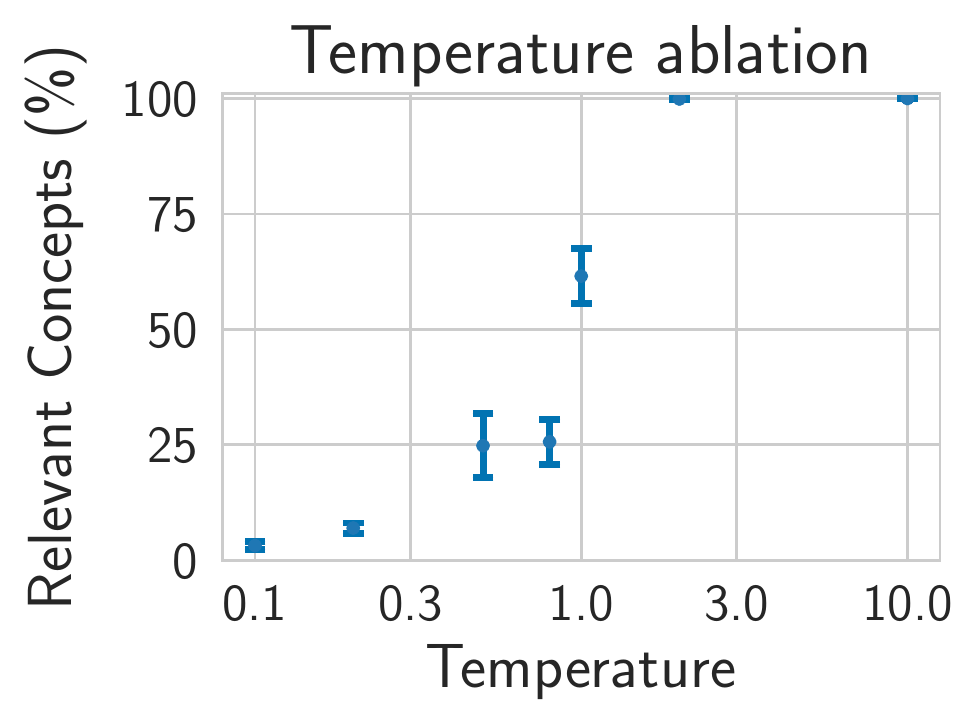}
    \caption{Temperature ablation on pre-trained concept embeddings from the CUB dataset.}
    \label{fig:ablation-temperature}
\end{figure}

\section{Number of concepts effect on training and test time}
We evaluate the computational cost of DCR as a function of the number of training concepts. To this end, we train DCR on the embeddings of a pre-trained Concept Embedding Model on the Caltech-UCSD Birds-200-2011 dataset~\cite{wah2011caltech} as it contains a large number of concepts. We then randomly select $10$, $50$, $100$, and $150$ concepts to train DCR. We train DCR using $5$ different initialization seeds. We observe that the computational time increases linearly when the number of concepts is small, and then it becomes almost constant.

\begin{figure}[H]
    \centering
    \includegraphics[width=0.4\textwidth]{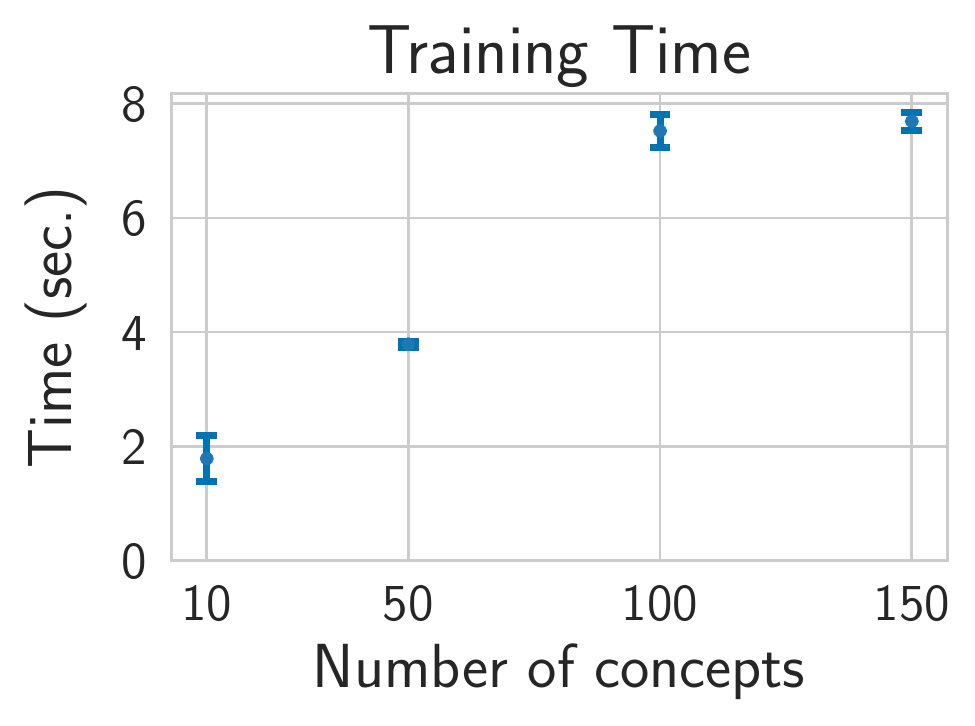} \quad
    \includegraphics[width=0.45\textwidth]{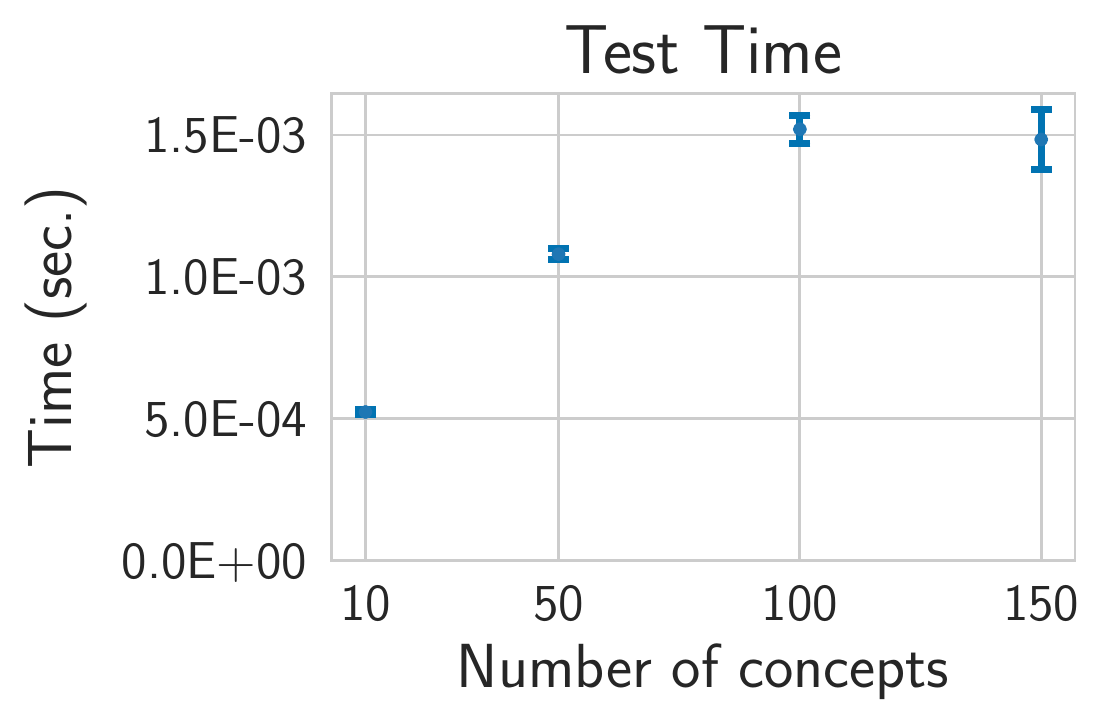}
    \caption{DCR computational time on pre-trained concept embeddings from the CUB dataset.}
    \label{fig:ablation-time}
\end{figure}



\section{Sensitivity analysis}
\label{app:sensitivity}
In Table \ref{tab:exp_sensitivity}, we report the results of the sensitivity analysis comparing DCR with interpretable models and local post-hoc explainers. More precisely, we report the area under the sensitivity curves of Figure \ref{fig:sensitivity}, when increasing the perturbation radius. The lower the values, the more stable the local explanations are on similar samples. These samples $x^\star$ correspond to randomly perturbed sample $x$ drawn by from the test set. More precisely, we draw them from a Gaussian distribution with maximum radius $\epsilon$. These perturbations must be non-significant, i.e., the model prediction must not change. We can see how DCR sensitivity is typically close to existing interpretable models. On the contrary, the compared explanation-based methods Lime and ReluNet have higher sensitivity, strongly reducing the user trust in these explanation methods. Indeed, since samples are very similar and the model predictions do not change, a user expects that also the corresponding explanation should not change, which does not happen for these methods. 

\begin{table}[H]
    \caption{AUC of the explanation sensitivity curves when increasing the perturbation radius $\epsilon$. The lower, the better. }
    \label{tab:exp_sensitivity}
    \centering
    \begin{tabular}{lcccc}
    \toprule
    Model &                        XOR &                       Trig &                        Vec &                      Mutag               \\
    \midrule
    DT              &\bf0.000{\tiny $\pm 0.000$ } &\bf0.000{\tiny $\pm 0.000$ } &\bf0.000{\tiny $\pm 0.000$ } &\bf0.000{\tiny $\pm 0.000$ } \\
    LR              &\bf0.000{\tiny $\pm 0.000$ } &\bf0.000{\tiny $\pm 0.000$ } &\bf0.000{\tiny $\pm 0.000$ } &\bf0.000{\tiny $\pm 0.000$ } \\
    ReluNet         &   0.939{\tiny $\pm 1.301$ } &   0.110{\tiny $\pm 0.181$ } &   0.148{\tiny $\pm 0.247$ } &   0.995{\tiny $\pm 1.480$ } \\
    LIME            &   0.984{\tiny $\pm 0.885$ } &   0.013{\tiny $\pm 0.009$ } &   0.592{\tiny $\pm 0.534$ } &   1.900{\tiny $\pm 0.969$ } \\
    DCR             &\bf0.000{\tiny $\pm 0.000$ } &\bf0.000{\tiny $\pm 0.000$ } &   0.165{\tiny $\pm 0.614$ } &\bf0.000{\tiny $\pm 0.000$ } \\
    \bottomrule
    \end{tabular}
\end{table}

\section{Counterfactual explanations}
\label{app:counterfactual}
In table \ref{tab:counterfactuals} we report the Areas under the model confidence curves of Figure \ref{fig:counterfactuals} when increasing the number of perturbed features. The lower the values, the easier it is to find a counterfactual sample. By comparing DCR with existing interpretable models and local post-hoc explainers, we can see how DCR provides the lowest values in three datasets out of four, confirming that the provided explanations are very precise as they correctly indicate the most important features for a given prediction. 

In Table \ref{tab:counterfactuals_examples}, instead, we report some examples of counterfactual rules provided by DCR on some benchmarked datasets.

\begin{table}[H]
    \caption{Area under the model confidence curves reported in Figure \ref{fig:counterfactuals} against counterfactual samples when increasing the number of perturbed features. The lower, the better. Our model reports the lowest value on three datasets out of four, confirming that DCR explanations can be effectively used to find counterfactual examples.}
    \label{tab:counterfactuals}
    \centering
    \begin{tabular}{lcccc}
    \toprule
    Model &       XOR &      Trig &       Vec &     Mutag \\
    \midrule
    DT          &\bf0.339{\tiny $\pm 0.468$ } &\bf0.395{\tiny $\pm 0.380$ } &\bf0.443{\tiny $\pm 0.442$ } &\bf0.185{\tiny $\pm 0.311$ } \\
    LR          &   0.992{\tiny $\pm 0.015$ } &   0.451{\tiny $\pm 0.402$ } &   0.530{\tiny $\pm 0.391$ } &   0.347{\tiny $\pm 0.351$ } \\
    ReluNet      &   0.622{\tiny $\pm 0.476$ } &   0.469{\tiny $\pm 0.429$ } &   0.448{\tiny $\pm 0.457$ } &\bf0.279{\tiny $\pm 0.387$ } \\
    LIME        &   0.674{\tiny $\pm 0.462$ } &   0.424{\tiny $\pm 0.422$ } &   0.450{\tiny $\pm 0.438$ } &   0.249{\tiny $\pm 0.372$ } \\
    XGBoost     &   0.680{\tiny $\pm 0.460$ } &   0.739{\tiny $\pm 0.431$ } &   0.804{\tiny $\pm 0.426$ } &   0.924{\tiny $\pm 0.226$ } \\
    DCR         &\bf0.344{\tiny $\pm 0.505$ } &\bf0.255{\tiny $\pm 0.436$ } &\bf0.394{\tiny $\pm 0.489$ } &   0.705{\tiny $\pm 0.467$ } \\
    \bottomrule
    \end{tabular}
\end{table}

\begin{table}[H]
\centering
\caption{Counterfactuals (Boolean)}
\label{tab:counterfactuals_examples}
\resizebox{0.8\textwidth}{!}{%
\begin{tabular}{lllll}
\hline
\textbf{Dataset} & \textbf{Old Concepts} & \textbf{Old prediction} & \textbf{New Concepts} & \textbf{New Prediction} \\ \hline
XOR & $\neg f0$, $\neg f1$ & $y=0$ & $\neg f0$, $ f1$ & $y=1$ \\
XOR & $f0$, $\neg f1$ & $y=1$ & $\neg f0$, $\neg f1$ & $y=0$ \\
Trigonometry & $\neg f0$, $\neg f1$, $\neg f2$ & $y=0$ & $\neg f0$, $f1$, $f2$ & $y=1$ \\
Trigonometry & $f0$, $f1$, $\neg f2$ & $y=1$ & $\neg f0$, $\neg f1$, $\neg f2$ & $y=0$ \\
Vector & $f0$, $\neg f1$ & $y=0$ & $\neg f0$, $f1$ & $y=1$ \\
Vector & $\neg f0$, $\neg f1$ & $y=1$ & $f0$, $f1$ & $y=0$ \\
CelebA & $\neg f0$, $\neg f1$, $\neg f2$ & $y=0$ & $f0$, $f1$, $f2$ & $y=4$ \\
\bottomrule
\end{tabular}%
}
\end{table}

\section{MNIST addition experiment}
\label{app:mnist}
In this experiment, we tested DCR in a task where it is not provided with any label on the concepts. In the MNIST addition dataset \cite{manhaeve2018deepproblog}, pairs of MNIST images are labelled with the sum of the corresponding digit. The single images are, therefore, never labelled. The idea behind the task is that an image classifier can still be  asked to predict the class of the single images, while a differentiable symbolic program can be used to map the class of the images to their sum. In terms of learning, the knowledge of both the label on the addition and the symbolic program provides a distant supervision signal to the image classifier. 

This task can be easily mapped in terms of a concept-based model. The output of the classifier for the two images constitutes a set of 20 concepts (i.e. 10 class predictions for each of the two images). The set of all possible additions constitutes a set of 19 tasks. The MNIST addition task could be considered a first example of a more structured (i.e. relational) setting, where the input is a list of two images. However, it is still simple enough not to require any specific modelling. 

The absence of direct supervision on the concepts puts our system in a different regime. In fact, there is no loss that forces the concept probabilities to represent crisp decisions. The softmax activation function tends to crisp decisions only when coupled with a categorical cross-entropy loss. In the absence of such loss, the network can still exploit the entire categorical distribution as an embedding to latently encode the identity of the digits.

Our solution to the absence of a concept loss is made of two ingredients. First, the softmax output distribution is substituted with a Gumbel-softmax sampling layer. The Gumbel-softmax forces the network to always make crisp decisions by sampling from the corresponding categorical distribution. Notice that a categorical distribution and its one-hot samples coincide when the distribution becomes very peaked on its prediction (e.g. at the end of the learning).
Second, we introduce a second task predictor function $f_{NN}: C \to Y$, that akin to standard concept bottleneck models, predicts the task only from the probabilities, and we add a corresponding loss encouraging $f_{NN}(g(\mathbf{x})) = \mathbf{y}$. The goal here is to force the model to exploit (and thus learn) the concept probabilities $\hat{c}_i$ and not to rely only on their embeddings $\hat{\mathbf{c}}_i$. 

In Table~\ref{tab:mnist-addition}, we show the comparison with state-of-the-art Neural Symbolic frameworks, as described in the main text. Moreover, in Table~\ref{tab:mnist-addition-rules}, we show the entire list of global rules learned by DCR, showing that it actually captured perfectly the semantics of the addition relation. 

\begin{table}[H]
    \centering
    \caption{MNIST addition global rules for 10000 training examples. $f_{ij}$ reads "class of the digit in position $i$ is $j$. Therefore, the rule $y_{0} \leftarrow f_{00} \land f_{10}$ means that if the first digit is a $0$ and the second digit is a $0$ then the sum is a $0$.  The semantics is correct except for a single rule $y_{8} \leftarrow f_{03} \land f_{16}$, which is easily identifiable as having a count of 1. Notice that we had to map the network concept IDs to the corresponding human digits, as there was no supervision on concepts during training.
    }
    \label{tab:mnist-addition-rules}
\scalebox{0.8}{
    \begin{tabular}{cc}
         \textbf{\textsc{Rule}} & \textbf{\textsc{Count}} \\
$y_{0} \leftarrow f_{00} \land f_{10}$ & $ 93 $ \\
$y_{1} \leftarrow f_{00} \land f_{11}$ & $ 110 $ \\
$y_{1} \leftarrow f_{01} \land f_{10}$ & $ 102 $ \\
$y_{2} \leftarrow f_{00} \land f_{12}$ & $ 89 $ \\
$y_{2} \leftarrow f_{01} \land f_{11}$ & $ 119 $ \\
$y_{2} \leftarrow f_{02} \land f_{10}$ & $ 101 $ \\
$y_{3} \leftarrow f_{01} \land f_{12}$ & $ 124 $ \\
$y_{3} \leftarrow f_{03} \land f_{10}$ & $ 96 $ \\
$y_{3} \leftarrow f_{02} \land f_{11}$ & $ 115 $ \\
$y_{3} \leftarrow f_{00} \land f_{13}$ & $ 100 $ \\
$y_{4} \leftarrow f_{03} \land f_{11}$ & $ 121 $ \\
$y_{4} \leftarrow f_{04} \land f_{10}$ & $ 84 $ \\
$y_{4} \leftarrow f_{01} \land f_{13}$ & $ 137 $ \\
$y_{4} \leftarrow f_{02} \land f_{12}$ & $ 105 $ \\
$y_{4} \leftarrow f_{00} \land f_{14}$ & $ 112 $ \\
$y_{5} \leftarrow f_{01} \land f_{14}$ & $ 104 $ \\
$y_{5} \leftarrow f_{03} \land f_{12}$ & $ 105 $ \\
$y_{5} \leftarrow f_{04} \land f_{11}$ & $ 113 $ \\
$y_{5} \leftarrow f_{00} \land f_{15}$ & $ 95 $ \\
$y_{5} \leftarrow f_{02} \land f_{13}$ & $ 90 $ \\
$y_{5} \leftarrow f_{05} \land f_{10}$ & $ 95 $ \\
$y_{6} \leftarrow f_{02} \land f_{14}$ & $ 92 $ \\
$y_{6} \leftarrow f_{05} \land f_{11}$ & $ 96 $ \\
$y_{6} \leftarrow f_{00} \land f_{16}$ & $ 109 $ \\
$y_{6} \leftarrow f_{04} \land f_{12}$ & $ 91 $ \\
$y_{6} \leftarrow f_{01} \land f_{15}$ & $ 86 $ \\
$y_{6} \leftarrow f_{03} \land f_{13}$ & $ 107 $ \\
$y_{6} \leftarrow f_{06} \land f_{10}$ & $ 92 $ \\
$y_{7} \leftarrow f_{00} \land f_{17}$ & $ 100 $ \\
$y_{7} \leftarrow f_{04} \land f_{13}$ & $ 108 $ \\
$y_{7} \leftarrow f_{01} \land f_{16}$ & $ 103 $ \\
$y_{7} \leftarrow f_{02} \land f_{15}$ & $ 81 $ \\
$y_{7} \leftarrow f_{07} \land f_{10}$ & $ 103 $ \\
$y_{7} \leftarrow f_{06} \land f_{11}$ & $ 137 $ \\
$y_{7} \leftarrow f_{05} \land f_{12}$ & $ 87 $ \\
$y_{7} \leftarrow f_{03} \land f_{14}$ & $ 117 $ \\
$y_{8} \leftarrow f_{05} \land f_{13}$ & $ 72 $ \\
$y_{8} \leftarrow f_{01} \land f_{17}$ & $ 122 $ \\
$y_{8} \leftarrow f_{03} \land f_{15}$ & $ 99 $ \\
$y_{8} \leftarrow f_{02} \land f_{16}$ & $ 97 $ \\
$y_{8} \leftarrow f_{06} \land f_{12}$ & $ 90 $ \\
$y_{8} \leftarrow f_{08} \land f_{10}$ & $ 96 $ \\
$y_{8} \leftarrow f_{07} \land f_{11}$ & $ 116 $ \\
$y_{8} \leftarrow f_{04} \land f_{14}$ & $ 106 $ \\
$y_{8} \leftarrow f_{00} \land f_{18}$ & $ 100 $ \\
$y_{8} \leftarrow f_{03} \land f_{16}$ & $ 1 $ \\
$y_{9} \leftarrow f_{04} \land f_{15}$ & $ 87 $ \\
$y_{9} \leftarrow f_{08} \land f_{11}$ & $ 112 $ \\
$y_{9} \leftarrow f_{06} \land f_{13}$ & $ 76 $ \\
$y_{9} \leftarrow f_{01} \land f_{18}$ & $ 113 $ \\
$y_{9} \leftarrow f_{00} \land f_{19}$ & $ 94 $ \\
\end{tabular}
\hspace{4cm}
\begin{tabular}{cc}
         \textbf{\textsc{Rule}} & \textbf{\textsc{Count}} \\
$y_{9} \leftarrow f_{03} \land f_{16}$ & $ 89 $ \\
$y_{9} \leftarrow f_{09} \land f_{10}$ & $ 100 $ \\
$y_{9} \leftarrow f_{07} \land f_{12}$ & $ 110 $ \\
$y_{9} \leftarrow f_{02} \land f_{17}$ & $ 102 $ \\
$y_{9} \leftarrow f_{05} \land f_{14}$ & $ 89 $ \\
$y_{10} \leftarrow f_{01} \land f_{19}$ & $ 115 $ \\
$y_{10} \leftarrow f_{06} \land f_{14}$ & $ 97 $ \\
$y_{10} \leftarrow f_{09} \land f_{11}$ & $ 100 $ \\
$y_{10} \leftarrow f_{08} \land f_{12}$ & $ 100 $ \\
$y_{10} \leftarrow f_{07} \land f_{13}$ & $ 113 $ \\
$y_{10} \leftarrow f_{04} \land f_{16}$ & $ 94 $ \\
$y_{10} \leftarrow f_{03} \land f_{17}$ & $ 89 $ \\
$y_{10} \leftarrow f_{02} \land f_{18}$ & $ 103 $ \\
$y_{10} \leftarrow f_{05} \land f_{15}$ & $ 75 $ \\
$y_{11} \leftarrow f_{08} \land f_{13}$ & $ 89 $ \\
$y_{11} \leftarrow f_{03} \land f_{18}$ & $ 105 $ \\
$y_{11} \leftarrow f_{07} \land f_{14}$ & $ 94 $ \\
$y_{11} \leftarrow f_{09} \land f_{12}$ & $ 97 $ \\
$y_{11} \leftarrow f_{04} \land f_{17}$ & $ 111 $ \\
$y_{11} \leftarrow f_{05} \land f_{16}$ & $ 86 $ \\
$y_{11} \leftarrow f_{02} \land f_{19}$ & $ 105 $ \\
$y_{11} \leftarrow f_{06} \land f_{15}$ & $ 104 $ \\
$y_{12} \leftarrow f_{03} \land f_{19}$ & $ 98 $ \\
$y_{12} \leftarrow f_{04} \land f_{18}$ & $ 87 $ \\
$y_{12} \leftarrow f_{06} \land f_{16}$ & $ 105 $ \\
$y_{12} \leftarrow f_{07} \land f_{15}$ & $ 96 $ \\
$y_{12} \leftarrow f_{09} \land f_{13}$ & $ 106 $ \\
$y_{12} \leftarrow f_{05} \land f_{17}$ & $ 94 $ \\
$y_{12} \leftarrow f_{08} \land f_{14}$ & $ 87 $ \\
$y_{13} \leftarrow f_{06} \land f_{17}$ & $ 106 $ \\
$y_{13} \leftarrow f_{08} \land f_{15}$ & $ 85 $ \\
$y_{13} \leftarrow f_{09} \land f_{14}$ & $ 82 $ \\
$y_{13} \leftarrow f_{07} \land f_{16}$ & $ 118 $ \\
$y_{13} \leftarrow f_{05} \land f_{18}$ & $ 79 $ \\
$y_{13} \leftarrow f_{04} \land f_{19}$ & $ 100 $ \\
$y_{14} \leftarrow f_{06} \land f_{18}$ & $ 105 $ \\
$y_{14} \leftarrow f_{07} \land f_{17}$ & $ 98 $ \\
$y_{14} \leftarrow f_{05} \land f_{19}$ & $ 78 $ \\
$y_{14} \leftarrow f_{09} \land f_{15}$ & $ 74 $ \\
$y_{14} \leftarrow f_{08} \land f_{16}$ & $ 101 $ \\
$y_{15} \leftarrow f_{09} \land f_{16}$ & $ 107 $ \\
$y_{15} \leftarrow f_{08} \land f_{17}$ & $ 95 $ \\
$y_{15} \leftarrow f_{07} \land f_{18}$ & $ 103 $ \\
$y_{15} \leftarrow f_{06} \land f_{19}$ & $ 111 $ \\
$y_{16} \leftarrow f_{07} \land f_{19}$ & $ 115 $ \\
$y_{16} \leftarrow f_{09} \land f_{17}$ & $ 100 $ \\
$y_{16} \leftarrow f_{08} \land f_{18}$ & $ 84 $ \\
$y_{17} \leftarrow f_{09} \land f_{18}$ & $ 100 $ \\
$y_{17} \leftarrow f_{08} \land f_{19}$ & $ 86 $ \\
$y_{18} \leftarrow f_{09} \land f_{19}$ & $ 102 $ \\
& \\
          
    \end{tabular}
}
\end{table}

Our solution to the MNIST addition task shows that  DCR can be enhanced with an unsupervised (or distantly supervised) criterion for the learning of meaningful concepts. This creates interesting links with generative models for learning representations, but we leave such interpretation for future works.

The architecture of the image classifiers is those in \cite{manhaeve2018deepproblog}. The additional task network is MLP with 1 hidden layer of 30 hidden neurons and relu activations.  We searched over the following grid of parameters (bold selected): embedding size [10, 20, \textbf{30}, 50]; gumbel-softmax temperature [1, 1.25, 1.50, \textbf{1.75}, 2.0].

\section{Complexity of logic rules}
We compute rule complexity as the average size of the learnt logic rules. Table~\ref{tab:tab-complexity} summarizes the main outcomes comparing DCR rules with decision tree rules. In most datasets, such as Trigonometry, Dot, or CelebA, the rule complexity of DCR matches that of decision tree rules while providing superior task performance. However, in Mutagenicity, there is a tradeoff between performance and complexity compared to decision trees. Nevertheless, we don't observe a significant increase in rule complexity as shown in the plot, partly because DCR rules are "per sample." However, if we were to learn global rules, the complexity would likely increase, especially if multiple combinations of concepts could result in the same task prediction. It is worth noting that overly complex rules may not be a machine error, but rather a limitation of the human side. For example, asking a model to explain complex tasks using raw features like pixel intensities as concepts would lead to complex rules.

\begin{table}[H]
\centering
\caption{Complexity of logic rules}
\label{tab:tab-complexity}
\begin{tabular}{llll}
\toprule
 & \textbf{CE+DCR (ours)} & \textbf{CT+Decision Tree} & \textbf{CE+Decision Tree} \\
 \midrule
XOR & $2.00 \pm 0.00$ & $2.00 \pm 0.00$ & $1.40 \pm 0.16$ \\
Trigonometry & $3.00 \pm 0.00$ & $3.00 \pm 0.00$ & $1.40 \pm 0.16$ \\
Dot & $2.00 \pm 0.00$ & $2.00 \pm 0.00$ & $1.93 \pm 0.07$ \\
Mutagenicity & $13.57 \pm 0.62$ & $4.84 \pm 0.74$ & $2.35 \pm 0.35$ \\
CelebA & $1.00 \pm 0.00$ & $1.00 \pm 0.00$ & $5.86 \pm 0.56$\\
\bottomrule
\end{tabular}%
\end{table}

\section{Code, Licences, Resources}
\label{sec:appendix_code}

\paragraph{Libraries} For our experiments, we implemented all baselines and methods in Python 3.7 and relied upon open-source libraries such as PyTorch 1.11~\citep{paszke2019pytorch} (BSD license) and Scikit-learn~\citep{ pedregosa2011scikit} (BSD license). To produce the plots seen in this paper, we made use of Matplotlib  3.5 (BSD license). We will release all of the code required to recreate our experiments in an MIT-licensed public repository.

\paragraph{Resources} All of our experiments were run on a private machine with 8 Intel(R) Xeon(R) Gold 5218 CPUs (2.30GHz), 64GB of RAM, and 2 Quadro RTX 8000 Nvidia GPUs. We estimate that approximately 240-GPU hours were required to complete all of our experiments.



\end{document}